%% file: main.tex
\documentclass[10pt,twocolumn,letterpaper]{article}

\PassOptionsToPackage{table}{xcolor}

\usepackage[pagenumbers]{cvpr}

\input{preamble}

\definecolor{cvprblue}{rgb}{0.21,0.49,0.74}
\usepackage[
    pagebackref,
    breaklinks,
    colorlinks,
    allcolors=cvprblue
]{hyperref}

\def\confName{CVPR}
\def\confYear{2027}

\input{paper_info}

\begin{document}


\twocolumn[{
    \maketitle
    \input{figs/teaser}
    \bigbreak
}]

\begin{abstract}
\input{sec/0_abstract}
\end{abstract}

\section{Introduction}
\input{sec/1_intro}

\section{Related Works}
\input{sec/2_related}

\section{Methods}
\input{sec/3_method}

\section{Experiments}
\input{sec/4_experiment}

\section{Conclusion}
\input{sec/5_conclusion}


\clearpage



{
    \small
    \bibliographystyle{ieeenat_fullname}
    \bibliography{main}
}


\clearpage
\onecolumn

\begin{center}
    \Large
    \textbf{\thetitle}\\
    \vspace{0.5em}
    Supplementary Material\\
    \vspace{1.0em}
\end{center}

\appendix

\setcounter{figure}{0}
\setcounter{table}{0}
\setcounter{equation}{0}

\renewcommand{\thefigure}{\thesection.\arabic{figure}}
\renewcommand{\thetable}{\thesection.\arabic{table}}
\renewcommand{\theequation}{\thesection.\arabic{equation}}

\renewcommand{\theHsection}{appendix.\Alph{section}}
\renewcommand{\theHfigure}{appendix.\Alph{section}.\arabic{figure}}
\renewcommand{\theHtable}{appendix.\Alph{section}.\arabic{table}}
\renewcommand{\theHequation}{appendix.\Alph{section}.\arabic{equation}}

\makeatletter
\@addtoreset{figure}{section}
\@addtoreset{table}{section}
\@addtoreset{equation}{section}
\makeatother


\input{sec/appendix_contents}

\clearpage


\input{sec/6_appendix}

\end{document}

%% file: preamble.tex
\input{math_commands.tex}

\usepackage{multirow}
\usepackage{float}
\usepackage{wrapfig}
\usepackage{pifont}
\usepackage{algorithm}
\usepackage{algpseudocode}
\usepackage{titletoc}
\usepackage{marvosym}
\definecolor{colorfirst}{RGB}{220,242,220}
\definecolor{colorsecond}{RGB}{255,244,204}

\newcommand{\cellfirst}[1]{\cellcolor{colorfirst}{#1}}
\newcommand{\cellsecond}[1]{\cellcolor{colorsecond}{#1}}

\newcommand{\cmark}{\textcolor{green!60!black}{\ding{51}}}
\newcommand{\xmark}{\textcolor{red!75!black}{\ding{55}}}

\newcommand{\OurMethod}{COHERE4D}
\newcommand{\HumanThreeR}{Human3R}
\newcommand{\UniSHMethod}{UniSH}
\newcommand{\SHOWMethod}{SHOW}
\newcommand{\MHMRtwo}{Multi-HMR~2}
\newcommand{\SAMThreeDBody}{3DB}
\newcommand{\VGGTOmega}{VGGT-$\Omega$}
\newcommand{\DAThree}{DA3}
\newcommand{\GVHMRMethod}{GVHMR}
\newcommand{\PiThreeX}{Pi3X}
\newcommand{\CHROMMMethod}{CHROMM}
\newcommand{\UniConThreeR}{UniCon3R}
\newcommand{\MetricHMSRMethod}{MetricHMSR}
\newcommand{\HSfMMethod}{HSfM}
\newcommand{\HMRtwoMethod}{HMR2.0}
\newcommand{\JOSHMethod}{JOSH}
\newcommand{\JOSHThreeR}{JOSH3R}

\newcommand{\GEMMethod}{GEM}

\newcommand{\FastSAMThreeDBody}{Fast SAM 3D Body}

\makeatletter
\patchcmd{\@maketitle}
  {\vspace*{24pt}}
  {\vspace*{14pt}}
  {}{}
\makeatother

\makeatletter
\newcommand{\wrapfill}{  \par
  \ifnum\value{WF@wrappedlines}>0
    \addtocounter{WF@wrappedlines}{-1}    \null\vspace{\arabic{WF@wrappedlines}\baselineskip}    \WFclear
  \fi
}
\makeatother

\newcommand{\bestmark}{  \begingroup
  \setlength{\fboxsep}{0pt}  \colorbox{colorfirst}{    \rule{0pt}{1.2ex}\hspace{1.2ex}  }  \endgroup
}

\newcommand{\secondmark}{  \begingroup
  \setlength{\fboxsep}{0pt}  \colorbox{colorsecond}{    \rule{0pt}{1.2ex}\hspace{1.2ex}  }  \endgroup
}

%% file: math_commands.tex
\usepackage{amsmath,amsfonts,bm}

\def\eqref#1{equation~\ref{#1}}

\def\1{\bm{1}}

\DeclareMathAlphabet{\mathsfit}{\encodingdefault}{\sfdefault}{m}{sl}
\SetMathAlphabet{\mathsfit}{bold}{\encodingdefault}{\sfdefault}{bx}{n}

%% file: paper_info.tex
\title{\OurMethod: Training-Free Dense Surface Alignment for\\
Coherent Metric 4D Human--Scene Reconstruction}

\author{%
\begin{tabular}{c}
\large
Zhumei Wang\textsuperscript{1,2}
\quad
Shaofei Wang\textsuperscript{2,\Letter}
\quad
Xiang An\textsuperscript{3}
\quad
Mingtao Pei\textsuperscript{1}
\quad
Siyuan Huang\textsuperscript{2,\Letter}
\\[0.25em]
\small
\textsuperscript{1}Beijing Institute of Technology
\qquad
\textsuperscript{2}State Key Laboratory of General Artificial Intelligence, BIGAI
\qquad
\textsuperscript{3}MiniMax
\\[0.20em]
\small
\Letter~Corresponding authors
\qquad
Project page:
\href{https://wangzhumei.github.io/cohere4D/}
{\texttt{https://wangzhumei.github.io/cohere4D/}}
\end{tabular}%
}

%% file: figs/teaser.tex
\begin{minipage}{\textwidth}
    \vspace{-8mm}
    \centering
    \includegraphics[width=\textwidth]{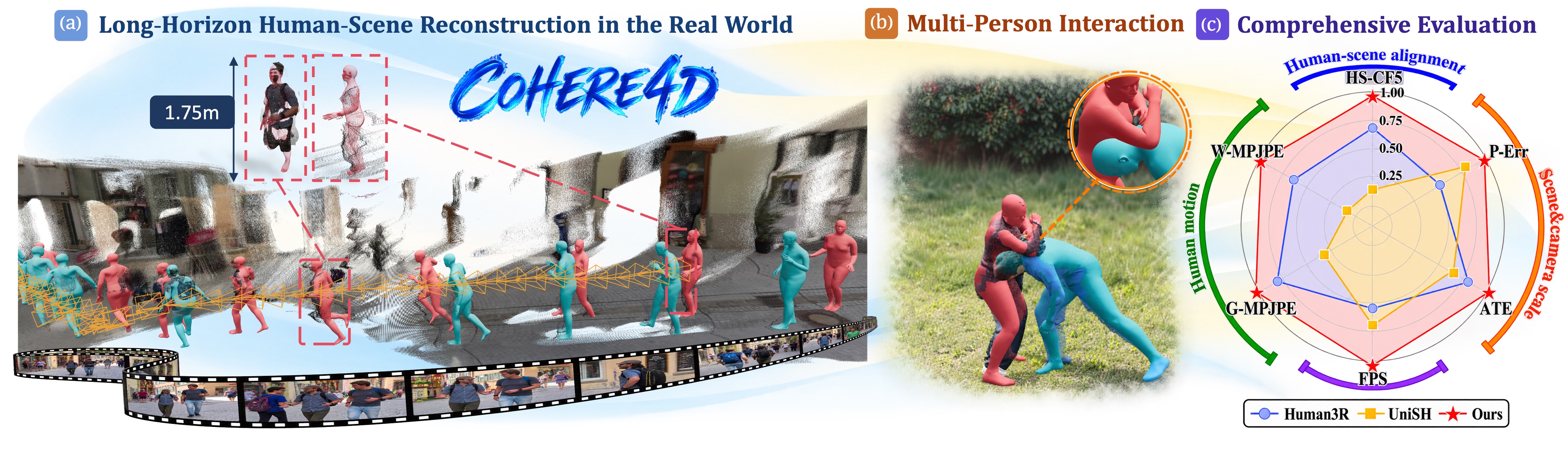}
    \vspace{-9mm}
    \captionof{figure}{
                                                                We present \textbf{\OurMethod{}}, a training-free framework for coherent 4D human--scene reconstruction.
        (a) Metric humans anchor scene scale to a unified metric world.
                (b) Scene geometry grounds humans via dense surface correspondences, yielding better alignment and more plausible interactions.
        (c) Normalized performance across human--scene alignment (3DPW), scene/camera metric scale (BEDLAM2), human motion (EMDB-2) and efficiency.
            }
    \label{fig:teaser}
\end{minipage}

\vspace{-1mm}

%% file: sec/0_abstract.tex
\vspace{-8mm}

Human and scene foundation models offer complementary strengths for monocular 4D reconstruction: human models recover metric-scale body meshes but lack scene grounding, whereas scene models estimate globally consistent 3D geometry and camera motion without reliable metric scale.
Crucially, both models reconstruct the same foreground human surface.
Yet existing methods do not directly exploit this shared surface: lacking reliable foreground human geometry in scene reconstructions, earlier approaches either resort to sparse contact constraints or learn data-driven alignment modules that generalize poorly to out-of-distribution scenes.
We present \OurMethod{}, a \textbf{training-free} framework that directly registers the human surfaces predicted by independently pretrained human and scene foundation models to reconstruct a \textbf{coherent, metric} 4D world.
\OurMethod{} establishes dense, pixel-aligned 3D correspondences over the visible human surface and employs RANSAC with cross-frame consensus to suppress outliers from occlusion, background contamination, and depth noise.
This formulation solves for two complementary geometric quantities: a sequence-level shared scale that calibrates scene geometry and camera motion to metric human size, and per-frame human translations that align camera-space human meshes with the scaled scene geometry to enforce spatial and temporal consistency.
Across diverse backbones and benchmarks, \OurMethod{} substantially outperforms prior methods in \textbf{human--scene alignment, metric scene and camera pose reconstruction, and global human motion estimation}, adding only 4\,ms/frame, enabling overall inference at up to 14 FPS.

\vspace{-6mm}

%% file: sec/1_intro.tex
\vspace{-2mm}

\label{sec:intro}
Jointly reconstructing humans, 3D scene geometry, and camera motion from a monocular video is fundamental to 4D perception.
Such a reconstruction is only physically meaningful when the human and the scene are aligned within a shared metric space, enabling geometrically consistent reasoning for applications ranging from AR/VR and gaming to robotic teleoperation, imitation learning, and embodied data collection~\citep{shi2026scene,lyu2026reviv,muller2025reconstructing}.
Although human and scene foundation models have progressed rapidly along separate tracks, neither suffices on its own: parametric human models provide strong anthropometric priors for metric body size yet lack spatial grounding in the environment, while 3D scene models recover coherent camera trajectories and dense geometry only up to an ambiguous or drifting scale.
Importantly, both models reconstruct the same foreground human in the video, providing a shared geometric surface to bridge their coordinate systems.

Despite this natural connection, existing methods do not directly register this shared human surface, as doing so requires the scene reconstruction to recover reliable foreground geometry for dynamic humans.
Before scene foundation models were available, prior methods relied on sparse SLAM point clouds that largely omit dynamic foregrounds, restricting alignment to sparse human--background contact constraints: \JOSHMethod{}~\citep{liu2026joint} enforces contact priors through costly offline optimization, whereas HAC~\citep{yang2025humans} uses contact joints to calibrate a global camera scale.
Yet such contact-based calibration relies on visible ground contact and only rescales camera motion, leaving the reconstructed human locally misaligned with the 3D scene.
As early 3D foundation models~\citep{wang2025continuous,wang2025vggt} began estimating dense foreground point clouds with coarse or distorted human geometry, unified frameworks turned to data-driven learning to bridge the gap: \HumanThreeR{}~\citep{chen2025human3r} and \SHOWMethod{}~\citep{shi2026scene} learn metric scale and human placement through joint training, while \UniSHMethod{}~\citep{li2026unish} trains a dedicated alignment network to match SMPL depth with scene point clouds.
However, real-world data that jointly annotates metric humans, dense scenes, and cameras is scarce, forcing reliance on synthetic data or noisy pseudo-labels and making learned alignment modules vulnerable to out-of-distribution shift.

Recent advances in scene foundation models such as \DAThree{}~\citep{lin2025depth}, \PiThreeX{}~\citep{wang2026pi}, and \VGGTOmega{}~\citep{wang2026vggt} have substantially improved the geometric fidelity of foreground human point clouds, making it possible to revisit training-free alignment through direct 3D surface registration.
Based on this insight, we introduce \OurMethod{}, a \textbf{training-free framework for coherent 4D human--scene reconstruction}.
Given a monocular video, we first use pretrained human and scene foundation models to independently initialize metric human meshes, dense scene geometry, and camera motion.
We then pair visible points on the projected human mesh with back-projected scene points at shared pixels.
Because these correspondences are inevitably contaminated by occlusion, background clutter, and depth noise, we formulate alignment as a robust RANSAC-based~\citep{fischler1981random} registration with cross-frame scale consensus (Eqs.~\ref{eq:scale_translation_ls}--\ref{eq:scale_translation_closed_form}) that decouples two quantities with different temporal properties: a \textbf{global scale factor} anchors the scene and camera trajectory to the metric human, while \textbf{frame-dependent translation offsets} ground the camera-space human meshes in the calibrated scene.
The resulting humans, scene geometry, and cameras are thus directly transformed into a common metric world coordinate system without any additional training.

Because this registration is entirely training-free, \OurMethod{} flexibly pairs with diverse backbones: human-anchored scale calibration corrects metric drift in \DAThree{}~\citep{lin2025depth} and \PiThreeX{}~\citep{wang2026pi} and unlocks scale-normalized \VGGTOmega{}~\citep{wang2026vggt} for metric reconstruction, while scene-grounded translation alignment refines camera-space human placement in fast multi-person models such as \MHMRtwo{}~\citep{fiche2026multi} and scales directly with stronger scene foundation models.
Furthermore, unlike contact-based calibration~\citep{yang2025humans}, which fails when the human is airborne (\eg\ jumping), when the lower body is occluded, or when truncated legs violate flat-ground assumptions, our dense surface registration operates on any visible body region without any contact assumptions.

Our contributions are summarized as follows:

\noindent-- We present \OurMethod{}, a training-free, plug-and-play framework for coherent metric 4D human--scene reconstruction that registers the shared human surfaces predicted by independently pretrained human and scene foundation models at test time.

\noindent-- We formulate a robust RANSAC and cross-frame consensus estimator over dense pixel-aligned correspondences that decouples sequence-level scene and camera scale calibration from per-frame camera-space human translation alignment.

\noindent-- \OurMethod{} achieves state-of-the-art performance in human--scene alignment (36.2\% lower HS-CF$_5$), metric scene and camera reconstruction (57.1\% lower point error), and global human motion (53.6\% lower G-MPJPE), while running at 6--14 FPS across backbones and generalizing robustly to out-of-distribution scenes and severe occlusion.

%% file: sec/2_related.tex
\input{tables/method_comparison}
\subsection{Human Mesh Recovery}
Human mesh recovery estimates camera-space 3D pose and shape from monocular images~\citep{kanazawa2018end,kolotouros2019learning,li2022cliff,goel2023humans,dwivedi2024tokenhmr,patel2025camerahmr,baradel2024multi,yang2026fast}, with state-of-the-art methods \MHMRtwo{}~\citep{fiche2026multi} and \SAMThreeDBody{}~\citep{yang2026sam} offering fast multi-person reconstruction and strong in-the-wild generalization, respectively.
World-grounded methods recover global human motion using motion priors, camera motion, SLAM, gravity, and contact constraints~\citep{yuan2022glamr,ye2023decoupling,kocabas2024pace,sun2023trace,shin2024wham,wang2024tram,wang2025prompthmr,wang2026mocap,wang2026duomo,zhao2026onlinehmr}, including \GVHMRMethod{}'s~\citep{shen2024world} gravity-view coordinates and \GEMMethod{}'s~\citep{li2025genmo} generative motion priors.
However, these methods rarely exploit dense scene geometry to constrain trajectories, and temporal rollout can accumulate drift.
We use reconstructed scene geometry to improve less reliable human trajectories and joint human--scene reconstruction.

\subsection{Foundation Models for 3D/4D Reconstruction}
Feed-forward 3D/4D reconstruction models predict depth or point maps, camera parameters, and scene geometry from images or videos~\citep{wang2024dust3r,leroy2024grounding,wang2025continuous,wang2025vggt}.
Despite recent progress, monocular reconstruction remains inherently scale-ambiguous: models either operate in a normalized space, as in \VGGTOmega{}~\citep{wang2026vggt}, or predict metric scale using learned priors~\citep{wang2025continuous,keetha2026mapanything}, including \PiThreeX{}'s metric token~\citep{wang2026pi,pi3xgithub} and \DAThree{}'s metric-depth supervision~\citep{lin2025depth}.
Absolute metric accuracy remains underexplored because common evaluation protocols emphasize relative geometry or remove scale errors through scale or Sim(3) alignment~\citep{wang2024dust3r,wang2025vggt,wang2026pi}.
We instead calibrate scene geometry using more reliable human-scale priors, improving robustness under distribution shift.

\subsection{Joint Human--Scene Reconstruction and Alignment}
Training-free methods couple humans, cameras, and scenes through geometric optimization or calibration~\citep{yang2026humanba,zhao2024synergistic}: \HSfMMethod{}~\citep{muller2025reconstructing} performs multi-view human--scene bundle adjustment, \JOSHMethod{}~\citep{liu2026joint} optimizes monocular human--scene contact constraints offline, and HAC~\citep{yang2025humans} uses contact joints to calibrate global SLAM camera scale without optimization.
Unlike these approaches, we neither require offline optimization nor rely on sparse ground-contact priors that only rescale camera motion; instead, we directly register dense visible human surfaces with scene predictions online to both calibrate scene scale and align per-frame human translations.
Feed-forward methods learn joint reconstruction: \HumanThreeR{}~\citep{chen2025human3r} extends CUT3R's metric 4D prior~\citep{wang2025continuous}, \UniSHMethod{}~\citep{li2026unish} predicts scene scale and human translations with AlignNet, and \SHOWMethod{}~\citep{shi2026scene} uses scale, placement, and geometric alignment losses.
Their metric grounding and human placement depend on joint training, which can limit generalization under distribution shift.
More recent methods incorporate human--scene constraints through model-specific fine-tuning, using either human geometry to improve scene reconstruction~\citep{kim2026coherent,xiong2025huprior3r,song2026metrichmsr} or scene geometry to refine humans~\citep{sur2026unicon3r,ym2026graft}.
These approaches remain tied to their fine-tuned human or scene models; in contrast, we align independently pretrained models without additional training, anchoring scene scale to metric human meshes while simultaneously using scene geometry to refine human translations.
Table~\ref{tab:method_comparison} compares representative methods.

%% file: tables/method_comparison.tex
\begin{table}[t]
\centering

\caption{
    Comparison of representative human--scene reconstruction methods.
}
\label{tab:method_comparison}
\vspace{-8pt}

\setlength{\tabcolsep}{2.6pt}
\renewcommand{\arraystretch}{1.08}

\resizebox{\linewidth}{!}{\begin{tabular}{@{}lccccc@{}}
\toprule

Method
& Mono.
& 4D
& \shortstack{Multi-Person}
& \shortstack{Training-Free}
& Online \\

\midrule

\HSfMMethod{}~\citep{muller2025reconstructing}
& \xmark
& \xmark
& \cmark
& \cmark
& \xmark \\

\JOSHMethod{}~\citep{liu2026joint}
& \cmark
& \cmark
& \cmark
& \cmark
& \xmark \\

HAC~\citep{yang2025humans}
& \cmark
& \cmark
& \xmark
& \cmark
& \xmark \\

\MetricHMSRMethod{}~\citep{song2026metrichmsr}
& \cmark
& \xmark
& \xmark
& \xmark
& \cmark \\

\HumanThreeR{}~\citep{chen2025human3r}
& \cmark
& \cmark
& \cmark
& \xmark
& \cmark \\

\UniSHMethod{}~\citep{li2026unish}
& \cmark
& \cmark
& \xmark
& \xmark
& \cmark \\

\CHROMMMethod{}~\citep{kim2026coherent}
& \cmark
& \cmark
& \cmark
& \xmark
& \cmark \\

\SHOWMethod{}~\citep{shi2026scene}
& \cmark
& \cmark
& \cmark
& \xmark
& \cmark \\

\midrule

\textbf{Ours}
& \cmark
& \cmark
& \cmark
& \cmark
& \cmark \\

\bottomrule
\end{tabular}}

\vspace{-16pt}
\end{table}

%% file: sec/3_method.tex
\label{sec:method}
\input{figs/pipeline}
Given a monocular RGB video
$\mathcal{I}=\{\mathbf{I}_{t}\}_{t=1}^{T}$,
where
$\mathbf{I}_{t}\in\mathbb{R}^{H_t\times W_t\times 3}$,
our goal is to reconstruct humans, scene geometry, and camera motion in a unified metric world coordinate system.
As illustrated in Figure~\ref{fig:pipeline}, we employ two independently pretrained foundation models to initialize the human and scene components.
At each frame, the human branch predicts one or more metric human meshes, while the scene branch estimates dense scene geometry and camera motion (Section~\ref{sec:preliminaries}).
Since the absolute scale predicted by scene foundation models can be inaccurate, we treat the metric human meshes as anchors and use RANSAC over multi-frame human--scene correspondences to jointly estimate a sequence-level scale and per-frame human translations; the scalar scale is used to calibrate the scene prediction to the human's metric scale (Section~\ref{sec:scene_calibration}), while the per-frame translations ground the humans in the calibrated scene (Section~\ref{sec:human_alignment}).
The calibrated humans, scene geometry, and camera trajectory are finally transformed into the same metric world coordinate system.
Both foundation models' weights remain unchanged throughout the pipeline, making our approach training-free.

\subsection{Preliminaries}
\label{sec:preliminaries}

\paragraph{Parametric Human Models.}
For person $n$ at frame $t$, a human reconstruction backbone $\mathcal{F}_{H}$ predicts pose, shape, orientation, and camera-space translation:
\begin{equation}
    \mathcal{F}_{H}(\mathcal{I})
    =
    \{
        \boldsymbol{\theta}_{t,n},
        \boldsymbol{\beta}_{t,n},
        \mathbf{Q}_{t,n},
        \boldsymbol{\gamma}_{t,n}
    \}_{t,n},
    \; n=1,\ldots,N_t
    \label{eq:human_state}
\end{equation}
where $\boldsymbol{\theta}_{t,n}$ and $\boldsymbol{\beta}_{t,n}$ are model-specific pose and shape parameters, $\mathbf{Q}_{t,n}\in\mathrm{SO}(3)$ and $\boldsymbol{\gamma}_{t,n}\in\mathbb{R}^{3}$ are the global body orientation and translation in camera coordinates.

The body decoder $\mathcal{B}$ produces translation-free and camera-space metric meshes:
\begin{equation}
    \begin{aligned}
    \bar{\mathbf{M}}_{t,n}
    &=
    \mathcal{B}
    \left(
        \boldsymbol{\theta}_{t,n},
        \boldsymbol{\beta}_{t,n},
        \mathbf{Q}_{t,n}
    \right)
    \in \mathbb{R}^{N_v\times3},
    \\
    \mathbf{M}^{c}_{t,n}
    &=
    \bar{\mathbf{M}}_{t,n}
    +
    \mathbf{1}_{N_v}\boldsymbol{\gamma}_{t,n}^{\top},
    \end{aligned}
    \label{eq:human_mesh}
\end{equation}
where $\bar{\mathbf{M}}_{t,n}$ and $\mathbf{M}^{c}_{t,n}$ are the translation-free and camera-space meshes, respectively, and $N_v$ is the model-specific vertex count.
Our framework is agnostic to the underlying parametric body representation and supports diverse human reconstruction backbones ~\citep{fiche2026multi,goel2023humans,yang2026sam,chen2025human3r}, requiring only metric human surfaces and camera-space root transformations, regardless of parameterization or mesh topology.

\vspace{-5mm}
\paragraph{Scene Foundation Models.}
Given a video or temporal subset, a scene foundation model $\mathcal{F}_{S}$ predicts camera parameters and dense geometry:
\begin{equation}
    \mathcal{F}_{S}(\mathcal{I})
    =
    \left\{
        \mathbf{K}_{t},
        \mathbf{R}_{t},
        \mathbf{t}_{t},
        \mathcal{G}_{t}
    \right\}_{t},
    \label{eq:scene_output}
\end{equation}
where $\mathbf{K}_{t}\in\mathbb{R}^{3\times3}$ contains camera intrinsics, $\mathcal{G}_{t}$ denotes dense geometry, and $(\mathbf{R}_{t},\mathbf{t}_{t})\in\mathrm{SO}(3)\times\mathbb{R}^{3}$ maps camera coordinates to the first frame of the current scene-model input.

We convert scene geometry~\citep{wang2026vggt,wang2026pi,lin2025depth} to camera-space points by back-projecting depth maps $\mathcal{G}_{t}=\mathbf{D}_{t}\in\mathbb{R}^{H_t\times W_t}$:
\begin{equation}
    \mathbf{p}^{c}_{t}(\mathbf{u})
    =
    \mathbf{D}_{t}(\mathbf{u})\,
    \mathbf{K}_{t}^{-1}\tilde{\mathbf{u}},
    \qquad
    \tilde{\mathbf{u}}=[u,v,1]^{\top}.
    \label{eq:depth_to_point}
\end{equation}
For models that predict point maps, we transform them from the scene-model's reference frame to camera-space points $\mathbf{p}^{c}_{t}(\mathbf{u})$ using the predicted camera extrinsics.

\subsection{Unified Human--Scene Alignment}
\label{sec:scene_calibration}

Although some scene foundation models can predict metric-scale geometry, their data-driven scale estimates remain sensitive to distribution shift. Using reconstructed humans as near-metric references, we estimate a shared scene scale and per-frame human translations from pixel-aligned human--scene correspondences.

\vspace{-5mm}
\paragraph{Overlap-Aware Scene Reconstruction.}
Scene foundation models typically process only a limited number of frames jointly. We therefore partition the input video into a sequence of overlapping chunks $\{\mathcal{I}^{(k)}\}_{k=1}^{K}$ along the temporal dimension, where each chunk contains at most $L$ frames and two consecutive chunks share $O$ overlapping frames.
All frames within a chunk are jointly processed by the scene foundation model, yielding chunk-local scene geometry and camera motion.

To align chunk $k$ with the previously reconstructed sequence, we use the shared frames to establish pixel-aligned 3D correspondences
$\{(\mathbf{p}^{\mathrm{ref}}_{i},
\mathbf{p}^{\mathrm{cur}}_{i})\}_{i=1}^{N}$,
where $\mathbf{p}^{\mathrm{ref}}_{i}$ and
$\mathbf{p}^{\mathrm{cur}}_{i}$ denote the same overlapping pixel observed in the aligned reference reconstruction and the current chunk, respectively.
To prevent inter-chunk scale jitter from independent chunk-wise inference, we estimate a $\mathrm{Sim}(3)$ similarity transformation (rather than $\mathrm{SE}(3)$, see Appendix~\ref{sec:scene_ablation}):
\vspace{-3mm}
\begin{equation}
    \mathbf{p}^{\mathrm{ref}}_{i}
    \simeq
    \alpha_k
    \mathbf{A}_{k}
    \mathbf{p}^{\mathrm{cur}}_{i}
    +
    \mathbf{d}_{k},
    \qquad
    \mathbf{A}_{k}\in\mathrm{SO}(3),
    \label{eq:chunk_sim3}
\end{equation}
where $\alpha_k\in\mathbb{R}_{+}$, $\mathbf{A}_k$, and $\mathbf{d}_k\in\mathbb{R}^{3}$ denote the inter-chunk scale, rotation, and translation, respectively.
We first estimate $\alpha_k$ from overlap depth ratios using a weighted median with outlier rejection based on the median absolute deviation (MAD). With $\alpha_k$ fixed, we recover $(\mathbf{A}_k,\mathbf{d}_k)$ using confidence-weighted Procrustes within iteratively reweighted least squares (IRLS)~\citep{holland1977robust}, which downweights high-residual correspondences.
The estimated transformation is applied to both scene geometry and camera motion.
By sequentially composing these transformations, we stitch all scene chunks and camera poses into the first chunk's coordinate system, yielding a temporally consistent long-sequence reconstruction in a chunk-wise online manner.

\vspace{-5mm}
\paragraph{Dense Human--Scene Surface Correspondences.}
Simultaneously with scene reconstruction, we estimate a sequence-level metric
scene scale from the first chunk using the metric human surface as a geometric reference.
For clarity, we omit the person index $n$ below and describe one human at frame $t$. For multi-person inputs, we construct correspondences separately for each valid human mesh and combine their scale evidence to estimate one shared scene scale, while estimating a separate translation for each person in each frame.
Using scene-predicted intrinsics $\mathbf{K}_t$ to express the camera-space human mesh $\mathbf{M}^{c}_{t}$, we project its triangles onto the image plane via $\Pi_{\mathbf{K}_t}$ and rasterize them with a z-buffer.
Let $\Omega_t^{H}$ denote the resulting visible human region under self-occlusion.

For each pixel $\mathbf{u}_i=(u_i,v_i)\in\Omega_t^{H}$, perspective-correct barycentric interpolation on the front-most triangle recovers a continuous 3D human surface point $\mathbf{m}_{t,i}$ on the surface of $\bar{\mathbf{M}}_t$ before global translation.
At the same pixel, we recover the corresponding camera-space scene point
$\mathbf{p}^{c}_{t,i}$ by back-projecting the predicted depth
(Eq.~\ref{eq:depth_to_point}).
As shown in Figure~\ref{fig:pipeline}, this gives the pixel-aligned 3D correspondence set
\begin{equation}
    \mathcal{C}_{t}
    =
    \left\{
        (\mathbf{p}^{c}_{t,i},\mathbf{m}_{t,i})
        \;\middle|\;
        \mathbf{u}_i\in\Omega_t^{H},
        \ D_t(\mathbf{u}_i)>0
    \right\}.
    \label{eq:human_scene_correspondence}
\end{equation}
Unlike nearest-neighbor matching in 3D, these correspondences are induced directly by the shared image pixel and preserve the visible human-surface association.

\vspace{-5mm}
\paragraph{Joint Scale and Translation Estimation.}
We formulate human--scene alignment as a least-squares problem over a shared scale $s$ and per-frame human translations $\{\boldsymbol{\gamma}_t\}$:
\begin{equation}
    \min_{s>0,\{\boldsymbol{\gamma}_t\}}
    \sum_{\substack{t,i:\;(\mathbf{p}^{c}_{t,i},\mathbf{m}_{t,i})\in\mathcal{C}_t}}
    \left\|
        s\mathbf{p}^{c}_{t,i}
        -
        \mathbf{m}_{t,i}
        -
        \boldsymbol{\gamma}_t
    \right\|_2^2 ,
    \label{eq:scale_translation_ls}
\end{equation}
where $s$ converts the scene prediction to the metric human scale and $\boldsymbol{\gamma}_t$ places the translation-free human surface in the camera coordinate system.
To robustly estimate these quantities, we generate per-frame scale hypotheses, select a shared scale through cross-frame consensus, and refine human translations with the selected scale fixed.
For a single-frame subset $\mathcal{S}\subseteq\mathcal{C}_t$, the least-squares scale--translation hypothesis is
\begin{equation}
    \begin{aligned}
    s
    &=
    \frac{
        \sum_{(\mathbf{p}^{c}_{t,i},\mathbf{m}_{t,i})\in\mathcal{S}}
        (\mathbf{p}^{c}_{t,i}-\bar{\mathbf{p}}^{c})^{\top}
        (\mathbf{m}_{t,i}-\bar{\mathbf{m}})
    }{
        \sum_{(\mathbf{p}^{c}_{t,i},\mathbf{m}_{t,i})\in\mathcal{S}}
        \|\mathbf{p}^{c}_{t,i}-\bar{\mathbf{p}}^{c}\|_2^2
    },
    \\[2pt]
    \boldsymbol{\gamma}
    &=
    s\bar{\mathbf{p}}^{c}
    -
    \bar{\mathbf{m}} ,
    \end{aligned}
    \label{eq:scale_translation_closed_form}
\end{equation}
where $\bar{\mathbf{p}}^{c}$ and $\bar{\mathbf{m}}$ are the respective means over $\mathcal{S}$.

Background geometry, occlusion boundaries, other people, and erroneous depths can contaminate the projected human region, biasing scale and translation estimates and thus human placement.
We therefore generate RANSAC hypotheses from minimal correspondence subsets and score them by inlier support using $r_{t,i}(s,\boldsymbol{\gamma})=\|s\mathbf{p}^{c}_{t,i}-\mathbf{m}_{t,i}-\boldsymbol{\gamma}\|_2\leq\tau_t$, where $\tau_t$ adapts to human body size.
We refine the best-supported hypothesis on its inliers with Huber weighting (Appendix~\ref{app:metric_calibration}).

\vspace{-5mm}
\paragraph{Cross-Frame Shared Metric Scale.}
To suppress depth-induced frame-wise scale fluctuations, we select $s^\star=s_t$ with the greatest cross-frame inlier support among valid frames in the first chunk.
We apply this shared scale $s^\star$ to scene geometry and camera translation while preserving the metric human body:
\begin{equation}
    D_t^{\mathrm{metric}} = s^\star D_t,\,
    \mathbf{p}_t^{c,\mathrm{metric}} = s^\star\mathbf{p}_t^c,\,
    \mathbf{t}_t^{\mathrm{metric}} = s^\star\mathbf{t}_t.
    \label{eq:metric_scale_propagation}
\end{equation}

\subsection{Scene-Grounded Translation Refinement}
\label{sec:human_alignment}

Because monocular human models estimate camera-space translation from human cues alone, the calibrated scene geometry provides a complementary spatial constraint for grounding the human in the surrounding 3D environment.
We therefore introduce scene-grounded translation refinement to refine the human translation after determining the shared metric scale $s^\star$.

For each frame, we reuse the pixel-aligned human--scene correspondences
$\mathcal{C}_t$ defined in Section~\ref{sec:scene_calibration} and keep
$s^\star$ fixed.
Each correspondence induces a translation estimate
$s^\star\mathbf{p}^{c}_{t,i}-\mathbf{m}_{t,i}$.
We robustly aggregate these estimates using iterative inlier filtering and coordinate-wise median refinement:
\begin{equation}
\begin{aligned}
&\hat{\boldsymbol{\gamma}}_{t}
=
\operatorname*{RobustMed}_{(\mathbf{p}^{c}_{t,i},\mathbf{m}_{t,i})
\in\mathcal{C}_t}
\left(
    s^\star\mathbf{p}^{c}_{t,i}-\mathbf{m}_{t,i}
\right),\\[3pt]
&\hat{\mathbf{M}}^{c}_{t}
=
\bar{\mathbf{M}}_{t}
+
\mathbf{1}_{N_v}\hat{\boldsymbol{\gamma}}_{t}^{\top}.
\end{aligned}
\label{eq:scene_grounded_translation}
\end{equation}
Here, $\operatorname{RobustMed}$ denotes the robust translation refitting
procedure detailed in Appendix~\ref{app:scene_grounded_alignment}.
Only the global human translation is updated to align the human with the reconstructed scene, while the predicted body pose, shape, and global orientation remain unchanged.

This refinement improves human--scene alignment across backbones and also improves global human motion for human models with less accurate camera-space translation predictions, such as \MHMRtwo~\citep{fiche2026multi}.

%% file: figs/pipeline.tex
\begin{figure*}[t]
    \vspace{-2mm}
    \centering
    \includegraphics[width=\linewidth]{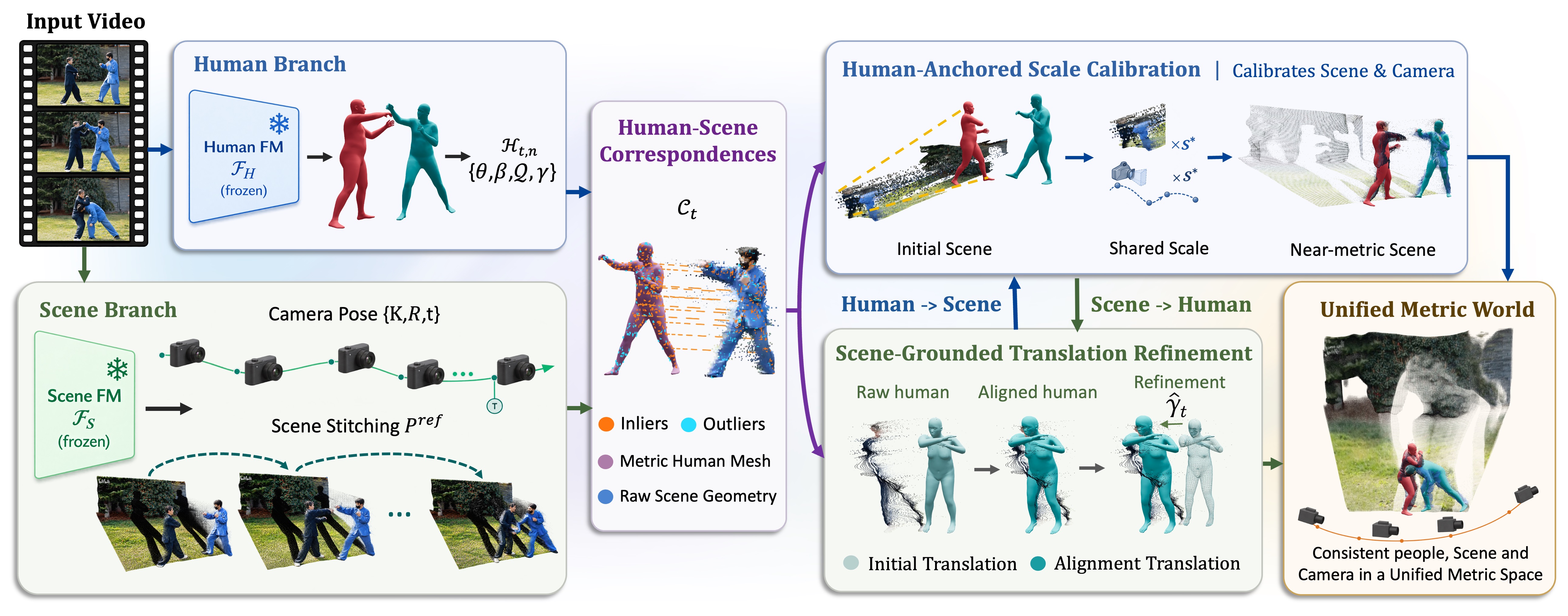}
    \vspace{-7mm}
    \caption{
                \textbf{Overview of \OurMethod{}.}
        Given monocular video frames, frozen human and scene branches independently reconstruct human meshes and scene geometry.
        Dense human--scene correspondences are established, and RANSAC selects robust inliers for geometric alignment.
        Human-anchored scale calibration uses humans as natural metric anchors to calibrate scene scale, while scene-grounded translation refinement uses the reconstructed scene to refine human translation, yielding a unified metric 4D world.
    }
    \label{fig:pipeline}
\vspace{-5mm}
\end{figure*}

%% file: sec/4_experiment.tex
We evaluate \OurMethod{} from three complementary perspectives:
human--scene geometric alignment (Section~\ref{sec:exp_hsa}),
metric scene and camera reconstruction (Section~\ref{sec:exp_metric_scene}),
and global human motion reconstruction (Section~\ref{sec:exp_global_motion}).
We provide ablation and additional discussion in
Section~\ref{sec:ablation}.

\subsection{Human--Scene Geometric Alignment}
\label{sec:exp_hsa}
\input{tables/hsa}
\input{figs/scene_scale}
\input{tables/metric_scene}

\vspace{-5mm}

We evaluate human--scene geometric alignment on 3DPW~\citep{von2018recovering} and EMDB-1~\citep{kaufmann2023emdb} in Table~\ref{tab:hsa}.
Following the metric formulation in \SHOWMethod{}~\citep{shi2026scene}, we use HS-CF$_5$ and HS-CF$_{10}$, which measure the one-way nearest-neighbor distance from the visible human surface to scene points within the human mask, normalized by the mean absolute depth deviation of the human surface.
The subscripts 5 and 10 indicate removing the outermost 5\% and 10\% of scene points along each coordinate axis, respectively, where lower values indicate better human--scene alignment.
All methods are re-evaluated under this unified protocol using the same SAM3~\citep{carion2026sam} human mask (details in Appendix~\ref{sec:hscf_protocol}).
Our method combines \MHMRtwo{}~\citep{fiche2026multi} with \DAThree{}~\citep{lin2025depth}, \PiThreeX{}~\citep{wang2026pi}, or \VGGTOmega{}~\citep{wang2026vggt} through human-anchored scale calibration (Section~\ref{sec:scene_calibration}) and scene-grounded translation refinement (Section~\ref{sec:human_alignment}).

Despite learning joint human--scene reconstruction, \HumanThreeR{}~\citep{chen2025human3r} and \UniSHMethod{}~\citep{li2026unish} exhibit visible human--scene misalignment and implausible contact on out-of-distribution inputs (Figure~\ref{fig:comparison}, first row).
Our training-free alignment consistently achieves lower HS-CF errors on both datasets and more plausible human--scene contact.
Compared with \HumanThreeR{}, \MHMRtwo{} + \VGGTOmega{} reduces HS-CF$_5$ by 36.2\% on 3DPW and 34.6\% on EMDB-1.
Applying \OurMethod{} to \HumanThreeR{} also improves consistency without backbone retraining (Appendix~\ref{app:human3r_refine}).
Stronger scene foundation models, like \PiThreeX{} and \VGGTOmega{}, generally yield lower errors: their cleaner, more complete geometry provides more reliable correspondences for scale and translation estimation.

\input{figs/comparison}

\subsection{Metric Scene Reconstruction}
\label{sec:exp_metric_scene}

To evaluate near-metric scene geometry and camera motion, we benchmark on BEDLAM2~\citep{tesch2026bedlam2} and Bonn~\citep{palazzolo2019refusion}, which provide metric ground-truth geometry and camera trajectories, as shown in Table~\ref{tab:metric_scene}.
Following standard 3D reconstruction metrics~\citep{chen2025human3r,wang2026pi,wang2026vggt}, we evaluate scene geometry with Point Error (P-Err.), Absolute Relative Error (AbsRel), and $\delta<1.25$ (percentage), and camera motion with Absolute Trajectory Error (ATE) and translational Relative Pose Error (RPE-t).
We align the reconstructed scene and camera trajectories to the ground-truth coordinate frame using a single global rotation and translation determined from the first camera pose.
Unlike scale-aligned evaluations, we apply no ground-truth scale correction, preserving each method's predicted scale and re-evaluating all baselines consistently.
BEDLAM2 (with scene metrics evaluated at ground-truth depth $< 70\,\text{m}$) has a much larger scene depth range than Bonn, making its absolute 3D point errors naturally larger for comparable relative errors.

We compare against both holistic human--scene reconstruction methods and
scene foundation models.

\vspace{-5mm}
\paragraph{Holistic human--scene reconstruction methods.}
Although \HumanThreeR{}~\citep{chen2025human3r} and
\UniSHMethod{}~\citep{li2026unish} incorporate human cues, their metric scene
accuracy can still be inferior to that of dedicated scene foundation models.
These results highlight the difficulty of generalizing learned metric-scale
prediction to out-of-distribution scenes, even when human cues are incorporated.
Meanwhile, contact-based HAC~\citep{yang2025humans} only calibrates sparse SLAM cameras, performing reasonably on flat-terrain indoor Bonn sequences but degrading markedly on the more diverse BEDLAM2 (Appendix~\ref{sec:hac}).

\vspace{-5mm}
\paragraph{Scene foundation models.}
For \DAThree{}~\citep{lin2025depth} and \PiThreeX{}~\citep{wang2026pi}, which natively predict metric-scale geometry, our human-anchored calibration consistently improves their raw metric predictions.
On Bonn, calibrating \PiThreeX{} reduces P-Err. by up to 57.1\% and ATE by 35.1\%; on BEDLAM2, it reduces AbsRel by up to 44.9\% and ATE by 37.7\%.
Figure~\ref{fig:scene_scale} further shows that our calibrated scene and camera trajectory are closer to the ground-truth metric scale than the raw predictions.
\VGGTOmega{}~\citep{wang2026vggt}, which outputs normalized, scale-ambiguous geometry, can also be calibrated into a near-metric scene and camera trajectory for world-space human reconstruction.
Overall, our method substantially improves the metric-scale accuracy of scene foundation models.

\subsection{Global Human Motion Reconstruction}

\input{tables/human_traj}

\label{sec:exp_global_motion}

We evaluate global human motion in world coordinates on BEHAVE~\citep{bhatnagar2022behave}, RICH~\citep{huang2022capturing}, and EMDB-2~\citep{kaufmann2023emdb}, which respectively provide human--object interactions with frequent occlusions, challenging in-the-wild scenes, and moving-camera sequences (Table~\ref{tab:human_traj}).
Following prior protocols~\citep{shin2024wham,shen2024world}, we report
W-MPJPE, WA-MPJPE, and Root Translation Error (RTE, \%).
W-MPJPE and WA-MPJPE align each 100-frame segment to ground truth using its first two frames and all frames, respectively.
Following recent long-sequence evaluation~\citep{qian2026masked}, we additionally report Global MPJPE (G-MPJPE), which aligns the full sequence using only its first frame, exposing long-term drift.

Table~\ref{tab:human_traj} compares our framework across backbone combinations against existing methods.
For \MHMRtwo{}~\citep{fiche2026multi}, scene-grounded translation refinement updates the human translation frame by frame,
whereas \SAMThreeDBody{}~\citep{yang2026sam,yang2026fast} combines its original camera-frame translation with the scale-calibrated scene camera trajectory to obtain the world-space human trajectory.
On BEHAVE~\citep{bhatnagar2022behave}, \MHMRtwo{}+\PiThreeX{}$^\dagger$ outperforms all baselines, showing that our scene grounding remains robust under object occlusion, while \SAMThreeDBody{} achieves even lower errors.
On RICH, \GEMMethod{}~\citep{li2025genmo} remains stronger than the \MHMRtwo{} variants on
W-MPJPE and WA-MPJPE, while the \SAMThreeDBody{} variants achieve better
results.
On the moving-camera EMDB-2 benchmark, the optimization-based \JOSHMethod{}~\citep{liu2026joint} remains competitive with the \MHMRtwo{} variants, while our \SAMThreeDBody{}+\VGGTOmega{} configuration outperforms \JOSHMethod{} across all reported metrics without costly offline optimization.
Our geometric alignment adds only about 10\,ms/frame during initialization and 4\,ms/frame in steady-state inference on an NVIDIA RTX 4090, while inference reaches 6--14 FPS across backbones (see Appendix~\ref{app:runtime} for runtime details).
Meanwhile, contact-based HAC~\citep{yang2025humans} also achieves competitive results on EMDB-2, but is less robust under non-contact motions, lower-body occlusion, or violated flat-ground assumptions, whereas our dense surface registration requires no contact priors (see Appendix~\ref{sec:hac}).

Most importantly, across two human backbones and three scene backbones, most of our configurations achieve substantially lower G-MPJPE across all three datasets than all baselines, indicating less full-sequence trajectory drift.
On EMDB-2, our best \MHMRtwo{}+\DAThree{} and \SAMThreeDBody{}+\VGGTOmega{} configurations reduce G-MPJPE by 40.7\% and 53.6\% relative to the best-performing contact-based baseline HAC~\citep{yang2025humans}, by 55.8\% and 65.5\% relative to the strongest human-only baseline \GVHMRMethod{}~\citep{shen2024world}, and by 68.4\% and 75.3\% relative to the strongest holistic baseline \HumanThreeR{}~\citep{chen2025human3r}.
The last row of Figure~\ref{fig:comparison} further shows that our human trajectories remain
consistently closer to the ground truth than the baselines over long sequences.
More qualitative results are provided in Appendices~B.2--B.3.

\subsection{Ablation and Discussion}
\label{sec:ablation}

Table~\ref{tab:traj_ablation} compares scale calibration and translation refinement for both human backbones with \PiThreeX{}~\citep{wang2026pi}; additional backbone results and analysis appear in Appendix~\ref{app:ablation_human_motion}.

We find that on the static-camera RICH~\citep{huang2022capturing} dataset,
scene geometry is sufficiently reliable for translation refinement to
consistently improve \MHMRtwo{}~\citep{fiche2026multi} (\textit{+scale\&trans}).
On the moving-camera EMDB-2~\citep{kaufmann2023emdb}, \textit{+scale} substantially improves over \textit{Raw}, showing that accurate metric camera translation directly benefits world-space human motion.
The additional gains from \textit{+scale\&trans} depend more strongly on scene quality under camera motion, and improve as stronger scene foundation models provide more reliable foreground geometry.

Unlike the fast multi-person \MHMRtwo{}~\citep{fiche2026multi}, \SAMThreeDBody{}~\citep{yang2026sam,yang2026fast} is trained on vast human-centric data and already predicts more accurate camera-space depth than current SOTA scene models. Consequently, retaining \SAMThreeDBody{}'s native camera-space translation (\textit{+scale}, as reported in Table~\ref{tab:human_traj}) preserves the best global trajectory accuracy, whereas enabling \textit{+scale\&trans} trades minor G-MPJPE (still outperforming all prior baselines) to enforce strict human--scene surface alignment when visual coherence is prioritized (see supplementary video and Appendix~\ref{app:exp_details}).
Overall, our robust, training-free alignment framework consistently improves metric scene and camera reconstruction, global human trajectories, and coherent 4D human-scene interactions.

\input{tables/traj_ablation}

%% file: tables/hsa.tex
\begin{table}[t]
    \centering
    
                                        \caption{
    \textbf{Quantitative comparison of human--scene alignment} on 3DPW and
    EMDB-1 using HS-CF$_5$ and HS-CF$_{10}$.
    All methods are evaluated under a unified protocol following the metric formulation in \SHOWMethod{}.
    Our method uses scale calibration and translation refinement to improve human--scene geometric alignment.
    \bestmark~Best result; \secondmark~Second-best result.
    }
    \label{tab:hsa}
    \vspace{-2mm}
    
    \setlength{\tabcolsep}{3.5pt}
    \renewcommand{\arraystretch}{1.0}

    \resizebox{\linewidth}{!}{    \begin{tabular}{@{}lcc|cc@{}}
        \toprule
        \multirow{2}{*}{Method}
        & \multicolumn{2}{c|}{3DPW}
        & \multicolumn{2}{c}{EMDB-1} \\

        \cmidrule(lr){2-3}
        \cmidrule(lr){4-5}

        & HS-CF$_5$ $\downarrow$
        & HS-CF$_{10}$ $\downarrow$
        & HS-CF$_5$ $\downarrow$
        & HS-CF$_{10}$ $\downarrow$ \\

        \midrule

                \HumanThreeR{}~\citep{chen2025human3r}
        & 1.079 & 1.315
        & 0.719 & 0.924 \\

        \UniSHMethod{}~\citep{li2026unish}
        & 3.676 & 4.183
        & 1.166 & 1.528 \\

        \midrule

        \multicolumn{5}{l}{\textbf{Ours}} \\

        \MHMRtwo{} + \DAThree{} ~\citep{fiche2026multi,lin2025depth}
        & 0.791 & 1.032
        & 0.571 & 0.792 \\

        \MHMRtwo{} + \PiThreeX{}~\citep{fiche2026multi,wang2026pi}
        & \cellsecond{0.700}
        & \cellsecond{0.959}
        & \cellsecond{0.532}
        & \cellsecond{0.749} \\

        \MHMRtwo{} + \VGGTOmega{}~\citep{fiche2026multi,wang2026vggt}
        & \cellfirst{\textbf{0.688}}
        & \cellfirst{\textbf{0.942}}
        & \cellfirst{\textbf{0.470}}
        & \cellfirst{\textbf{0.691}} \\

        \bottomrule
    \end{tabular}    }

    \vspace{-6mm}
\end{table}\

%% file: figs/scene_scale.tex
\begin{figure*}[t]
        \centering
    \includegraphics[width=\linewidth]{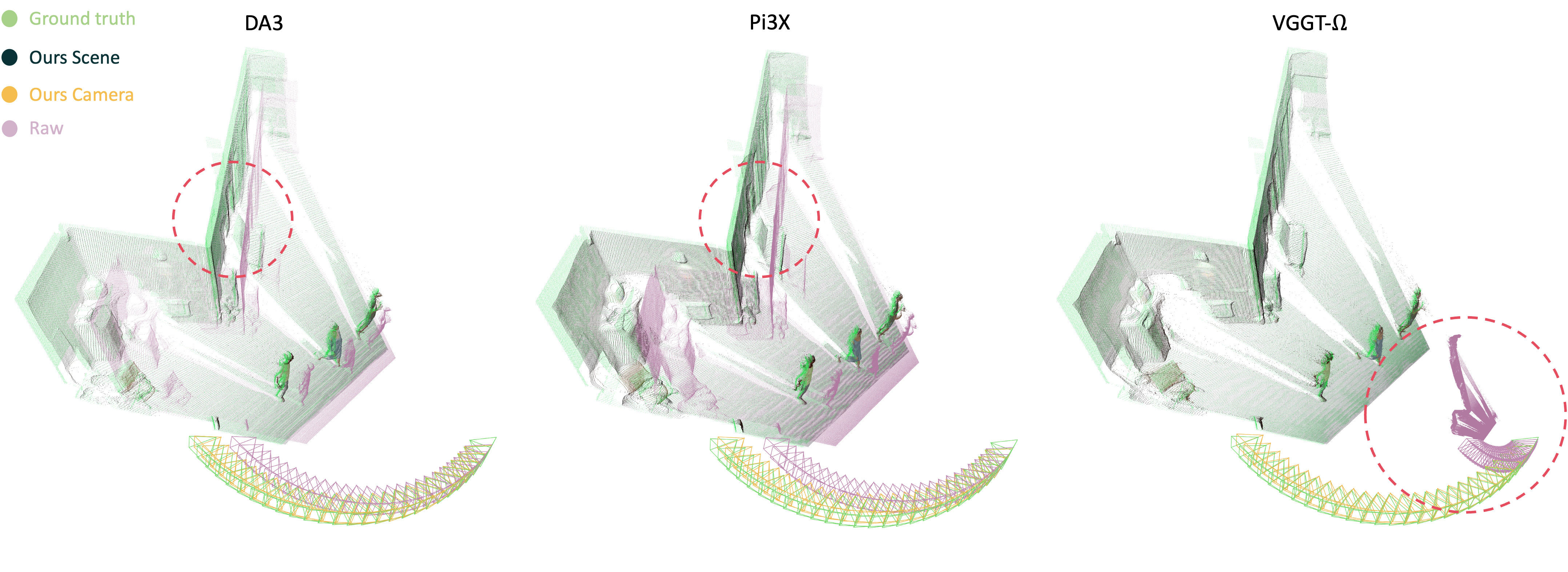}
    \vspace{-11mm}
    \caption{
                \textbf{Metric scene and camera motion on BEDLAM.}
        Our calibration brings the raw scene geometry and camera trajectory closer to the ground-truth metric scale (\VGGTOmega{}: normalized). 
    }
    \label{fig:scene_scale}
\end{figure*}

%% file: tables/metric_scene.tex
\begin{table*}[!t]
    \centering
    \vspace{-8pt}
    \caption{
        \textbf{Metric scene reconstruction and camera pose accuracy}
        on BEDLAM2 and Bonn.
        BEDLAM2 scene metrics are evaluated at GT depth
        $<70$\,m.
        Within each scene-backbone block, Raw denotes the
        native scene prediction (\VGGTOmega{}: normalized),
        while our variants calibrate scene scale using
        \MHMRtwo{} or \SAMThreeDBody{} as metric anchors.
        Point Error, ATE, and RPE-t are reported in centimeters.
        The best result within each block is boldfaced;
        \bestmark~overall best and
        \secondmark~overall second-best.
    }
    \label{tab:metric_scene}
    \vspace{-8pt}

    \setlength{\tabcolsep}{3.0pt}
    \renewcommand{\arraystretch}{1.05}

    \resizebox{\linewidth}{!}{    \begin{tabular}{@{}llccccc|ccccc@{}}
        \toprule

        \multirow{2}{*}{Scene Backbone}
        & \multirow{2}{*}{Scale Calibration}
        & \multicolumn{5}{c|}{BEDLAM2}
        & \multicolumn{5}{c}{Bonn} \\

        \cmidrule(lr){3-7}
        \cmidrule(lr){8-12}

        &
        & P-Err. $\downarrow$
        & AbsRel $\downarrow$
        & $\delta < 1.25$ $\uparrow$
        & ATE $\downarrow$
        & RPE-t $\downarrow$

        & P-Err. $\downarrow$
        & AbsRel $\downarrow$
        & $\delta < 1.25$ $\uparrow$
        & ATE $\downarrow$
        & RPE-t $\downarrow$ \\

        \midrule

        \multicolumn{2}{l}{HAC~\citep{yang2025humans}}
        & --
        & --
        & --
        & 38.67
        & 16.85
        & --
        & --
        & --
        & 5.42
        & 3.67 \\
        
        \multicolumn{2}{l}{\HumanThreeR~\citep{chen2025human3r}}
        & 345.86
        & 0.217
        & 60.20
        & 26.14
        & 11.99
        & 30.69
        & 0.077
        & 95.28
        & 28.83
        & 10.79 \\
        
        \multicolumn{2}{l}{\UniSHMethod~\citep{li2026unish}}
        & 278.37
        & 0.184
        & 67.43
        & 70.80
        & 35.13
        & 98.07
        & 0.318
        & 55.41
        & 36.76
        & 15.24 \\
        
        \midrule

        \multirow{3}{*}{\DAThree{}~\citep{lin2025depth}}
        & Raw
        & 275.82
        & 0.202
        & 57.66
        & 23.98
        & 9.98
        & 29.22
        & 0.117
        & 93.18
        & 6.04
        & 3.45 \\

        & Ours w/ \MHMRtwo{}~\citep{fiche2026multi}
        & 233.46
        & 0.157
        & 74.43
        & \textbf{16.29}
        & \textbf{6.96}
        & \textbf{19.67}
        & \textbf{0.070}
        & \textbf{97.43}
        & \textbf{5.30}
        & \textbf{3.20} \\

        & Ours w/ \SAMThreeDBody{}~\citep{yang2026sam}
        & \textbf{212.49}
        & \textbf{0.155}
        & \textbf{77.99}
        & 17.41
        & 7.42
        & 24.91
        & 0.085
        & 96.54
        & 5.81
        & 3.40 \\

        \midrule

        \multirow{3}{*}{\PiThreeX{}~\citep{wang2026pi}}
        & Raw
        & 193.41
        & 0.158
        & 70.00
        & 17.41
        & 7.66
        & 30.48
        & 0.090
        & 98.49
        & 8.70
        & 3.53 \\

        & Ours w/ \MHMRtwo{}~\citep{fiche2026multi}
        & \cellfirst{\textbf{123.18}}
        & \cellfirst{\textbf{0.087}}
        & \cellfirst{\textbf{94.02}}
        & \cellfirst{\textbf{10.85}}
        & \cellfirst{\textbf{4.97}}
        & \cellfirst{\textbf{13.08}}
        & \cellfirst{\textbf{0.036}}
        & \cellsecond{98.67}
        & 6.15
        & 3.15 \\

        & Ours w/ \SAMThreeDBody{}~\citep{yang2026sam}
        & 160.11
        & 0.133
        & 86.29
        & 15.97
        & 7.02
        & 18.04
        & 0.066
        & \cellfirst{\textbf{98.68}}
        & \textbf{5.65}
        & \textbf{3.10} \\

        \midrule

        \multirow{3}{*}{\VGGTOmega{}~\citep{wang2026vggt}}
        & Raw
        & 1017.13
        & 0.898
        & 0.01
        & 111.87
        & 44.93
        & 170.0
        & 0.572
        & 0.315
        & 22.58
        & 7.40 \\

        & Ours w/ \MHMRtwo{}~\citep{fiche2026multi}
        & \cellsecond{\textbf{136.13}}
        & \cellsecond{\textbf{0.098}}
        & \cellsecond{\textbf{92.81}}
        & \cellsecond{\textbf{10.98}}
        & \cellsecond{\textbf{5.22}}
        & \cellsecond{\textbf{16.75}}
        & \cellsecond{\textbf{0.064}}
        & \textbf{97.72}
        & \cellfirst{\textbf{3.58}}
        & \cellfirst{\textbf{2.16}} \\

        & Ours w/ \SAMThreeDBody{}~\citep{yang2026sam}
        & 143.30
        & 0.131
        & 88.11
        & 16.55
        & 7.24
        & 28.00
        & 0.097
        & 97.27
        & \cellsecond{3.64}
        & \cellsecond{2.40} \\

        \bottomrule
    \end{tabular}    }
    \vspace{-15pt}
\end{table*}

%% file: figs/comparison.tex
\begin{figure*}[t]
        \centering
    \includegraphics[width=\linewidth]{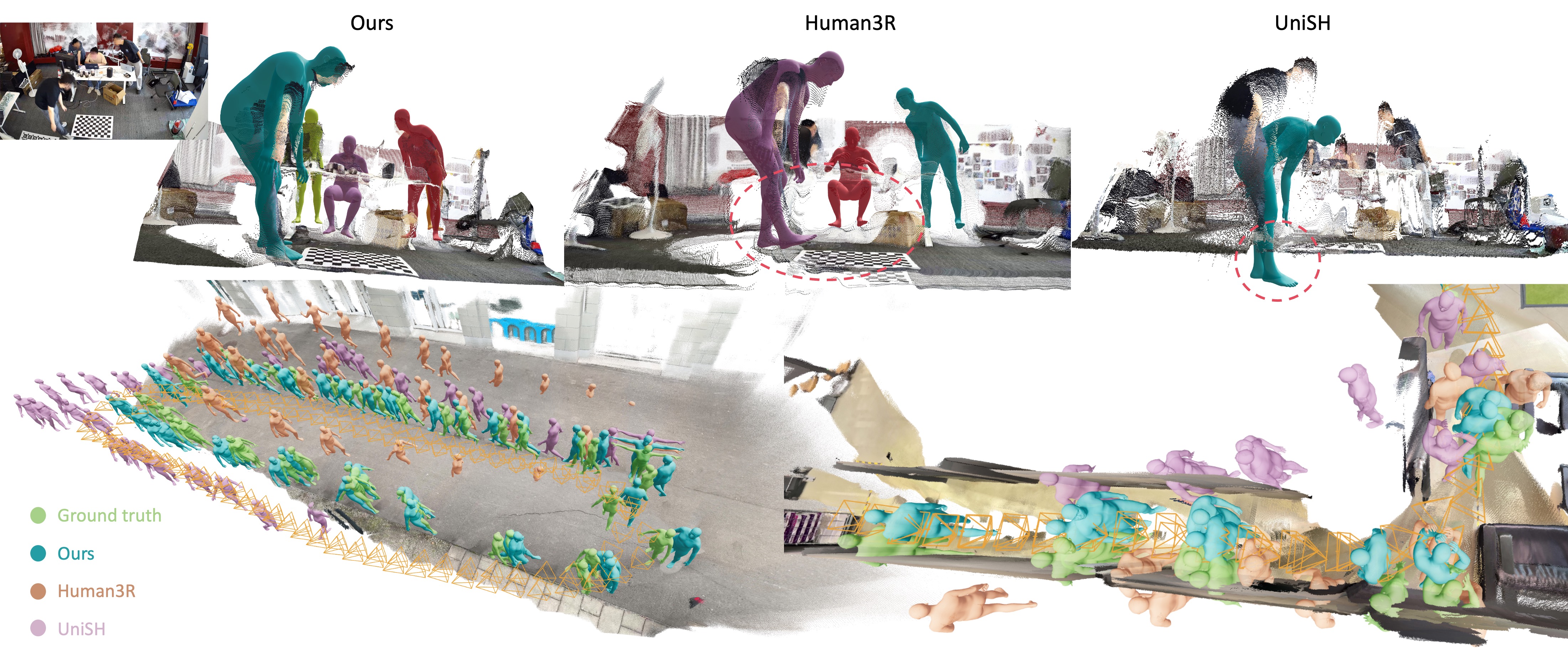}
    \vspace{-8mm}
    \caption{
                \textbf{Qualitative comparison of human--scene reconstruction under occlusion and global human motion} on Internet videos and EMDB-2.
        Top: \OurMethod{} yields more accurate human--scene alignment and contact than \HumanThreeR{} and \UniSHMethod{}.
        Bottom: \OurMethod{} exhibits less long-term global trajectory drift and better matches the ground truth than the baselines.
    }
    \label{fig:comparison}
    \vspace{-5mm}
\end{figure*}

%% file: tables/human_traj.tex
\begin{table*}[!t]
    \centering
        \caption{
        \textbf{Global human motion reconstruction in world coordinates} on BEHAVE, RICH, and EMDB-2.
        The first block contains human-only motion methods without explicit scene reconstruction, while the second contains optimization / contact-based human--scene methods, and the third contains feed-forward human--scene reconstruction methods.
        $^\dagger$ applies both scale calibration and translation refinement, while $^\ddagger$ applies scale calibration only, retaining the original \SAMThreeDBody{} camera-frame translation.
        \bestmark~overall best and \secondmark~overall second-best.
    }
    \label{tab:human_traj}
    \vspace{-3mm}

    \setlength{\tabcolsep}{1.8pt}
    \renewcommand{\arraystretch}{1.05}

    \resizebox{\linewidth}{!}{    \begin{tabular}{@{}lcccc|cccc|cccc|c@{}}
        \toprule

        \multirow{2}{*}{Method}
        & \multicolumn{4}{c|}{BEHAVE}
        & \multicolumn{4}{c|}{RICH}
        & \multicolumn{4}{c|}{EMDB-2}
        & \multirow{2}{*}{FPS $\uparrow$} \\

        \cmidrule(lr){2-5}
        \cmidrule(lr){6-9}
        \cmidrule(lr){10-13}

        & G-MPJPE $\downarrow$
        & W-MPJPE $\downarrow$
        & WA-MPJPE $\downarrow$
        & RTE $\downarrow$

        & G-MPJPE $\downarrow$
        & W-MPJPE $\downarrow$
        & WA-MPJPE $\downarrow$
        & RTE $\downarrow$

        & G-MPJPE $\downarrow$
        & W-MPJPE $\downarrow$
        & WA-MPJPE $\downarrow$
        & RTE $\downarrow$

        & \\

        \midrule

        \GVHMRMethod{}~\citep{shen2024world}
        & 509.8
        & 133.8
        & 80.0
        & 1.00

        & 277.7
        & 126.3
        & 78.8
        & 2.40

        & 3487.5
        & 276.5
        & 111.0
        & 2.00

        & \cellfirst{\textbf{30.9}} \\

        \GEMMethod{}~\citep{li2025genmo}
        & 550.6
        & 151.7
        & 90.6
        & 1.10

        & 314.8
        & 118.6
        & 75.3
        & 1.90

        & 3522.5
        & 202.1
        & 74.3
        & \cellsecond{1.20}

        & -- \\

        \cmidrule(lr){1-14}

                HAC~\citep{yang2025humans}
        & 117.8
        & 97.3
        & 64.7
        & 0.28

        & 459.2
        & 244.9
        & 158.4
        & 4.3

        & 2594.9
        & 197.3
        & 69.1
        & 1.23

        & -- \\

        \JOSHMethod{}~\citep{liu2026joint}
        & --
        & --
        & --
        & --

        & --
        & 184.3
        & 102.6
        & 3.40

        & --
        & 174.7
        & 68.9
        & 1.30

        & 0.8 \\

        \cmidrule(lr){1-14}

        \JOSHThreeR{}~\citep{liu2026joint,chen2025human3r}
        & --
        & --
        & --
        & --

        & --
        & 334.9
        & 190.4
        & 6.30

        & --
        & 661.7
        & 220.0
        & 13.10

        & \cellsecond{15.4} \\

        \MetricHMSRMethod{}~\citep{song2026metrichmsr}
        & --
        & --
        & --
        & --

        & --
        & 165.5
        & 109.6
        & 1.90

        & --
        & 199.5
        & 72.1
        & 1.40

        & -- \\

        \HumanThreeR{}~\citep{chen2025human3r}
        & 360.3
        & 112.2
        & 70.3
        & 0.73

        & 349.3
        & 184.9
        & 110.0
        & 3.30

        & 4866.0
        & 267.9
        & 112.2
        & 2.20

        & 4.9 \\

        \UniSHMethod{}~\citep{li2026unish}
        & 349.4
        & 239.9
        & 134.8
        & 1.11

        & 409.4
        & 183.2
        & 118.1
        & 4.80

        & 6099.7
        & 270.1
        & 118.5
        & 5.80

        & 6.2 \\

        \CHROMMMethod{}-mono~\citep{kim2026coherent}
        & --
        & --
        & --
        & --

        & --
        & 138.3
        & 87.5
        & 3.30

        & --
        & 255.0
        & 102.6
        & 1.70

        & 5.1 \\

        \UniConThreeR{}~\citep{sur2026unicon3r}
        & --
        & --
        & --
        & --

        & --
        & 129.8
        & 81.5
        & 1.70

        & --
        & 285.6
        & 113.7
        & 2.30

        & 2.4 \\

        \SHOWMethod{}~\citep{shi2026scene}
        & --
        & --
        & --
        & --

        & --
        & 172.7
        & 107.3
        & 2.20

        & --
        & 262.3
        & 109.1
        & 2.10

        & -- \\

        \midrule

        \multicolumn{14}{l}{\textbf{Ours}} \\

        \MHMRtwo{} + \DAThree{}~\citep{fiche2026multi,lin2025depth}$^\dagger$
        & 113.2
        & 101.6
        & 68.7
        & 0.32

        & 246.1
        & 179.4
        & 104.9
        & 2.97

        & 1539.9
        & 234.4
        & 89.9
        & 1.41

        & 6.9 \\

        \MHMRtwo{} + \PiThreeX{}~\citep{fiche2026multi,wang2026pi}$^\dagger$
        & 112.1
        & 93.8
        & 64.4
        & 0.26

        & 211.3
        & 151.3
        & 92.8
        & 2.32

        & 3313.8
        & 328.2
        & 119.7
        & 3.29

        & 6.5 \\

        \MHMRtwo{} + \VGGTOmega{}~\citep{fiche2026multi,wang2026vggt}$^\dagger$
        & 120.1
        & 99.9
        & 66.7
        & 0.28

        & 226.9
        & 167.0
        & 96.3
        & 2.70

        & 1580.1
        & 210.9
        & 78.9
        & 1.31

        & 13.8 \\

        \addlinespace[1pt]

        \SAMThreeDBody{} + \DAThree{}~\citep{yang2026sam,lin2025depth}$^\ddagger$
        & 85.3
        & 72.3
        & 46.8
        & 0.26

        & \cellsecond{142.9}
        & 100.6
        & 66.9
        & 1.51

        & \cellsecond{1411.2}
        & \cellsecond{162.4}
        & \cellsecond{68.6}
        & 1.21

        & 5.9 \\

        \SAMThreeDBody{} + \PiThreeX{}~\citep{yang2026sam,wang2026pi}$^\ddagger$
        & \cellsecond{79.7}
        & \cellfirst{\textbf{64.7}}
        & \cellfirst{\textbf{42.5}}
        & \cellfirst{\textbf{0.23}}

        & 145.0
        & \cellsecond{100.3}
        & \cellsecond{66.6}
        & \cellsecond{1.48}

        & 2303.4
        & 286.3
        & 103.9
        & 2.67

        & 5.7 \\

        \SAMThreeDBody{} + \VGGTOmega{}~\citep{yang2026sam,wang2026vggt}$^\ddagger$
        & \cellfirst{\textbf{77.6}}
        & \cellsecond{69.7}
        & \cellsecond{45.4}
        & \cellsecond{0.25}

        & \cellfirst{\textbf{138.9}}
        & \cellfirst{\textbf{99.3}}
        & \cellfirst{\textbf{66.2}}
        & \cellfirst{\textbf{1.47}}

        & \cellfirst{\textbf{1203.9}}
        & \cellfirst{\textbf{152.9}}
        & \cellfirst{\textbf{68.1}}
        & \cellfirst{\textbf{1.15}}

        & 10.3 \\

        \bottomrule
    \end{tabular}    }
\vspace{-3mm}
\end{table*}

%% file: tables/traj_ablation.tex
\begin{table}[t]
        \centering
    
    \caption{
                \textbf{Ablation} of scale calibration and translation refinement for global human motion.
        \textit{Raw} uses the original human and camera trajectories; \textit{+scale} calibrates the camera trajectory while retaining the original human translation; \textit{+scale\&trans} further applies scene-grounded human translation refinement.
    }
    \label{tab:traj_ablation}
    \vspace{-4pt}
    
    \scriptsize
    \setlength{\tabcolsep}{2.0pt}
    \renewcommand{\arraystretch}{0.86}

    \resizebox{\linewidth}{!}{    \begin{tabular}{@{}llccc@{}}
        \toprule
        Human + Scene
        & Setting
        & G-MPJPE $\downarrow$
        & W-MPJPE $\downarrow$
        & WA-MPJPE $\downarrow$ \\
        \midrule

                                \multicolumn{5}{l}{\textbf{RICH}} \\
        \midrule

        \multirow{2}{*}{\MHMRtwo{}+\PiThreeX{}~\citep{fiche2026multi,wang2026pi}}
        & +scale
        & 271.8
        & 209.2
        & 119.8 \\

        & +scale\&trans
        & \textbf{211.3}
        & \textbf{151.3}
        & \textbf{92.8} \\

        \cmidrule(lr){1-5}

                \multirow{2}{*}{\SAMThreeDBody{}+\PiThreeX{}~\citep{yang2026sam,wang2026pi}}
        & +scale
        & \textbf{145.0}
        & 100.3
        & 66.6 \\

        & +scale\&trans
        & 156.5
        & \textbf{98.0}
        & \textbf{65.5} \\

                                \midrule
        \multicolumn{5}{l}{\textbf{EMDB-2}} \\
        \midrule

        \multirow{3}{*}{\MHMRtwo{}+\PiThreeX{}~\citep{fiche2026multi,wang2026pi}}
        & Raw
        & 5877.1
        & 411.6
        & 157.2 \\

        & +scale
        & \textbf{3303.5}
        & 329.2
        & 122.5 \\

        & +scale\&trans
        & 3313.8
        & \textbf{328.2}
        & \textbf{119.7} \\

        \cmidrule(lr){1-5}

                \multirow{3}{*}{\SAMThreeDBody{}+\PiThreeX{}~\citep{yang2026sam,wang2026pi}}
        & Raw
        & 5359.4
        & 377.5
        & 144.6 \\

        & +scale
        & \textbf{2303.4}
        & \textbf{286.3}
        & \textbf{103.9} \\

        & +scale\&trans
        & 2308.8
        & 288.3
        & 105.9 \\

        \bottomrule
    \end{tabular}    }

    \vspace{-16pt}
\end{table}

%% file: sec/5_conclusion.tex
We presented \OurMethod{}, a training-free framework for coherent, metrically
aligned 4D human--scene reconstruction from monocular video.
Our framework uses metric human reconstructions to calibrate scene scale
and scene geometry to refine human translation.
Experiments demonstrate gains in human--scene alignment, metric scene
reconstruction, and global human motion across diverse settings, including
out-of-distribution inputs and severe occlusions. See Appendix~\ref{app:limitation} for limitations and failure cases.

%% file: sec/appendix_contents.tex
\startcontents[appendix]

{\LARGE\bfseries Table of Contents}
\vspace{0.4em}
\hrule
\vspace{0.8em}

\titlecontents{subsection}
    [3.0em]
    {}
    {\contentslabel{3.2em}}
    {\hspace*{-3.2em}}
    {\titlerule*[0.5pc]{.}\contentspage}

\printcontents[appendix]{}{1}{
    \setcounter{tocdepth}{2}
}

%% file: sec/6_appendix.tex
\section{Method}
\label{app:method_details}

\subsection{Unified Human--Scene Alignment}
\label{app:metric_calibration}

\paragraph{Human--scene correspondence construction.}
When more than 1,024 valid human pixels are available, we sample 1,024 pixels without replacement using a frame-dependent deterministic random seed.
At each selected pixel, the visible human surface point is recovered using z-buffering and perspective-correct barycentric interpolation.

\paragraph{RANSAC configuration.}
Since human and scene points are expressed in the same camera coordinate frame, each minimal hypothesis samples two non-degenerate human--scene correspondences and solves for scale and translation in closed form.
We evaluate one hypothesis fitted using all correspondences and 1,000 hypotheses fitted using randomly sampled minimal subsets.
The adaptive inlier threshold is
\begin{equation}
    \tau_t
    =
    \max
    \left(
        0.04~\mathrm{m},
        0.02\,\operatorname{diag}(\bar{\mathbf{M}}_t)
    \right).
\end{equation}
where $\operatorname{diag}(\bar{\mathbf{M}}_t)$ denotes the diagonal length of the 3D bounding box of the translation-free human mesh.
Candidate hypotheses are ranked by decreasing inlier count, breaking ties by increasing median and then 90th-percentile inlier residuals.
The selected hypothesis is refitted on its inliers and further refined with up to three Huber-weighted iterations~\citep{huber1992robust}, which downweight large-residual correspondences, using a transition point $\delta_t=\tau_t/2$.
Our RANSAC~\citep{fischler1981random} solver parameterizes the alignment as $\mathbf{m}_{t,i}\simeq s_t\mathbf{p}^{c}_{t,i}+\mathbf{b}_t$.
Accordingly, the human translation that aligns the translation-free body surface to the scaled scene satisfies $\mathbf{m}_{t,i}+\hat{\boldsymbol{\gamma}}_t
\simeq s_t\mathbf{p}^{c}_{t,i}$,
and is therefore given by
$\hat{\boldsymbol{\gamma}}_t=-\mathbf{b}_t$.
The offset $\mathbf{b}_t$ is thus an internal fitting variable rather than the final camera-space human translation.

\paragraph{Shared-scale selection.}
The sequence-level scale is estimated from the first scene chunk.
We discard frame-level RANSAC scale candidates with an inlier ratio below $0.25$ or an inlier median residual above $0.08$~m; if none pass, we retain all finite positive candidates.
For each retained candidate $s^{(k)}$, we fix the scale and independently re-estimate the translation offset in every valid frame as
\begin{equation}
    \mathbf{b}^{(k)}_t
    =
    \operatorname*{median}_{i}
    \left(
        \mathbf{m}_{t,i}
        -
        s^{(k)}\mathbf{p}^{c}_{t,i}
    \right).
    \label{eq:appendix_fixed_scale_offset}
\end{equation}
The corresponding inlier sets are then refined and aggregated across the first chunk.
We rank each candidate by its total cross-frame inlier support, using the median and 90th-percentile residuals as secondary criteria:
\begin{equation}
    s^\star
    =
    \operatorname*{arg\,max}_{s^{(k)}}
    \left(
        N_{\mathrm{in}}^{(k)},
        -\operatorname{Med}^{(k)},
        -\operatorname{P90}^{(k)}
    \right),
    \label{eq:appendix_shared_scale}
\end{equation}
where the tuple is compared lexicographically.
The selected $s^\star$ remains fixed for the sequence, while the frame-dependent human translation is re-estimated under this fixed scale as described in Sec.~\ref{sec:human_alignment}.

\subsection{Scene-Grounded Translation Refinement}
\label{app:scene_grounded_alignment}

After estimating the sequence-level shared scale $s^\star$, we keep it fixed for all subsequent frames and refine only the camera-space human translation.
The human pose, shape, and global orientation remain unchanged.

\paragraph{Camera-frame Translation.}
Different human backbones use different camera parameterizations before
producing the camera-frame human translation used in our alignment.
Following Sec.~\ref{sec:preliminaries}, we denote the translation of person
$n$ at frame $t$ by $\boldsymbol{\gamma}_{t,n}\in\mathbb{R}^{3}$.
To ensure pixel-aligned human--scene correspondence, we express the
camera-frame translation from all human backbones using the
scene-predicted intrinsics $\mathbf{K}_t$.

For \HMRtwoMethod{}~\citep{goel2023humans} and
\SAMThreeDBody{}~\citep{yang2026sam}, the camera head predicts crop-based
camera parameters $(s^{\mathrm{cam}},t_x,t_y)$, which we convert to a
full-perspective translation using the detected bounding box and
scene-predicted intrinsics $\mathbf{K}_t$.
Given the bounding-box center $(c_x,c_y)$, box size $b$, focal length $f$,
and principal point $(c_x^K,c_y^K)$,
\begin{equation}
    \begin{aligned}
    T_z&=\frac{2f}{s^{\mathrm{cam}}b},\\
    T_x&=t_x+\frac{2(c_x-c_x^K)}{s^{\mathrm{cam}}b},\\
    T_y&=t_y+\frac{2(c_y-c_y^K)}{s^{\mathrm{cam}}b},
    \end{aligned}
    \label{eq:crop_camera_translation}
\end{equation}
and
$\boldsymbol{\gamma}_{t,n}=(T_x,T_y,T_z)^\top$.

In contrast, \MHMRtwo{}~\citep{fiche2026multi} directly predicts a
full-perspective pelvis translation in camera coordinates.
We preserve its predicted pelvis depth $T_{z,t,n}$ and image location
$\tilde{\mathbf{u}}_{t,n}$, but re-express the translation using the
scene-predicted intrinsics $\mathbf{K}_t$:
\begin{equation}
    \boldsymbol{\gamma}_{t,n}
    =
    T_{z,t,n}\,
    \mathbf{K}_t^{-1}
    \tilde{\mathbf{u}}_{t,n},
    \label{eq:mhmr2_camera_translation}
\end{equation}
where $\tilde{\mathbf{u}}_{t,n}$ is the homogeneous pelvis image
coordinate. Thus, all human translations used by our alignment are
defined under the same camera intrinsics $\mathbf{K}_t$ as the scene
branch.

For all human backbones, $\boldsymbol{\gamma}_{t,n}$ henceforth denotes
the camera-frame human translation expressed under the scene-predicted
intrinsics $\mathbf{K}_t$.
For notational simplicity, we omit $n$ below and write
$\boldsymbol{\gamma}_t$, while the final refined translation is
$\hat{\boldsymbol{\gamma}}_t$.

\paragraph{Per-frame correspondence construction.}
For each frame $t$, we use the original camera-frame human translation to project the human mesh onto the scene image plane and construct the same pixel-aligned 3D correspondences as in Sec.~\ref{sec:scene_calibration}.
Specifically, rasterization and z-buffering determine the visible human surface, while perspective-correct barycentric interpolation recovers the continuous human surface point $\mathbf{m}_{t,i}$ at each valid pixel.
The corresponding scene point $\mathbf{p}^{c}_{t,i}$ is obtained from the scene point map at the same pixel.
At most 1,024 valid correspondences are sampled.

\paragraph{Fixed-scale robust translation refitting.}
With $s^\star$ fixed, each correspondence provides a candidate human
translation
\begin{equation}
    \boldsymbol{\gamma}^{\mathrm{cand}}_{t,i}
    =
    s^\star\mathbf{p}^{c}_{t,i}
    -
    \mathbf{m}_{t,i}.
    \label{eq:translation_candidate}
\end{equation}
We initialize the refined translation using the coordinate-wise median
\[
    \boldsymbol{\gamma}^{(0)}_t
    =
    \operatorname{median}_i
    \left(
        \boldsymbol{\gamma}^{\mathrm{cand}}_{t,i}
    \right).
\]
At iteration $k$, we compute the residual
$r_{t,i}^{(k)}
=
\|\mathbf{m}_{t,i}
+\boldsymbol{\gamma}^{(k)}_t
-s^\star\mathbf{p}^{c}_{t,i}\|_2$
and retain correspondences satisfying
$r_{t,i}^{(k)}\leq\tau_t$, where
\begin{equation}
    \tau_t
    =
    \max
    \left(
        0.04~\mathrm{m},
        0.02\,\operatorname{diag}(\bar{\mathbf{M}}_t)
    \right).
\end{equation}
The translation is then updated by taking the coordinate-wise median of $\boldsymbol{\gamma}^{\mathrm{cand}}_{t,i}$ over the retained inliers.
We perform at most three iterations and stop early when the translation change is below $10^{-7}$.
If no valid refinement is obtained, we retain the original human translation $\boldsymbol{\gamma}_t$.

\paragraph{World-space reconstruction.}
After refinement, the aligned camera-space human mesh is
$\hat{\mathbf{M}}^c_t
=
\bar{\mathbf{M}}_t
+
\mathbf{1}_{N_v}\hat{\boldsymbol{\gamma}}_t^\top$.
Using the calibrated scene camera trajectory, the human is transformed into the common world coordinate system, where the human and scene are spatially aligned and the scene geometry and camera translation share the same metric scale $s^\star$.

\section{Experiments}
\label{app:exp_details}

\subsection{Metric Scene Reconstruction on BEDLAM}

Table~\ref{tab:metric_scene_bedlam} provides additional evaluation of metric
scene geometry and camera trajectories on BEDLAM~\citep{black2023bedlam}.
Since \HumanThreeR{}~\citep{chen2025human3r} and
\UniSHMethod{}~\citep{li2026unish} are trained on BEDLAM~\citep{black2023bedlam}, we exclude them for a fair comparison and report only the raw outputs of scene foundation models and their scale-calibrated outputs using \MHMRtwo{}~\citep{fiche2026multi} as the metric human anchor. P-Err., ATE, and RPE-t are reported in centimeters, while AbsRel is unitless.
Our calibration again yields more accurate metric scale than the native metric predictions of the scene foundation models.
Although \VGGTOmega{}~\citep{wang2026vggt} outputs normalized, scale-ambiguous geometry, its stronger human-foreground reconstruction provides more reliable correspondences, leading to lower scale error after our alignment.

In addition, complementing Table~\ref{tab:metric_scene} in the main paper (evaluated at ground-truth depth $< 70\,\text{m}$), Table~\ref{tab:bedlam_bak} reports BEDLAM2~\citep{tesch2026bedlam2} scene reconstruction evaluated across all valid scene points without depth truncation.
Because BEDLAM2 contains large-scale outdoor scenes with extremely distant background regions, even small relative errors at far depths result in large absolute 3D point errors (P-Err.), yet our human-anchored scale calibration still consistently improves all scene and camera metrics across backbones.

Figure~\ref{fig:scene_scale_appendix} qualitatively compares raw scene geometry and camera motion with the results after scale calibration.
For \DAThree{}~\citep{lin2025depth} and \PiThreeX{}~\citep{wang2026pi}, the raw predictions show noticeable scale and camera-trajectory deviations from the ground truth in both indoor and outdoor scenes, whereas our scale calibration brings both closer to the correct metric scale.
For the normalized, scale-ambiguous \VGGTOmega{}~\citep{wang2026vggt} predictions, our scale calibration similarly recovers near-metric scene geometry and camera motion.

\input{figs/scene_scale_appendix}
\input{tables/metric_scene_bedlam_pair}

\wrapfill

\subsection{Additional Human Trajectory Results}
\label{app:additional_human_results}

\paragraph{Quantitative Results.} 

Table~\ref{tab:human_traj_appendix} provides additional global human motion results on BEHAVE~\citep{bhatnagar2022behave}, RICH~\citep{huang2022capturing}, and EMDB-2~\citep{kaufmann2023emdb}.
BEHAVE~\citep{bhatnagar2022behave} contains human--object interactions with frequent occlusions, making it well suited for evaluating the robustness of our alignment.
For consistency, all BEHAVE evaluations use the same subset, Sub05 with camera 2, where Sub05 provides more long sequences and camera 2 is randomly selected.
RICH~\citep{huang2022capturing} provides challenging in-the-wild scenes for testing out-of-distribution generalization, while EMDB-2~\citep{kaufmann2023emdb} evaluates performance under moving-camera conditions.

\input{tables/human_traj_appendix}
\input{figs/qualitative_appendix}

\input{figs/hsa_appendix}

We additionally evaluate \HMRtwoMethod{}~\citep{goel2023humans} with different scene foundation models, since several prior baselines, such as \GVHMRMethod{}~\citep{shen2024world}, are built on this earlier human
backbone.
Using the same \HMRtwoMethod{} backbone provides a more controlled comparison and shows that the gains of \OurMethod{} do not simply come from using stronger human reconstruction models.
Across long-sequence evaluation, our variants generally outperform prior baselines in G-MPJPE, while all \HMRtwoMethod{}-based variants outperform \GVHMRMethod{}, with trends consistent with those observed using \MHMRtwo{}~\citep{fiche2026multi}.
These results show that \OurMethod{}'s long-term trajectory improvements extend to the earlier \HMRtwoMethod{} backbone.
Moreover, as a model-independent, training-free framework, \OurMethod{} can be readily combined with newer human and scene foundation models.
Unlike unified baselines that require costly model retraining to incorporate new backbones, our framework can directly replace the corresponding pretrained modules, allowing it to benefit from future advances in human and scene foundation models with minimal additional cost.

\paragraph{Qualitative Results.} Figure~\ref{fig:qualitative_appendix} shows qualitative human--scene reconstruction results on 3DPW~\citep{von2018recovering} and EMDB-2~\citep{kaufmann2023emdb}. Our framework reconstructs multiple people together with the scene in a shared metric space within a single inference pipeline. Over long horizons, the recovered human trajectories remain consistently aligned with the reconstructed scene.

\subsection{Qualitative Comparison with Baselines}
\label{app:comp_with_baselines}

\input{figs/hsa_appendix_behave}

\paragraph{Human--Scene Alignment and Interaction.} Figure~\ref{fig:hsa_appendix} and Figure~\ref{fig:hsa_behave_appendix} qualitatively compare human--scene alignment on EMDB-2~\citep{kaufmann2023emdb} and BEHAVE~\citep{bhatnagar2022behave}. On EMDB-2~\citep{kaufmann2023emdb} (Figure~\ref{fig:hsa_appendix}), our improved human--scene alignment leads to more plausible object contact, e.g., the reconstructed humans are correctly supported by benches and other supporting surfaces. In contrast, inaccurate human scale in \HumanThreeR{}~\citep{chen2025human3r} can cause the body to intersect objects or be vertically misaligned with the supporting surface, while \UniSHMethod{}~\citep{li2026unish} shows residual human--scene misalignment that shifts the contact location. Under moving cameras, our reconstructed scene also remains more consistent across frames, whereas \HumanThreeR{}~\citep{chen2025human3r} and \UniSHMethod{}~\citep{li2026unish} may reconstruct the same object at inconsistent locations, producing visible ghosting artifacts. Similar behavior is observed on BEHAVE~\citep{bhatnagar2022behave} (Figure~\ref{fig:hsa_behave_appendix}), where our meshes better align with and interact with the objects. 
This also highlights a complementary benefit of scene-grounded translation refinement: beyond improving global trajectories when human translation is less reliable, it can improve local human--scene consistency in interaction-heavy scenes.
Even when a strong baseline sometimes shows limited gains in global trajectory metrics, these metrics may not fully capture the benefits of translation refinement.
In our in-the-wild videos, enabling translation refinement consistently yields visibly better human--scene alignment and more plausible human--scene contact, which is particularly beneficial for interaction-heavy scenes.
Overall, the improved human--scene alignment yields more geometrically consistent human--object interactions.

\paragraph{Long-Horizon Global Human Motion.}
Figure~\ref{fig:human_motion} qualitatively compares global human motion on EMDB-2~\citep{kaufmann2023emdb}.
For all methods, the reconstructed scene and human trajectory are aligned to the ground truth only at the first frame.
Over long sequences, our trajectories remain consistently closer to the ground truth with substantially less accumulated drift, consistent with the lower G-MPJPE in Table~\ref{tab:human_traj} in the main paper, while preserving better alignment with the reconstructed scene.
The last row further compares the estimated camera trajectories.
Our calibrated camera motion remains closer to the ground truth over time,
which in turn leads to more accurate world-space human trajectories under moving cameras.

\clearpage
\input{figs/human_motion}
\clearpage

\input{tables/hsa_physical_pair}

\subsection{Quantitative Physical Grounding Analysis}
\label{app:physical_grounding}

Before analyzing physical grounding, Table~\ref{tab:hsa_3db} presents the human--scene geometric alignment results using the \SAMThreeDBody{}~\citep{yang2026sam} backbone, where \OurMethod{} consistently achieves the lowest HS-CF errors on both 3DPW~\citep{von2018recovering} and EMDB-1~\citep{kaufmann2023emdb}.
Building upon this robust geometric alignment, we further evaluate physical grounding between the reconstructed human body and static supporting scene geometry on EMDB-2~\citep{kaufmann2023emdb} to quantitatively substantiate the plausible interactions observed in Figures~\ref{fig:hsa_appendix} and~\ref{fig:hsa_behave_appendix}.
Because monocular scene reconstructions provide noisy point clouds rather than watertight meshes, we adapt established physical grounding formulations from AvatarJLM~\citep{zheng2023realistic} and PhySIC~\citep{yalandur2025physic} to predicted static scene point clouds (excluding dynamic human regions via segmentation masks).

\paragraph{Ground Penetration (GP), Ground Floating (GF), and Ground Distance (GD).}
Following AvatarJLM~\citep{zheng2023realistic}, we evaluate foot--ground physical consistency. In each frame, we robustly fit a ground plane $\mathbf{n}^\top \mathbf{x} + d = 0$ ($\|\mathbf{n}\|_2 = 1$, oriented upward) to the static scene point cloud via RANSAC with planarity and footprint checks. Let $h_{t,k} = \min_{v \in \mathcal{V}_{\text{sole}}^{(k)}} (\mathbf{n}^\top v + d)$ denote the signed distance from the lowest sole vertex of foot $k \in \{L, R\}$ to the fitted ground plane at frame $t$. GP measures the average downward floor penetration across valid ground frames using $\text{GP}_t = \max(-\min_{k} h_{t,k}, 0)$. To penalize upward foot floating without falsely penalizing swing feet during walking, we identify stance feet from the lower height envelope of ground-truth motion and compute GF over valid stance-foot instances as $\text{GF}_{t,k} = \max(h_{t,k} - \tau_{\text{float}}, 0)$ with a $\tau_{\text{float}} = 2\text{ cm}$ tolerance. Following~\citep{zheng2023realistic}, GD measures the overall foot--ground contact error as the sum of penetration and floating:
\begin{equation}
    \text{GD} = \text{GP} + \text{GF}.
\end{equation}

\paragraph{Normal-Based Penetration (NBP).}
Evaluating human--object floating requires ground-truth 3D human--object contact annotations, which are unavailable on in-the-wild datasets such as EMDB-2~\citep{kaufmann2023emdb}. However, physical interpenetration between the body and surrounding objects (e.g., benches, chairs, walls) can be geometrically evaluated without contact labels. Inspired by the signed normal distance formulation in PhySIC~\citep{yalandur2025physic}, for each non-hand body vertex $v$, we query its local $K$-nearest static scene points to fit a reliable local tangent plane with center $\mathbf{c}_v$ and outward normal $\mathbf{n}_v$ oriented toward the camera. Using the signed distance $s_v = (v - \mathbf{c}_v)^\top \mathbf{n}_v$, NBP computes the average penetration depth over all valid near-surface vertices $\mathcal{V}$ (with a $\tau_{\text{pen}} = 5\text{ mm}$ tolerance):
\begin{equation}
    \text{NBP} = \frac{1}{|\mathcal{V}|} \sum_{v \in \mathcal{V}} \max(-s_v - \tau_{\text{pen}},\; 0).
\end{equation}

Table~\ref{tab:physical_grounding} reports the quantitative results on EMDB-2~\citep{kaufmann2023emdb}. GP and GF reveal two opposite failure modes in the baselines: inaccurate scale in \HumanThreeR{}~\citep{chen2025human3r} causes the reconstructed humans to sink deeply beneath the floor, which artificially suppresses its GF since buried feet yield zero upward floating error, whereas \UniSHMethod{}~\citep{li2026unish} suffers from severe upward foot floating. In contrast, \OurMethod{} maintains near-zero floating ($\text{GF} = 0.797\text{ cm}$) while achieving the lowest GP, thereby reducing overall GD by $60.6\%$ compared to \HumanThreeR{}~\citep{chen2025human3r} and by $66.4\%$ compared to \UniSHMethod{}~\citep{li2026unish}. Furthermore, our method achieves the lowest scene and object interpenetration in terms of NBP. 
Combined with the HS-CF evaluation in Table~\ref{tab:hsa_3db}, these results suggest that direct dense surface alignment, without explicit contact or physical optimization, not only improves human--scene geometric alignment but is also associated with reduced penetration and floating errors, indicating better physical grounding.

\subsection{Applying \OurMethod{} to Existing Baselines}
\label{app:human3r_refine}
\input{tables/human3r_pair}
\input{figs/ablation_hsa_baseline}

We further evaluate whether \OurMethod{} can be applied to existing unified human--scene reconstruction models without retraining, and analyze how the quality of the reconstructed scene affects scene-grounded human refinement.

\paragraph{Human--scene geometric alignment.}
Table~\ref{tab:human3r_hsa_ablation} compares the original \HumanThreeR{}~\citep{chen2025human3r} and \UniSHMethod{}~\citep{li2026unish} predictions with their corresponding results after applying \OurMethod{}.
Our human-anchored scale calibration brings the scene to the human metric scale, followed by scene-grounded translation refinement.
Applying this training-free geometric alignment consistently reduces HS-CF$_5$ and HS-CF$_{10}$ for both methods on 3DPW~\citep{von2018recovering} and EMDB-1~\citep{kaufmann2023emdb}.
Figure~\ref{fig:ablation_hsa_baseline} further illustrates these gains qualitatively: the original predictions exhibit noticeable mesh--scene misalignment, whereas our method brings the human mesh surfaces into tighter agreement with the reconstructed scene.
These results show that \OurMethod{} can be directly applied to existing human--scene reconstruction pipelines to improve geometric consistency without backbone retraining.

\paragraph{Global human motion and scene quality.}
Table~\ref{tab:human3r_motion} further evaluates global human motion on BEHAVE~\citep{bhatnagar2022behave} and EMDB-2~\citep{kaufmann2023emdb} when plugging \OurMethod{} into \HumanThreeR{} and \UniSHMethod{}, alongside an ablation using \MHMRtwo{} with the earlier Pi3~\citep{wang2026pi} scene model.

Applying \OurMethod{} improves all three global motion metrics for \HumanThreeR{} on the moving-camera EMDB-2~\citep{kaufmann2023emdb} dataset, while substantially reducing all motion errors for \UniSHMethod{} on BEHAVE and lowering its G-MPJPE on EMDB-2~\citep{kaufmann2023emdb}. At the same time, because these baselines build upon earlier scene foundation models, such motion gains are not always uniform across all settings. This primarily stems from the coarse or distorted geometry that earlier scene reconstruction models often generate around dynamic humans, which makes direct surface correspondences less reliable for trajectory refinement. Our additional comparison using the earlier Pi3~\citep{wang2026pi} scene model shows a similar dependence on scene quality: \textit{+Scale+Trans} improves all three motion metrics on BEHAVE, while on EMDB-2 it slightly increases G-MPJPE but reduces W-MPJPE and WA-MPJPE. Consistent with the progression discussed in the main paper, this observation explains why earlier unified approaches relied on end-to-end learning to compensate for noisy scene geometry rather than performing direct dense surface registration.
As modern scene foundation models (such as \PiThreeX{}~\citep{wang2026pi} and \VGGTOmega{}~\citep{wang2026vggt}) provide substantially cleaner foreground geometry, the advantages of our direct dense surface correspondence become fully pronounced for both alignment and motion refinement.
Crucially, whereas learning-based pipelines are tightly coupled to specific backbones and require \textbf{costly adaptation and fine-tuning} whenever stronger models emerge, our training-free framework can be directly integrated with new human and scene foundation models without additional training or adaptation.

\subsection{Ablation of Scale Calibration and Translation Refinement}
\label{app:ablation_human_motion}

Table~\ref{tab:human_traj_ablation} provides a comprehensive ablation of \MHMRtwo{}~\citep{fiche2026multi}, \HMRtwoMethod{}~\citep{goel2023humans}, and \SAMThreeDBody{}~\citep{yang2026sam} with \DAThree{}~\citep{lin2025depth}, \PiThreeX{}~\citep{wang2026pi}, and \VGGTOmega{}~\citep{wang2026vggt} on BEHAVE~\citep{bhatnagar2022behave}, RICH~\citep{huang2022capturing}, and EMDB-2~\citep{kaufmann2023emdb}. 
\textit{Raw} retains the original human and camera motion; \textit{+scale} preserves the camera-frame human translation while calibrating the scene scale and camera translation; and \textit{+scale\&trans} additionally refines the human translation using the reconstructed scene.

\input{tables/human_traj_ablation}
\input{tables/least_squares_ablation}

On the static-camera BEHAVE and RICH datasets, translation refinement consistently improves \MHMRtwo{} and \HMRtwoMethod{} across all scene backbones, showing that reliable scene geometry can effectively ground their less accurate global trajectory.
On the moving-camera EMDB-2~\citep{kaufmann2023emdb} dataset, \textit{+scale} substantially improves over \textit{Raw}, demonstrating that more accurate metric camera translation is critical for world-space human trajectory recovery.
The additional gains from translation refinement are less consistent on EMDB-2~\citep{kaufmann2023emdb}, where camera motion can degrade the scene foundation model's depth predictions in foreground human regions, making frame-wise translation refinement less stable than in static-camera scenes.
Across different scene backbones, \VGGTOmega{}~\citep{wang2026vggt} generally provides the most reliable scene geometry and camera poses.
Although its native output is scale-normalized and cannot be directly used in a metric world, our calibration maps it into a near-metric space without retraining, allowing its stronger geometry to directly benefit world-space human motion reconstruction.

\MHMRtwo{}~\citep{fiche2026multi} provides efficient multi-person reconstruction but has less reliable camera-space translation, making it particularly amenable to scene-grounded refinement. In contrast, \SAMThreeDBody{}~\citep{yang2026sam,yang2026fast}, trained with extensive human-centric data, already provides highly accurate camera-space depth. For this stronger human backbone, retaining its native translation while using the human as a metric anchor (\textit{+scale}) yields the best standalone global trajectory accuracy. Enabling full bidirectional grounding (\textit{+scale\&trans}) introduces a small trade-off in trajectory error, but still outperforms prior baselines in G-MPJPE while enforcing tighter human--scene alignment and improving spatial and temporal consistency between the reconstructed human and scene (see Appendix~\ref{app:comp_with_baselines} and the supplementary video). Thus, our framework flexibly adapts to different backbones: scene geometry refines human motion when human depth is less reliable, whereas a strong human model serves as a metric anchor for the scene with optional translation refinement for superior local coherence.

\input{figs/ransac_ablation}

\input{figs/HAMC_ablation}

\subsection{Ablation of Robust Metric Calibration}

We ablate the robust estimator used in our metric calibration by replacing our default RANSAC-based~\citep{fischler1981random} consensus with ordinary least squares and two robust alternatives. All settings fix \MHMRtwo{}~\citep{fiche2026multi} and \VGGTOmega{}~\citep{wang2026vggt} as the human and scene backbones.
As a baseline, ordinary least squares (Ordinary LS) treats all correspondences equally and is highly sensitive to outliers. To mitigate this, we additionally compare against trimmed least squares (Trimmed LS)~\citep{rousseeuw2003robust}, which explicitly discards a fixed ratio of the correspondences with the highest errors, and confidence-weighted iteratively reweighted least squares (Conf-weighted IRLS)~\citep{huber1992robust}, which continuously down-weights correspondences based on both fitting residuals and the scene backbone's geometric confidence.

As shown in Table~\ref{tab:least_squares_ablation}, Ordinary LS substantially degrades both human--scene geometric alignment and global human motion.
Figure~\ref{fig:ransac_ablation} illustrates this fundamental challenge by directly comparing Ordinary LS against our method.
When a human overlaps with an interacting object, ordinary least-squares fitting is easily biased toward the nearby background geometry.
While Trimmed LS and Conf-weighted IRLS improve upon this non-robust baseline by mitigating severe outliers, Table~\ref{tab:least_squares_ablation} shows they still fall short of our RANSAC-based approach.
Under occlusion or moving-camera settings, the reconstructed scene geometry can be highly unreliable. In such cases, fixed-ratio trimming (Trimmed LS) may inadvertently discard valid points or retain severe outliers when the error distribution is heavily skewed.
Similarly, Conf-weighted IRLS relies on the scene backbone's predicted confidence, which does not always perfectly correlate with true geometric accuracy.
In contrast, by isolating the maximum discrete inlier consensus across frames, our RANSAC estimator successfully rejects these challenging mismatches and reliably recovers the geometrically consistent human--scene inliers.

\input{figs/SGTR_ablation}

\subsection{Qualitative Ablation of Scale Calibration and Translation Refinement}

Figures~\ref{fig:hamc_ablation} and~\ref{fig:sgtrablation} qualitatively illustrate the effects of scale calibration and translation refinement on BEHAVE~\citep{bhatnagar2022behave} (which provides per-frame Kinect-scanned ground-truth scene geometry) as well as internet images. 

\paragraph{Effect of Scale Calibration.} Figure~\ref{fig:hamc_ablation} visualizes metric scene calibration. The left column shows the raw normalized output of \VGGTOmega{}, whose scale deviates substantially from the ground truth shown in green. After scale calibration (middle), the reconstructed scene is calibrated to a near-metric scale. The right column overlays the raw prediction, our calibrated result, and the ground truth in the same coordinate system, clearly showing the scale correction introduced by our method. 

\paragraph{Effect of Translation Refinement.} Figure~\ref{fig:sgtrablation} visualizes scene-grounded human translation refinement using \MHMRtwo{}~\citep{fiche2026multi}. The raw human prediction (left) exhibits a visible gap from the reconstructed scene, whereas translation refinement (middle) updates the human translation to better align the mesh with the scene geometry. 
The right column directly compares the predictions before and after refinement, with the lighter mesh denoting the raw result.
In particular, in the in-the-wild image shown in the last row, the raw predictions fail to establish proper contact between the two people, whereas our refinement brings them into plausible contact while remaining aligned with the reconstructed scene.
Overall, translation refinement more accurately grounds the human meshes in the scene and improves geometric consistency in both single- and multi-person interactions.

\subsection{Contact-Based vs. Dense Surface Calibration}
\label{sec:hac}
Contact-based calibration provides an effective and efficient metric-scale cue when reliable human--scene contacts are available. HAC~\citep{yang2025humans} introduces an insightful paradigm that leverages contact-point ratios to calibrate scene and camera scale. To isolate the effect of the geometric correspondence formulation from backbone differences, we conduct a controlled ablation within the same pipeline using \SAMThreeDBody{}~\citep{yang2026sam} and \VGGTOmega{}~\citep{wang2026vggt}, directly comparing contact-based scale calibration~\cite{yang2025humans} against our dense surface correspondence calibration.

\input{tables/hac_comp_3db}
\input{figs/hac}

\paragraph{Quantitative Results.}
Table~\ref{tab:hac_3db_comp} summarizes the quantitative comparison across different evaluation benchmarks.
On the EMDB-2~\citep{kaufmann2023emdb} dataset, where the full human body and flat supporting ground planes are consistently visible throughout the sequences, contact-based calibration performs competitively, while our dense surface calibration still achieves lower long-sequence trajectory error in terms of G-MPJPE.
However, on more challenging datasets characterized by diverse terrain, frequent partial occlusions, severe body truncation, and complex camera motion (e.g., 3DPW~\citep{von2018recovering} and BEDLAM2~\citep{tesch2026bedlam2}), our dense surface calibration demonstrates significantly stronger robustness and consistency across both human--scene alignment and metric scene reconstruction.

\paragraph{Qualitative Results.}
Nevertheless, contact-based calibration depends on accurate contact detection and reliable scene-ray intersections. 
These cues could easily degrade during non-contact motion, lower-body occlusion or truncation, and on sloped or uneven terrain. 
Figure~\ref{fig:hac} illustrates a representative example from 3DPW~\citep{von2018recovering}, where subjects walk over non-flat ground with truncated lower bodies.
Because the feet are out of view, the estimated mesh contact points drift, while the corresponding scene rays frequently intersect nearby geometry, resulting in biased scale estimates. When such truncation persists across a video, these errors accumulate into noticeable global scale drift.
In contrast, our dense surface formulation leverages geometric cues from any visible body region without explicit contact assumptions, maintaining robust metric scale calibration under severe truncation and non-flat terrain.

\input{tables/scene_ablation}

\input{tables/runtime}

\subsection{Scene Reconstruction Ablation}
\label{sec:scene_ablation}

We ablate two key design choices in our streaming scene
reconstruction pipeline: inter-chunk alignment and
sequence-level metric scale selection.
Results are summarized in Table~\ref{tab:scene_ablation}.

\paragraph{Scene chunk stitching.}
Although metric scene foundation models such as
\PiThreeX{}~\citep{wang2026pi} predict metric-scale geometry,
independent chunk-wise inference can still introduce scale
inconsistencies across overlapping chunks.
SE(3) stitching preserves these discrepancies, potentially
causing scale jitter in long sequences.
We therefore adopt Sim(3) stitching to align chunk scales
before applying the shared human-anchored metric calibration.
Sim(3) improves 3D scene geometry (P-Err. and AbsRel) and global human motion reconstruction, with only a minor trade-off in camera trajectory error on Bonn~\citep{palazzolo2019refusion}.
Moreover, it mitigates inter-chunk scale inconsistencies,
yielding more coherent long-sequence reconstruction.
We therefore adopt Sim(3) stitching in our streaming pipeline.

\paragraph{Shared-scale selection.}
Our streaming pipeline estimates a shared metric scale from
the first chunk with reliable human--scene correspondences.
If humans are absent or heavily occluded, scale initialization
can be deferred until valid observations become available.
We further compare this strategy with estimating the shared
scale using the entire sequence.
Both strategies achieve comparable scene and camera accuracy,
while sequence-wide estimation provides modest improvements
in global human motion.
However, sequence-wide estimation is computationally expensive
and requires offline processing.
We therefore adopt first-valid-chunk scale selection
to preserve efficient streaming reconstruction.

\subsection{Runtime Evaluation}
\label{app:runtime}

We benchmark our method on a single NVIDIA RTX 4090 and re-evaluate \HumanThreeR{}~\citep{chen2025human3r} and \UniSHMethod{}~\citep{li2026unish} on the same compute node used for our experiments, ensuring consistent hardware conditions.

As shown in Table~\ref{tab:runtime}, we report both end-to-end and core method inference FPS.
End-to-end FPS includes image preprocessing, model inference, decoding, and subsequent post-processing, and is measured on BEHAVE~\citep{bhatnagar2022behave} videos at their original resolution of $2048\times1536$; it is therefore affected by input resolution.
Core method FPS excludes only image preprocessing while retaining decoding and post-processing.
For our streaming pipeline, we additionally report initial and steady-state
FPS, where the initial stage includes the one-time shared-scale initialization, while the steady-state FPS measures subsequent streaming inference after initialization.

\MHMRtwo{}~\citep{fiche2026multi} provides the faster configuration, achieving 6.5--13.8 FPS in steady-state core inference, with a human inference cost of about 42\,ms/frame.

For the higher-accuracy configuration, we use \FastSAMThreeDBody{}~\citep{yang2026fast}, an accelerated variant of \SAMThreeDBody{}~\citep{yang2026sam} that incurs only a small accuracy loss while substantially improving efficiency.
All \SAMThreeDBody{}-based results in this paper use \FastSAMThreeDBody{}, achieving 5.7--10.3 FPS with about 60\,ms/frame for human inference.

Runtime also depends on the scene backbone.
\VGGTOmega{}~\citep{wang2026vggt} is consistently the most efficient, followed by \DAThree{}~\citep{lin2025depth}, while \PiThreeX{}~\citep{wang2026pi} is the most computationally expensive.
By comparison, our geometric alignment adds only about 10\,ms/frame during initialization and 4\,ms/frame in steady-state inference, showing that the overall runtime is dominated by the pretrained human and scene backbones.
Future advances in more efficient human and scene foundation models can therefore directly improve the overall throughput of our pipeline.

\input{figs/HSCF}

\subsection{HS-CF Evaluation Protocol}
\label{sec:hscf_protocol}

In the main paper (Table~\ref{tab:hsa}), we evaluate human--scene geometric alignment using the HS-CF metric proposed by \SHOWMethod{}~\citep{shi2026scene}, strictly following the definition in their paper. Following the recent release of the official \SHOWMethod{}~\citep{shi2026scene} evaluation code, we identified a critical discrepancy between their paper's definition and the released code regarding the query direction of the nearest-neighbor (NN) search.

Specifically, omitting percentile filtering for clarity, the published metric is
\begin{equation}
\mathrm{HS\text{-}CF}_{\mathrm{paper}}
=
\frac{
\sum_{\mathbf h\in\mathcal H}
\min_{\mathbf p\in\mathcal S}
\|\mathbf h-\mathbf p\|_2
}{
\sum_{\mathbf h\in\mathcal H}
|z_{\mathbf h}-\bar z_{\mathcal H}|
},
\end{equation}
where each visible human point queries its nearest scene point (Human$\rightarrow$Scene).

In contrast, the released evaluator computes
\begin{equation}
\mathrm{HS\text{-}CF}_{\mathrm{code}}
=
100
\frac{
\frac{1}{|\mathcal S|}
\sum_{\mathbf p\in\mathcal S}
\min_{\mathbf h\in\mathcal H}
\|\mathbf p-\mathbf h\|_2
}{
\frac{1}{|\mathcal S|}
\sum_{\mathbf p\in\mathcal S}z_{\mathbf p}
},
\end{equation}
where each scene point queries its nearest human point (Scene$\rightarrow$Human).
Thus, the released implementation differs from the published Eq.~19 in both correspondence direction and normalization, and the absolute values from the two protocols are not directly comparable.

\input{tables/HSCF}

\paragraph{Correspondence visualization.}
Figure~\ref{fig:hscf} further illustrates the effect of the correspondence direction.
The top row shows the original alignment, while the bottom row applies the same fixed translation to the human mesh to make the directional difference more visible.
In practice, the estimated scene points within the human mask inevitably contain noise, as the geometry is affected by varying camera distances and scene model limitations, and the cropped region often includes background clutter or interacting objects.
Under Human$\rightarrow$Scene, background or object points only serve as candidate targets, so visible human points typically match nearby scene support.
Under Scene$\rightarrow$Human, every scene point inside the human mask becomes a query, causing background or object points to form long correspondences to the human surface.
Therefore, the Scene$\rightarrow$Human query direction is more sensitive to noise within the human mask.

\paragraph{Released-code evaluation.}
For completeness, Table~\ref{tab:hscf} provides an additional evaluation using the officially released codebase protocol, serving as a supplement to Table~\ref{tab:hsa} in the main paper.
We rerun the officially released \SHOWMethod{}~\citep{shi2026scene} evaluation code within the same pipeline to reproduce and compare different baselines.
All variants of \OurMethod{} achieve substantially lower HS-CF errors on both the 3DPW~\citep{von2018recovering} and EMDB-1~\citep{kaufmann2023emdb} datasets.
Combined with Table~\ref{tab:hsa} in the main paper, these results demonstrate that \OurMethod{} consistently outperforms the baselines regardless of the underlying evaluation protocol, further highlighting the robustness and stability of our framework in achieving coherent human--scene geometric alignment. Note that \SHOWMethod{} has not released its checkpoint yet; thus, we were unable to evaluate \SHOWMethod{} with either their official evaluation script or our re-implementation according to the original paper. For completeness, we also copy its metrics, along with \HumanThreeR{} and \UniSHMethod{} metrics reported in the original paper, into Table~\ref{tab:hscf} only as a reference (denoted with * and not directly comparable with other numbers in the table), as \SHOWMethod{}'s current official evaluation script cannot reproduce these metrics.

\subsection{Out-of-Distribution Qualitative Results}

\input{figs/occlusion}
\input{figs/interaction}
We further evaluate \OurMethod{} on challenging out-of-distribution scenes with
severe occlusion and multi-person interaction, as shown in
Figures~\ref{fig:occlusion} and~\ref{fig:interaction}.
These examples are drawn from Internet videos together with PROX~\citep{hassan2019resolving} and Hi4D~\citep{yin2023hi4d}.

\paragraph{Robustness to Occlusion.}

Figure~\ref{fig:occlusion} compares human--scene reconstruction under severe occlusion.
In the first row, \OurMethod{} achieves better human--scene alignment and more plausible contact, whereas \HumanThreeR{} places the human above the ground and \UniSHMethod{} shows noticeable mesh--scene misalignment.
In the second row, where the lower body is heavily occluded by a table, our method recovers a more plausible occluded-leg configuration while remaining well aligned with the scene; the baselines fail to reconstruct the lower body reliably, and \UniSHMethod{} additionally exhibits human--scene misalignment.
Importantly, this case also highlights the benefit of our training-free design: while unified baselines typically rely on joint training with paired human--scene data, our framework can directly incorporate stronger human and scene foundation models without retraining, so improvements from newer backbones can be immediately combined with our geometric alignment.
These results demonstrate both the robustness and flexibility of \OurMethod{}
under strong occlusion.

\paragraph{Multi-Person Interaction.}
Figure~\ref{fig:interaction} shows challenging multi-person interactions.
By grounding the human meshes with foreground scene geometry, \OurMethod{}
better preserves relative human positions and produces more plausible contact
between interacting people.
Scene grounding also provides additional depth cues for resolving front--back
ordering, improving the geometric consistency of multi-person interactions.

\subsection{Qualitative Results on In-the-Wild Internet Videos}

\input{figs/internet}

We further evaluate \OurMethod{} on diverse Internet videos in Figure~\ref{fig:internet}, including multi-person scenes, dancing, long sequences, and sports motions.
Despite the substantial domain shift, \OurMethod{} maintains accurate human--scene alignment and consistent human--scene geometry over long sequences.
These results demonstrate that \OurMethod{} generalizes robustly beyond benchmark datasets and remains effective for diverse, unconstrained real-world videos.

\subsection{Limitations and Failure Cases}
\label{app:limitation}

\input{figs/failure}

Figure~\ref{fig:failure} illustrates representative failure cases.
As discussed in Sec.~\ref{app:ablation_human_motion}, our method remains dependent on scene reconstruction quality.
Large viewpoint changes or distant humans can degrade the predicted foreground geometry around human regions, reducing the reliability of scene-scale calibration and human translation refinement.
As shown on the left of Figure~\ref{fig:failure}, this can occasionally lead to local body--scene misalignment or sparse outliers.
Although our pipeline employs a fallback strategy that retains the raw camera-space human translation when the number of reliable inliers is insufficient, a small number of mismatched correspondences may still remain in challenging cases.
Among the evaluated scene backbones, \VGGTOmega{}~\citep{wang2026vggt} provides more reliable scene geometry, enabling stronger human translation refinement.
As scene foundation models continue to improve, increasingly accurate geometry is expected to further enhance human translation and global motion reconstruction.

Our framework also does not explicitly optimize the tracking capability of the underlying human foundation model.
As shown on the right, when multiple people are extremely close, occasional identity association errors can cause overlapping human reconstructions.
This limitation is orthogonal to our alignment framework and could be further improved with stronger dense human detection, tracking, or human mesh recovery models.
Finally, our geometric formulation does not explicitly address temporal jitter or foot sliding; learned temporal refinement with contact constraints would be an interesting direction for future work.

%% file: figs/scene_scale_appendix.tex
\begin{figure*}[t]
    \centering
    \includegraphics[width=\linewidth]{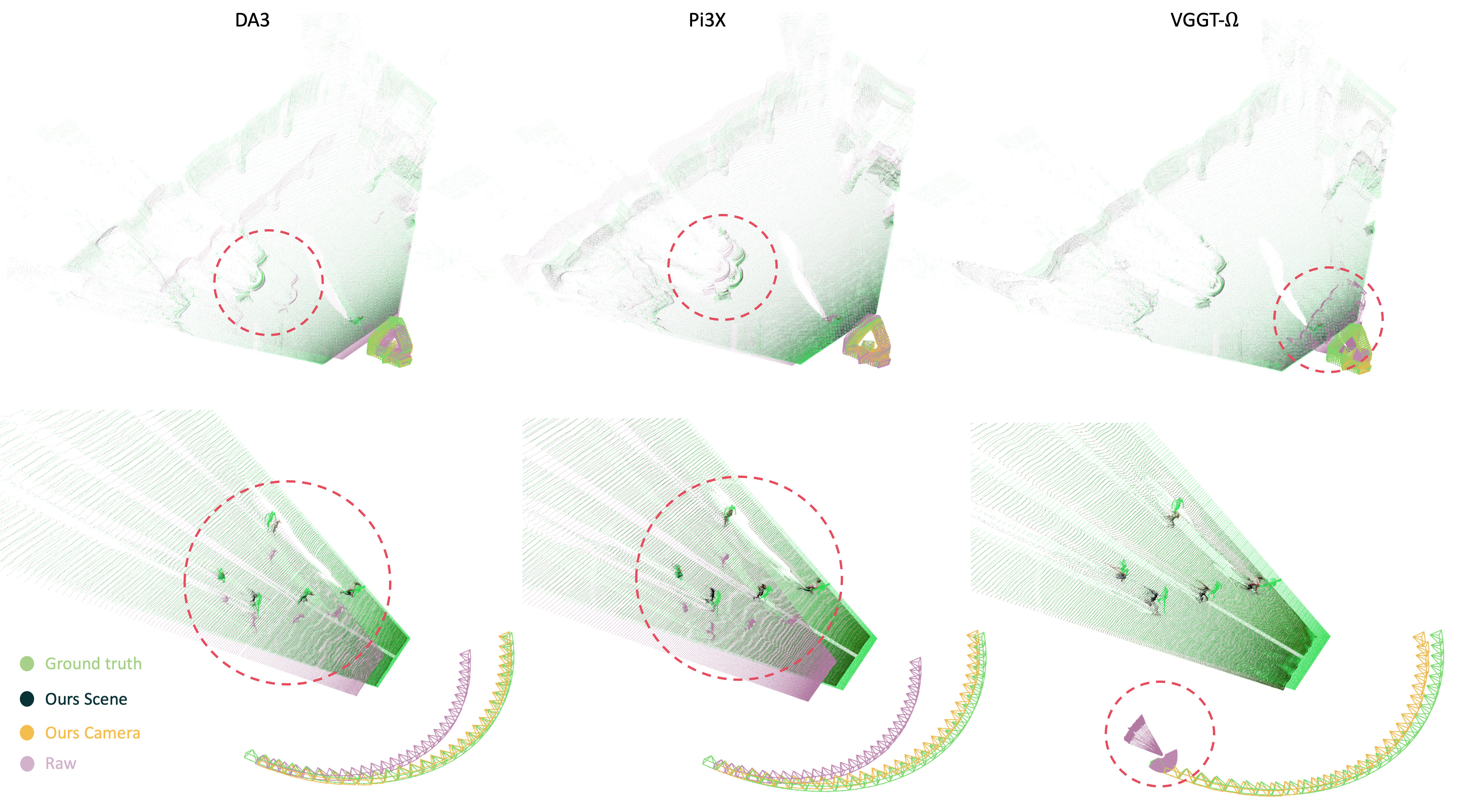}
    \vspace{-5mm}
    \caption{
        \textbf{Metric scene and camera motion on BEDLAM and BEDLAM2.}
        Our calibration brings the raw scene geometry and camera trajectory closer to the ground-truth metric scale. 
        Specifically, \VGGTOmega{} produces normalized predictions, and our scale calibration converts them into near-metric reconstructions.
    }
    \label{fig:scene_scale_appendix}
\end{figure*}

%% file: tables/metric_scene_bedlam_pair.tex
\begin{table*}[t]
    \centering

    \input{tables/metric_scene_bedlam}
    \hfill
    \input{tables/bedlam2_orig}

\end{table*}

%% file: tables/metric_scene_bedlam.tex
\begin{minipage}[t]{0.45\textwidth}
    \vspace{0pt}
    \centering

    \captionof{table}{
        Metric scene reconstruction and camera pose accuracy on BEDLAM.
        Raw denotes the native output of each scene foundation model.
        Our scale calibration uses \MHMRtwo{} as the metric human anchor.
        Lower is better.
    }
    \label{tab:metric_scene_bedlam}

    \scriptsize
    \setlength{\tabcolsep}{2.2pt}
    \renewcommand{\arraystretch}{0.90}

    \resizebox{\linewidth}{!}{    \begin{tabular}{@{}llcccc@{}}
        \toprule
        Scene Backbone
        & Scale Calibration
        & P-Err. $\downarrow$
        & AbsRel $\downarrow$
        & ATE $\downarrow$
        & RPE-t $\downarrow$ \\
        \midrule

        \multirow{2}{*}{\DAThree{}~\citep{lin2025depth}}
        & Raw
        & 2024.9
        & 0.417
        & 191.0
        & 76.51 \\

        & Ours w/ \MHMRtwo{}~\citep{fiche2026multi}
        & \textbf{1645.1}
        & \textbf{0.338}
        & \textbf{114.6}
        & \textbf{47.46} \\

        \cmidrule(lr){1-6}

        \multirow{2}{*}{\PiThreeX{}~\citep{wang2026pi}}
        & Raw
        & 1909.8
        & 0.311
        & 228.8
        & 93.67 \\

        & Ours w/ \MHMRtwo{}~\citep{fiche2026multi}
        & \textbf{1261.8}
        & \textbf{0.161}
        & \textbf{108.1}
        & \textbf{47.75} \\

        \cmidrule(lr){1-6}

        \multirow{2}{*}{\VGGTOmega{}~\citep{wang2026vggt}}
        & Raw
        & 3155.9
        & 0.915
        & 576.4
        & 228.0 \\

        & Ours w/ \MHMRtwo{}~\citep{fiche2026multi}
        & \textbf{993.4}
        & \textbf{0.268}
        & \textbf{24.83}
        & \textbf{10.34} \\

        \bottomrule
    \end{tabular}    }

\end{minipage}

%% file: tables/bedlam2_orig.tex
\begin{minipage}[t]{0.53\textwidth}
    \vspace{0pt}
    \centering

    \captionof{table}{
        \textbf{Metric scene reconstruction and camera pose accuracy
        on BEDLAM2 (all valid scene points).}
        Raw denotes the native scene prediction
        (\VGGTOmega{}: normalized), while our variants
        calibrate scene scale using \MHMRtwo{} or
        \SAMThreeDBody{} as metric anchors.
        Point Error, ATE, and RPE-t are reported in centimeters.
        The best result within each backbone is boldfaced;
        \bestmark~overall best and
        \secondmark~overall second-best.
    }
    \label{tab:bedlam_bak}

    \setlength{\tabcolsep}{2.5pt}
    \renewcommand{\arraystretch}{1.05}

    \resizebox{\linewidth}{!}{    \begin{tabular}{@{}llccccc@{}}
        \toprule

        Scene Backbone
        & Scale Calibration
        & P-Err. $\downarrow$
        & AbsRel $\downarrow$
        & $\delta < 1.25$ $\uparrow$
        & ATE $\downarrow$
        & RPE-t $\downarrow$ \\

        \midrule

                \multicolumn{2}{l}{HAC~\citep{yang2025humans}}
        & --
        & --
        & --
        & 38.66
        & 16.85 \\

        \multicolumn{2}{l}{\HumanThreeR~\citep{chen2025human3r}}
        & 3692.3
        & 0.250
        & 57.69
        & 26.14
        & 11.99 \\

        \multicolumn{2}{l}{\UniSHMethod~\citep{li2026unish}}
        & 3506.0
        & 0.210
        & 66.03
        & 70.80
        & 35.13 \\

        \midrule

                \multirow{3}{*}{\DAThree{}~\citep{lin2025depth}}
        & Raw
        & 3537.0
        & 0.235
        & 54.84
        & 23.98
        & 9.98 \\

        & Ours w/ \MHMRtwo{}~\citep{fiche2026multi}
        & 3470.2
        & 0.192
        & 70.89
        & \textbf{16.29}
        & \textbf{6.96} \\

        & Ours w/ \SAMThreeDBody{}~\citep{yang2026sam}
        & \textbf{3421.1}
        & \textbf{0.190}
        & \textbf{74.19}
        & 17.42
        & 7.42 \\

        \midrule

                \multirow{3}{*}{\PiThreeX{}~\citep{wang2026pi}}
        & Raw
        & 3529.5
        & 0.194
        & 66.39
        & 17.41
        & 7.66 \\

        & Ours w/ \MHMRtwo{}~\citep{fiche2026multi}
        & \textbf{3445.7}
        & \cellfirst{\textbf{0.125}}
        & \cellfirst{\textbf{90.01}}
        & \cellfirst{\textbf{10.85}}
        & \cellfirst{\textbf{4.97}} \\

        & Ours w/ \SAMThreeDBody{}~\citep{yang2026sam}
        & 3499.8
        & 0.169
        & 82.45
        & 15.96
        & 7.01 \\

        \midrule

                \multirow{3}{*}{\VGGTOmega{}~\citep{wang2026vggt}}
        & Raw
        & 4404.9
        & 0.901
        & 0.006
        & 111.9
        & 44.93 \\

        & Ours w/ \MHMRtwo{}~\citep{fiche2026multi}
        & \cellsecond{3288.5}
        & \cellsecond{\textbf{0.132}}
        & \cellsecond{\textbf{88.88}}
        & \cellsecond{\textbf{10.98}}
        & \cellsecond{\textbf{5.22}} \\

        & Ours w/ \SAMThreeDBody{}~\citep{yang2026sam}
        & \cellfirst{\textbf{3275.5}}
        & 0.165
        & 84.18
        & 16.56
        & 7.25 \\

        \bottomrule
    \end{tabular}    }

\end{minipage}

%% file: tables/human_traj_appendix.tex
\begin{table*}[t]
    \centering
    \vspace{-2mm}
    
    \caption{
        Additional global human motion reconstruction results on BEHAVE, RICH, and EMDB-2.
                        We extend our evaluation to the earlier \HMRtwoMethod{} backbone, presenting comprehensive results alongside \MHMRtwo{} across different scene backbones, including \DAThree{}, \PiThreeX{}, and \VGGTOmega{}.
        $^\dagger$ applies scale calibration followed by translation refinement.
        \bestmark~Best result; \secondmark~Second-best result.
        Lower is better for all metrics.
    }
    \label{tab:human_traj_appendix}

    \setlength{\tabcolsep}{1.8pt}
    \renewcommand{\arraystretch}{1.02}

    \resizebox{\linewidth}{!}{    \begin{tabular}{@{}lcccc|cccc|cccc@{}}
        \toprule

        \multirow{2}{*}{Method}
        & \multicolumn{4}{c|}{BEHAVE}
        & \multicolumn{4}{c|}{RICH}
        & \multicolumn{4}{c}{EMDB-2} \\

        \cmidrule(lr){2-5}
        \cmidrule(lr){6-9}
        \cmidrule(lr){10-13}

        & G-MPJPE $\downarrow$
        & W-MPJPE $\downarrow$
        & WA-MPJPE $\downarrow$
        & RTE $\downarrow$

        & G-MPJPE $\downarrow$
        & W-MPJPE $\downarrow$
        & WA-MPJPE $\downarrow$
        & RTE $\downarrow$

        & G-MPJPE $\downarrow$
        & W-MPJPE $\downarrow$
        & WA-MPJPE $\downarrow$
        & RTE $\downarrow$ \\

        \midrule

        \GVHMRMethod{}~\citep{shen2024world}
        & 509.8
        & 133.8
        & 80.0
        & 1.00

        & 277.7
        & \cellsecond{126.3}
        & \cellsecond{78.8}
        & 2.40

        & 3487.5
        & 276.5
        & 111.0
        & 2.00 \\

        \GEMMethod{}~\citep{li2025genmo}
        & 550.6
        & 151.7
        & 90.6
        & 1.10

        & 314.8
        & \cellfirst{\textbf{118.6}}
        & \cellfirst{\textbf{75.3}}
        & \cellsecond{1.90}

        & 3522.5
        & 202.1
        & 74.3
        & \cellfirst{\textbf{1.20}} \\

        \cmidrule(lr){1-13}

        HAC~\citep{yang2025humans}
        & 117.8
        & 97.3
        & 64.7
        & \cellsecond{0.28}

        & 459.2
        & 244.9
        & 158.4
        & 4.3

        & 2594.9
        & \cellsecond{197.3}
        & \cellsecond{69.1}
        & \cellsecond{1.23} \\

        \JOSHMethod{}~\citep{liu2026joint}
        & --
        & --
        & --
        & --

        & --
        & 184.3
        & 102.6
        & 3.40

        & --
        & \cellfirst{\textbf{174.7}}
        & \cellfirst{\textbf{68.9}}
        & 1.30 \\

        \cmidrule(lr){1-13}

        \JOSHThreeR{}~\citep{liu2026joint,chen2025human3r}
        & --
        & --
        & --
        & --

        & --
        & 334.9
        & 190.4
        & 6.30

        & --
        & 661.7
        & 220.0
        & 13.10 \\

        \MetricHMSRMethod{}~\citep{song2026metrichmsr}
        & --
        & --
        & --
        & --

        & --
        & 165.5
        & 109.6
        & \cellsecond{1.90}

        & --
        & 199.5
        & 72.1
        & 1.40 \\

        \HumanThreeR{}~\citep{chen2025human3r}
        & 360.3
        & 112.2
        & 70.3
        & 0.73

        & 349.3
        & 184.9
        & 110.0
        & 3.30

        & 4866.0
        & 267.9
        & 112.2
        & 2.20 \\

        \UniSHMethod{}~\citep{li2026unish}
        & 349.4
        & 239.9
        & 134.8
        & 1.11

        & 409.4
        & 183.2
        & 118.1
        & 4.80

        & 6099.7
        & 270.1
        & 118.5
        & 5.80 \\

        \CHROMMMethod{}-mono~\citep{kim2026coherent}
        & --
        & --
        & --
        & --

        & --
        & 138.3
        & 87.5
        & 3.30

        & --
        & 255.0
        & 102.6
        & 1.70 \\

        \UniConThreeR{}~\citep{sur2026unicon3r}
        & --
        & --
        & --
        & --

        & --
        & 129.8
        & 81.5
        & \cellfirst{\textbf{1.70}}

        & --
        & 285.6
        & 113.7
        & 2.30 \\

        \SHOWMethod{}~\citep{shi2026scene}
        & --
        & --
        & --
        & --

        & --
        & 172.7
        & 107.3
        & 2.20

        & --
        & 262.3
        & 109.1
        & 2.10 \\

        \midrule

        \multicolumn{13}{l}{\textbf{Ours}} \\

        \addlinespace[1pt]

        \MHMRtwo{} + \DAThree{}~\citep{fiche2026multi,lin2025depth}$^\dagger$
        & 113.2
        & 101.6
        & 68.7
        & 0.32

        & 246.1
        & 179.4
        & 104.9
        & 2.97

        & \cellfirst{\textbf{1539.9}}
        & 234.4
        & 89.9
        & 1.41 \\

        \MHMRtwo{} + \PiThreeX{}~\citep{fiche2026multi,wang2026pi}$^\dagger$
        & 112.1
        & \cellsecond{93.8}
        & 64.4
        & \cellfirst{\textbf{0.26}}

        & \cellfirst{\textbf{211.3}}
        & 151.3
        & 92.8
        & 2.32

        & 3313.8
        & 328.2
        & 119.7
        & 3.29 \\

        \MHMRtwo{} + \VGGTOmega{}~\citep{fiche2026multi,wang2026vggt}$^\dagger$
        & 120.1
        & 99.9
        & 66.7
        & \cellsecond{0.28}

        & \cellsecond{226.9}
        & 167.0
        & 96.3
        & 2.70

        & \cellsecond{1580.1}
        & 210.9
        & 78.9
        & 1.31 \\

        \addlinespace[1pt]

        \HMRtwoMethod{} + \DAThree{}~\citep{goel2023humans,lin2025depth}$^\dagger$
        & 117.9
        & 105.5
        & 65.3
        & 0.38

        & 270.1
        & 159.7
        & 97.5
        & 2.78

        & 2293.8
        & 300.2
        & 112.2
        & 1.69 \\

        \HMRtwoMethod{} + \PiThreeX{}~\citep{goel2023humans,wang2026pi}$^\dagger$
        & \cellsecond{109.3}
        & \cellfirst{\textbf{93.4}}
        & \cellfirst{\textbf{58.3}}
        & 0.29

        & 234.5
        & 130.5
        & 83.3
        & 1.95

        & 2869.3
        & 372.5
        & 131.3
        & 2.96 \\

        \HMRtwoMethod{} + \VGGTOmega{}~\citep{goel2023humans,wang2026vggt}$^\dagger$
        & \cellfirst{\textbf{105.2}}
        & 95.8
        & \cellsecond{59.1}
        & \cellsecond{0.28}

        & 252.5
        & 143.1
        & 85.9
        & 2.49

        & 1825.2
        & 261.7
        & 88.8
        & 1.41 \\

        \bottomrule
    \end{tabular}    }
    \vspace{-2mm}
    
\end{table*}

%% file: figs/qualitative_appendix.tex
\begin{figure*}[t]
    \centering
    \includegraphics[width=\linewidth]{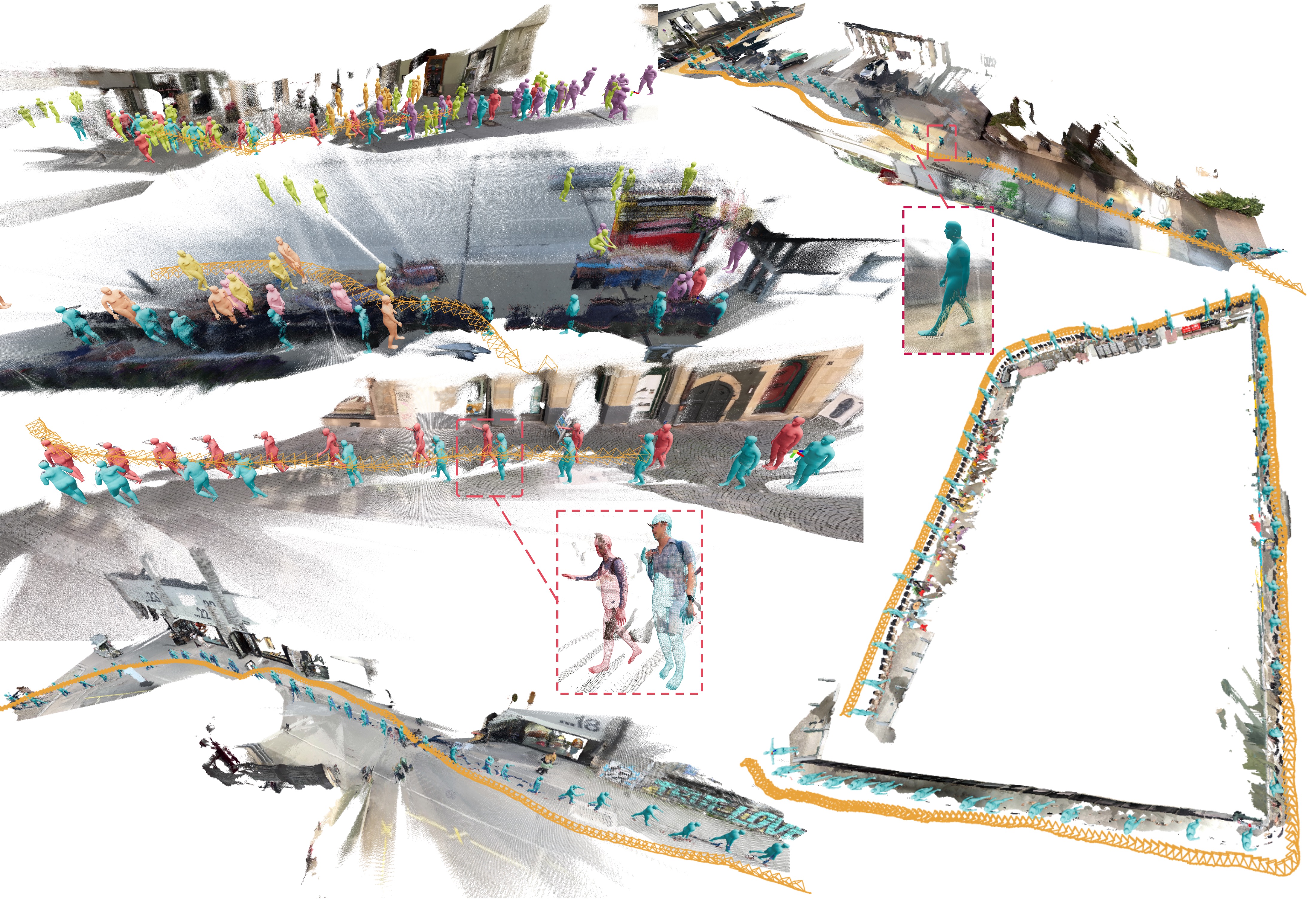}
        \vspace{-6mm}
        \caption{
        \textbf{Qualitative results of human--scene reconstruction} on 3DPW and EMDB-2, including multi-person scenes and long sequences.
    }
    \label{fig:qualitative_appendix}
\end{figure*}

%% file: figs/hsa_appendix.tex
\begin{figure*}[t]
        \centering
    \includegraphics[width=\linewidth]{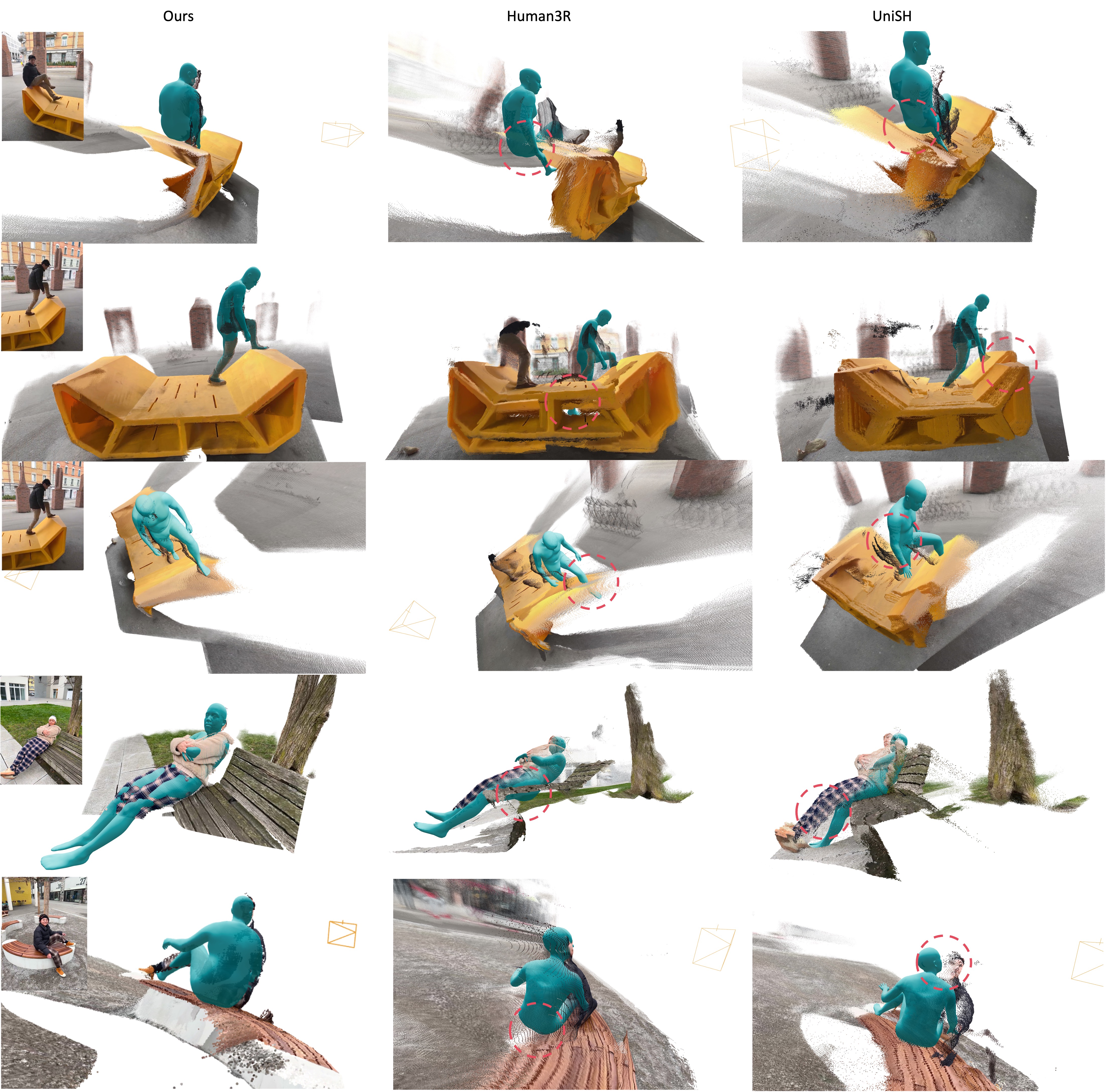}
    \vspace{-6mm}
    \caption{
        \textbf{Qualitative comparison of human--scene alignment} on EMDB-2.
        \OurMethod{} yields more accurate human--scene alignment and contact than \HumanThreeR{} and \UniSHMethod{}.
    }
    \label{fig:hsa_appendix}
    \vspace{-5mm}
\end{figure*}

%% file: figs/hsa_appendix_behave.tex
\begin{figure*}[t]
        \centering
    \includegraphics[width=\linewidth]{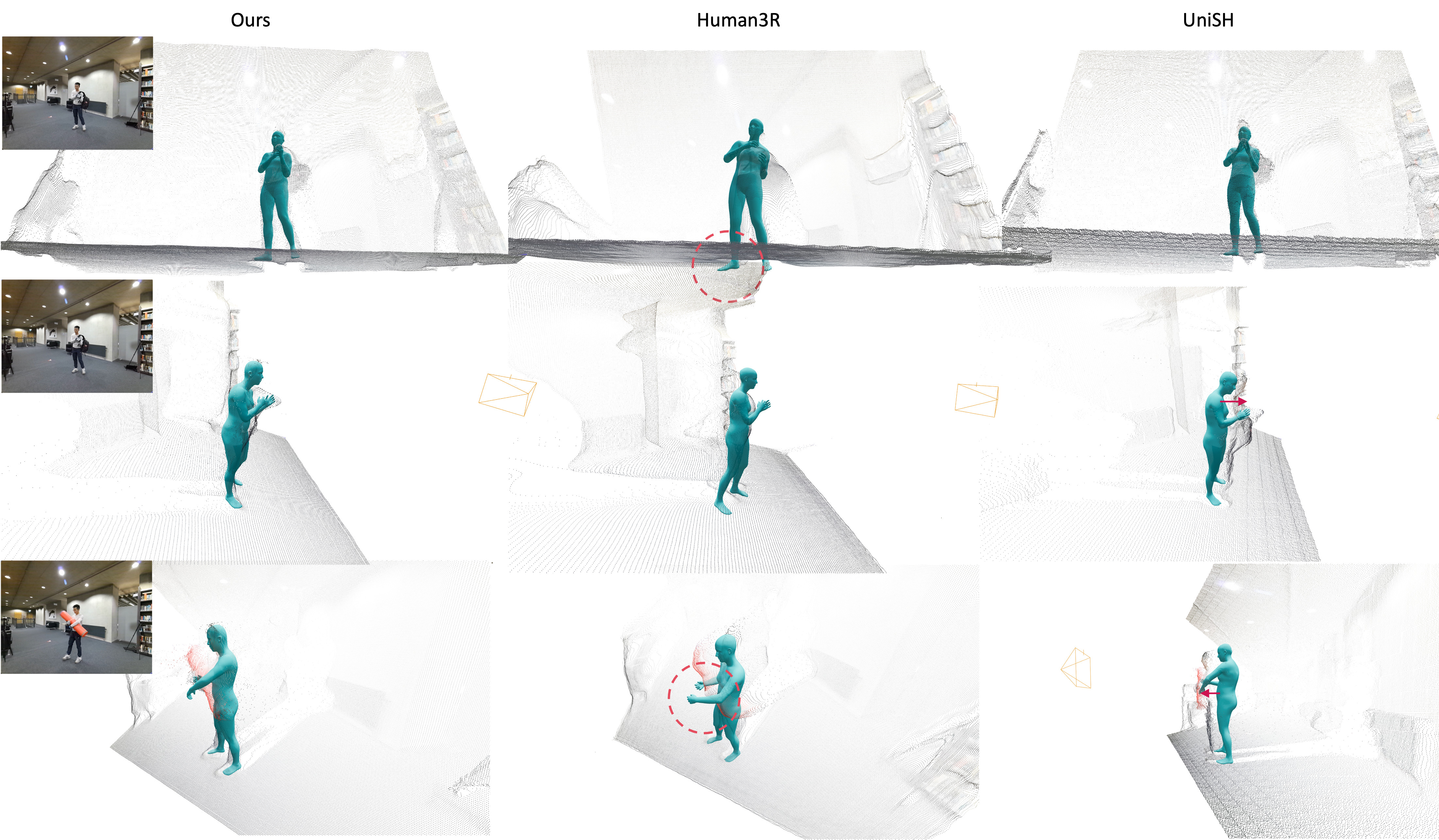}
    \vspace{-6mm}
    \caption{
        \textbf{Qualitative comparison of human--scene alignment} on BEHAVE.
        \OurMethod{} achieves more accurate human--scene alignment and more plausible human--object contact than \HumanThreeR{} and \UniSHMethod{}.
    }
    \label{fig:hsa_behave_appendix}
    \vspace{-6mm}
\end{figure*}

%% file: figs/human_motion.tex
\begin{figure}[p]
    \centering
        \includegraphics[width=\textwidth,height=\textheight,keepaspectratio]{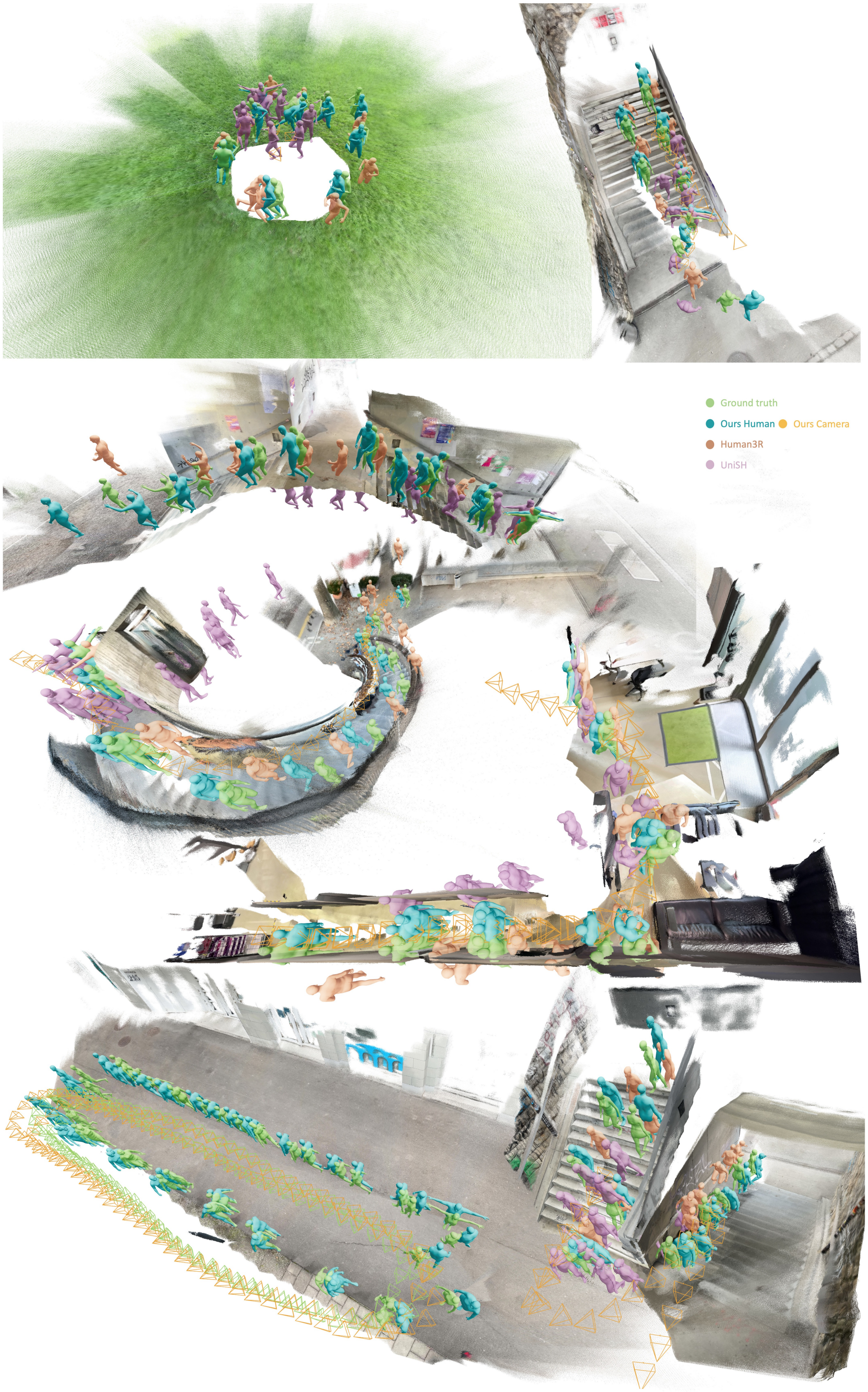}
    \vspace{-4mm}
    \caption{
        \textbf{Qualitative comparison of global human motion} on EMDB-2.
        \OurMethod{} exhibits less long-term global trajectory drift and better matches the ground truth than the baselines.
    }
    \label{fig:human_motion}
    \end{figure}

%% file: tables/hsa_physical_pair.tex
\begin{table*}[t]
    \centering

    \input{tables/hsa_3db}
    \hfill
    \input{tables/contact}

\end{table*}

%% file: tables/hsa_3db.tex
\begin{minipage}[t]{0.50\textwidth}
    \vspace{0pt}
    \centering

    \captionof{table}{
        \textbf{Quantitative comparison of human--scene alignment using 3DB}
        on 3DPW and EMDB-1 using HS-CF$_5$ and HS-CF$_{10}$.
        \bestmark~Best result; \secondmark~Second-best result.
    }
    \label{tab:hsa_3db}

    \vspace{-2mm}

    \setlength{\tabcolsep}{3.5pt}
    \renewcommand{\arraystretch}{1.0}

    \resizebox{\linewidth}{!}{    \begin{tabular}{@{}lcc|cc@{}}
        \toprule

        \multirow{2}{*}{Method}
        & \multicolumn{2}{c|}{3DPW}
        & \multicolumn{2}{c}{EMDB-1} \\

        \cmidrule(lr){2-3}
        \cmidrule(lr){4-5}

        & HS-CF$_5$ $\downarrow$
        & HS-CF$_{10}$ $\downarrow$
        & HS-CF$_5$ $\downarrow$
        & HS-CF$_{10}$ $\downarrow$ \\

        \midrule

        \HumanThreeR{}~\citep{chen2025human3r}
        & 1.079
        & 1.315
        & 0.719
        & 0.924 \\

        \UniSHMethod{}~\citep{li2026unish}
        & 3.676
        & 4.183
        & 1.166
        & 1.528 \\

        \midrule

        \multicolumn{5}{l}{\textbf{Ours}} \\

        \SAMThreeDBody{} + \DAThree{}~\citep{lin2025depth}
        & 0.751
        & 0.955
        & 0.482
        & 0.671 \\

        \SAMThreeDBody{} + \PiThreeX{}~\citep{wang2026pi}
        & \cellsecond{0.670}
        & \cellsecond{0.894}
        & \cellsecond{0.419}
        & \cellsecond{0.616} \\

        \SAMThreeDBody{} + \VGGTOmega{}~\citep{wang2026vggt}
        & \cellfirst{\textbf{0.641}}
        & \cellfirst{\textbf{0.856}}
        & \cellfirst{\textbf{0.401}}
        & \cellfirst{\textbf{0.601}} \\

        \bottomrule
    \end{tabular}    }

\end{minipage}

%% file: tables/contact.tex
\begin{minipage}[t]{0.48\textwidth}
    \vspace{0pt}
    \centering

    \captionof{table}{
        \textbf{Quantitative evaluation of physical grounding on EMDB-2.}
        We evaluate foot--ground contact via Ground Distance
        ($\text{GD} = \text{GP} + \text{GF}$),
        Ground Penetration (GP), and Ground Floating (GF),
        as well as full-body scene/object interpenetration via
        Normal-Based Penetration (NBP).
        Our method uses \SAMThreeDBody{} + \VGGTOmega{} with
        \textit{+scale\&trans}.
        All metrics are reported in $\text{cm}$ ($\downarrow$).
        \bestmark~Best result; \secondmark~Second-best result.
    }
    \label{tab:physical_grounding}

    \vspace{-2mm}

    \setlength{\tabcolsep}{4.0pt}
    \renewcommand{\arraystretch}{1.05}

    \resizebox{\linewidth}{!}{    \begin{tabular}{@{}lcccc@{}}
        \toprule

        Method
        & GD (cm) $\downarrow$
        & GP (cm) $\downarrow$
        & GF (cm) $\downarrow$
        & NBP (cm) $\downarrow$ \\

        \midrule

        \HumanThreeR{}~\citep{chen2025human3r}
        & \cellsecond{11.926}
        & 11.192
        & \cellfirst{0.734}
        & 3.440 \\

        \UniSHMethod{}~\citep{li2026unish}
        & 13.970
        & \cellsecond{6.723}
        & 7.247
        & \cellsecond{2.336} \\

        \midrule

        Ours
        & \cellfirst{4.699}
        & \cellfirst{3.902}
        & \cellsecond{0.797}
        & \cellfirst{1.795} \\

        \bottomrule
    \end{tabular}    }

\end{minipage}

%% file: tables/human3r_pair.tex
\begin{table*}[t]
    \centering

    \input{tables/human3r_hsa_ablation}
    \hfill
    \input{tables/human3r_motion_ablation}

\end{table*}

%% file: tables/human3r_hsa_ablation.tex
\begin{minipage}[t]{0.46\textwidth}
    \vspace{0pt}
    \centering

    \captionof{table}{
        \textbf{Ablation of applying \OurMethod{} to \HumanThreeR{} and \UniSHMethod{}
        for human--scene geometric alignment on 3DPW and EMDB-1.}
        The baselines w/ \OurMethod{} apply scale calibration and translation
        refinement to the baselines’ human and scene predictions.
        Lower is better, and the best result within each baseline block is boldfaced.
    }
    \label{tab:human3r_hsa_ablation}

    \setlength{\tabcolsep}{3.5pt}
    \renewcommand{\arraystretch}{1.0}

    \resizebox{\linewidth}{!}{    \begin{tabular}{@{}lcc|cc@{}}
        \toprule

        \multirow{2}{*}{Method}
        & \multicolumn{2}{c|}{3DPW}
        & \multicolumn{2}{c}{EMDB-1} \\

        \cmidrule(lr){2-3}
        \cmidrule(lr){4-5}

        & HS-CF$_5$ $\downarrow$
        & HS-CF$_{10}$ $\downarrow$
        & HS-CF$_5$ $\downarrow$
        & HS-CF$_{10}$ $\downarrow$ \\

        \midrule

        \HumanThreeR{}~\citep{chen2025human3r}
        & 1.079
        & 1.315
        & 0.719
        & 0.924 \\

        \HumanThreeR{} w/ \OurMethod{}
        & \textbf{0.793}
        & \textbf{1.035}
        & \textbf{0.579}
        & \textbf{0.782} \\

        \cmidrule(lr){1-5}

        \UniSHMethod{}~\citep{li2026unish}
        & 3.676
        & 4.183
        & 1.166
        & 1.528 \\

        \UniSHMethod{} w/ \OurMethod{}
        & \textbf{0.438}
        & \textbf{0.618}
        & \textbf{0.653}
        & \textbf{0.958} \\

        \bottomrule
    \end{tabular}    }

\end{minipage}

%% file: tables/human3r_motion_ablation.tex
\begin{minipage}[t]{0.52\textwidth}
    \vspace{0pt}
    \centering

    \captionof{table}{
        \textbf{Ablation of applying \OurMethod{} to existing baselines and an earlier scene backbone on BEHAVE and EMDB-2.}
        For \HumanThreeR{} and \UniSHMethod{}, w/ \OurMethod{} applies our scale calibration and scene-grounded translation refinement.
        For \MHMRtwo{} paired with the earlier scene model Pi3, \textit{+Scale} applies scale calibration only, while \textit{+Scale+Trans} additionally refines human translation.
        All metrics are reported in mm ($\downarrow$), and the best result within each block is boldfaced.
    }
    \label{tab:human3r_motion}

    \setlength{\tabcolsep}{3.5pt}
    \renewcommand{\arraystretch}{1.0}

    \resizebox{\linewidth}{!}{    \begin{tabular}{@{}lccc|ccc@{}}
        \toprule

        \multirow{2}{*}{Method}
        & \multicolumn{3}{c|}{BEHAVE}
        & \multicolumn{3}{c}{EMDB-2} \\

        \cmidrule(lr){2-4}
        \cmidrule(lr){5-7}

        & G-MPJPE $\downarrow$
        & W-MPJPE $\downarrow$
        & WA-MPJPE $\downarrow$
        & G-MPJPE $\downarrow$
        & W-MPJPE $\downarrow$
        & WA-MPJPE $\downarrow$ \\

        \midrule

        \multicolumn{7}{@{}l}{
            \textbf{\HumanThreeR{}}
            ~\citep{chen2025human3r}
        } \\

        \quad Original
        & \textbf{360.3}
        & \textbf{112.2}
        & \textbf{70.3}
        & 4866.0
        & 267.9
        & 112.2 \\

        \quad w/ \OurMethod{}
        & 374.2
        & 119.2
        & 76.1
        & \textbf{4865.0}
        & \textbf{259.6}
        & \textbf{109.3} \\

        \midrule

        \multicolumn{7}{@{}l}{
            \textbf{\UniSHMethod{}}
            ~\citep{li2026unish}
        } \\

        \quad Original
        & 349.4
        & 239.9
        & 134.8
        & 6099.7
        & \textbf{270.1}
        & \textbf{118.5} \\

        \quad w/ \OurMethod{}
        & \textbf{166.7}
        & \textbf{119.7}
        & \textbf{75.7}
        & \textbf{6038.3}
        & 304.1
        & 129.5 \\

        \midrule

        \multicolumn{7}{@{}l}{
            \textbf{\MHMRtwo{} + Pi3}
        } \\

        \quad +Scale
        & 132.8
        & 99.3
        & 68.0
        & \textbf{3399.3}
        & 344.7
        & 125.0 \\

        \quad +Scale+Trans
        & \textbf{116.4}
        & \textbf{95.7}
        & \textbf{65.2}
        & 3405.9
        & \textbf{342.3}
        & \textbf{122.5} \\

        \bottomrule
    \end{tabular}    }

\end{minipage}

%% file: figs/ablation_hsa_baseline.tex
\begin{figure*}[t]
    \centering
    \includegraphics[width=\linewidth]{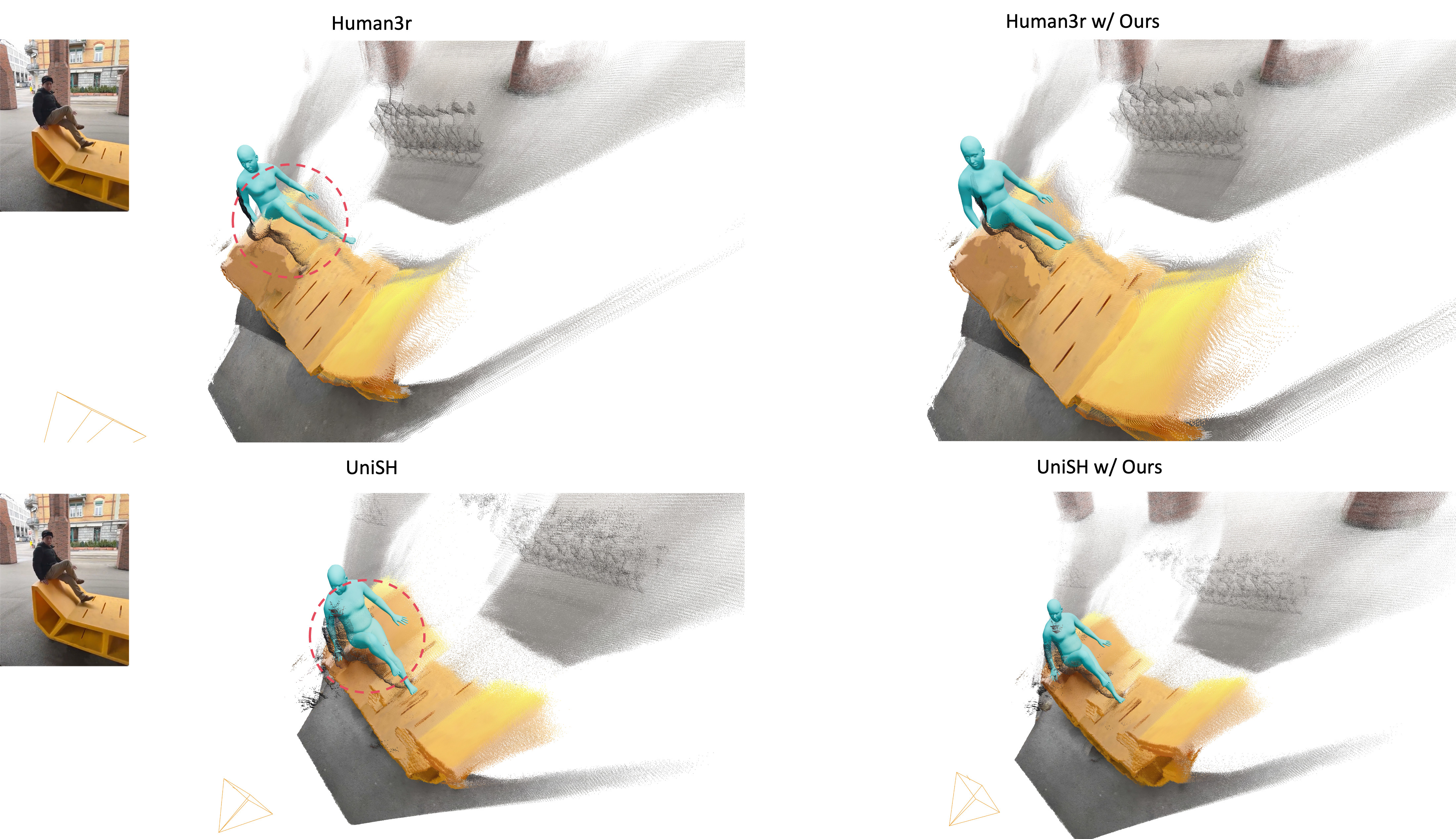}
    \vspace{-4mm}
    \caption{
        \textbf{Improving human--scene alignment in \HumanThreeR{} and \UniSHMethod{}.} Applying \OurMethod{} to either baseline improves human--scene alignment, bringing the human mesh into closer agreement with the reconstructed scene.
    }
    \label{fig:ablation_hsa_baseline}
\end{figure*}

%% file: tables/human_traj_ablation.tex
\begin{table*}[t]
    \centering
    \vspace{-2mm}
    \caption{
        \textbf{Ablation of scale calibration and translation refinement for global human motion}
        with \MHMRtwo{}, \HMRtwoMethod{}, and \SAMThreeDBody{} on BEHAVE, RICH, and EMDB-2.
        \textit{Raw}: original human/camera trajectories;
        \textit{+scale}: camera-scale calibration;
        \textit{+scale\&trans}: additional scene-grounded human translation refinement.
        Lower is better, and the best result for each human--scene backbone combination is boldfaced.
    }
    \label{tab:human_traj_ablation}

    \scriptsize
    \setlength{\tabcolsep}{2.2pt}
    \renewcommand{\arraystretch}{0.90}

    \resizebox{\linewidth}{!}{    \begin{tabular}{@{}llccc|ccc|ccc@{}}
        \toprule

        \multirow{2}{*}{Scene Backbone}
        & \multirow{2}{*}{Setting}
        & \multicolumn{3}{c|}{\MHMRtwo{}~\citep{fiche2026multi}}
        & \multicolumn{3}{c|}{\HMRtwoMethod{}~\citep{goel2023humans}}
        & \multicolumn{3}{c}{\SAMThreeDBody{}~\citep{yang2026sam}} \\

        \cmidrule(lr){3-5}
        \cmidrule(lr){6-8}
        \cmidrule(lr){9-11}

        &
        & G-MPJPE $\downarrow$
        & W-MPJPE $\downarrow$
        & WA-MPJPE $\downarrow$

        & G-MPJPE $\downarrow$
        & W-MPJPE $\downarrow$
        & WA-MPJPE $\downarrow$

        & G-MPJPE $\downarrow$
        & W-MPJPE $\downarrow$
        & WA-MPJPE $\downarrow$ \\

        \midrule

        \multicolumn{11}{l}{\textbf{BEHAVE}} \\
        \midrule

        \multirow{2}{*}{\DAThree{}~\citep{lin2025depth}}
        & +scale
        & 125.8
        & 105.1
        & 70.0

        & 143.1
        & 110.4
        & 68.7

                & \textbf{85.3}
        & \textbf{72.3}
        & \textbf{46.8} \\

        & +scale\&trans
        & \textbf{113.2}
        & \textbf{101.6}
        & \textbf{68.7}

        & \textbf{117.9}
        & \textbf{105.5}
        & \textbf{65.3}

                & 101.3
        & 84.7
        & 54.5 \\

        \cmidrule(lr){1-11}

        \multirow{2}{*}{\PiThreeX{}~\citep{wang2026pi}}
        & +scale
        & 123.4
        & 98.8
        & 67.7

        & 139.3
        & 105.0
        & 65.8

                & \textbf{79.7}
        & \textbf{64.7}
        & \textbf{42.5} \\

        & +scale\&trans
        & \textbf{112.1}
        & \textbf{93.8}
        & \textbf{64.4}

        & \textbf{109.3}
        & \textbf{93.4}
        & \textbf{58.3}

                & 92.5
        & 71.2
        & 47.1 \\

        \cmidrule(lr){1-11}

        \multirow{2}{*}{\VGGTOmega{}~\citep{wang2026vggt}}
        & +scale
        & 124.7
        & 103.2
        & 69.8

        & 130.6
        & 104.4
        & 65.6

                & \textbf{77.6}
        & \textbf{69.7}
        & \textbf{45.4} \\

        & +scale\&trans
        & \textbf{120.1}
        & \textbf{99.9}
        & \textbf{66.7}

        & \textbf{105.2}
        & \textbf{95.8}
        & \textbf{59.1}

                & 100.7
        & 79.1
        & 50.2 \\

        \midrule
        \multicolumn{11}{l}{\textbf{RICH}} \\
        \midrule

        \multirow{2}{*}{\DAThree{}~\citep{lin2025depth}}
        & +scale
        & 269.7
        & 209.2
        & 120.0

        & 311.0
        & 185.7
        & 113.1

                & \textbf{142.9}
        & \textbf{100.6}
        & \textbf{66.9} \\

        & +scale\&trans
        & \textbf{246.1}
        & \textbf{179.4}
        & \textbf{104.9}

        & \textbf{270.1}
        & \textbf{159.7}
        & \textbf{97.5}

                & 192.4
        & 131.2
        & 81.3 \\

        \cmidrule(lr){1-11}

        \multirow{2}{*}{\PiThreeX{}~\citep{wang2026pi}}
        & +scale
        & 271.8
        & 209.2
        & 119.8

        & 316.3
        & 185.2
        & 112.6

                & \textbf{145.0}
        & 100.3
        & 66.6 \\

        & +scale\&trans
        & \textbf{211.3}
        & \textbf{151.3}
        & \textbf{92.8}

        & \textbf{234.5}
        & \textbf{130.5}
        & \textbf{83.3}

                & 156.5
        & \textbf{98.0}
        & \textbf{65.5} \\

        \cmidrule(lr){1-11}

        \multirow{2}{*}{\VGGTOmega{}~\citep{wang2026vggt}}
        & +scale
        & 267.3
        & 207.4
        & 119.4

        & 314.8
        & 182.6
        & 111.4

                & \textbf{138.9}
        & \textbf{99.3}
        & \textbf{66.2} \\

        & +scale\&trans
        & \textbf{226.9}
        & \textbf{167.0}
        & \textbf{96.3}

        & \textbf{252.5}
        & \textbf{143.1}
        & \textbf{85.9}

                & 184.9
        & 118.3
        & 70.7 \\

        \midrule
        \multicolumn{11}{l}{\textbf{EMDB-2}} \\
        \midrule

        \multirow{3}{*}{\DAThree{}~\citep{lin2025depth}}
        & Raw
        & 2073.6
        & 251.0
        & 89.6

        & 2593.9
        & 307.1
        & 105.3

                & 1855.4
        & 182.3
        & 73.3 \\

        & +scale
        & \textbf{1520.1}
        & \textbf{216.1}
        & \textbf{80.6}

        & \textbf{2253.9}
        & \textbf{280.1}
        & \textbf{98.1}

                & \textbf{1411.2}
        & \textbf{162.4}
        & \textbf{68.6} \\

        & +scale\&trans
        & 1539.9
        & 234.4
        & 89.9

        & 2293.8
        & 300.2
        & 112.2

                & 1452.9
        & 189.1
        & 84.6 \\

        \cmidrule(lr){1-11}

        \multirow{3}{*}{\PiThreeX{}~\citep{wang2026pi}}
        & Raw
        & 5877.1
        & 411.6
        & 157.2

        & 5493.1
        & 446.2
        & 168.6

                & 5359.4
        & 377.5
        & 144.6 \\

        & +scale
        & \textbf{3303.5}
        & 329.2
        & 122.5

        & 2871.4
        & 374.9
        & 134.4

                & \textbf{2303.4}
        & \textbf{286.3}
        & \textbf{103.9} \\

        & +scale\&trans
        & 3313.8
        & \textbf{328.2}
        & \textbf{119.7}

        & \textbf{2869.3}
        & \textbf{372.5}
        & \textbf{131.3}

                & 2308.8
        & 288.3
        & 105.9 \\

        \cmidrule(lr){1-11}

        \multirow{2}{*}{\VGGTOmega{}~\citep{wang2026vggt}}
        & +scale
        & \textbf{1573.8}
        & 211.2
        & 80.8

        & 1828.1
        & 264.8
        & 92.3

                & \textbf{1203.9}
        & \textbf{152.9}
        & \textbf{68.1} \\

        & +scale\&trans
        & 1580.1
        & \textbf{210.9}
        & \textbf{78.9}

        & \textbf{1825.2}
        & \textbf{261.7}
        & \textbf{88.8}

                & 1208.6
        & 157.0
        & 71.8 \\

        \bottomrule
    \end{tabular}    }
\end{table*}

%% file: tables/least_squares_ablation.tex
\begin{table*}[t]
    \centering 
    \caption{
        \textbf{Comparison of estimators for robust human--scene metric calibration and global human motion.}
        All settings use \MHMRtwo{} and \VGGTOmega{}.
        We compare our RANSAC-based consensus estimator against ordinary least squares (Ordinary LS), trimmed least squares (Trimmed LS), and confidence-weighted iteratively reweighted least squares (Conf-weighted IRLS).
        Lower is better for all metrics, with the best result boldfaced.
    }
    \label{tab:least_squares_ablation}

    \setlength{\tabcolsep}{2.0pt}
    \renewcommand{\arraystretch}{1.02}

    \resizebox{\linewidth}{!}{    \begin{tabular}{@{}lcc|cc|ccc|ccc@{}}
        \toprule

        \multirow{3}{*}{Setting}
        & \multicolumn{4}{c|}{Human--Scene Alignment}
        & \multicolumn{6}{c}{Global Human Motion} \\

        \cmidrule(lr){2-5}
        \cmidrule(lr){6-11}

        & \multicolumn{2}{c|}{3DPW}
        & \multicolumn{2}{c|}{EMDB-1}
        & \multicolumn{3}{c|}{RICH}
        & \multicolumn{3}{c}{EMDB-2} \\

        \cmidrule(lr){2-3}
        \cmidrule(lr){4-5}
        \cmidrule(lr){6-8}
        \cmidrule(lr){9-11}

        & HS-CF$_5$ $\downarrow$
        & HS-CF$_{10}$ $\downarrow$
        & HS-CF$_5$ $\downarrow$
        & HS-CF$_{10}$ $\downarrow$

        & G-MPJPE $\downarrow$
        & W-MPJPE $\downarrow$
        & WA-MPJPE $\downarrow$

        & G-MPJPE $\downarrow$
        & W-MPJPE $\downarrow$
        & WA-MPJPE $\downarrow$ \\

        \midrule

        Ordinary LS
        & 5.614
        & 6.798
        & 3.194
        & 3.428

        & 730.3
        & 289.9
        & 156.3

        & 4865.0
        & 259.7
        & 109.3 \\

        Trimmed LS
        & 1.140
        & 1.404
        & 0.486
        & 0.720

        & 270.7
        & 167.5
        & 96.4

        & 1766.1
        & 222.7
        & 84.6 \\

        Conf-weighted IRLS
        & 2.096
        & 2.371
        & 0.502
        & 0.741

        & 315.4
        & 178.0
        & 98.3

        & 1944.2
        & 350.4
        & 94.7 \\

        \midrule

        Ours (RANSAC)
        & \textbf{0.688}
        & \textbf{0.942}
        & \textbf{0.470}
        & \textbf{0.691}

        & \textbf{226.9}
        & \textbf{167.0}
        & \textbf{96.3}

        & \textbf{1580.1}
        & \textbf{210.9}
        & \textbf{78.9} \\

        \bottomrule
    \end{tabular}    }

\end{table*}

%% file: figs/ransac_ablation.tex
\begin{figure*}[t]
    \centering
    \includegraphics[width=\linewidth]{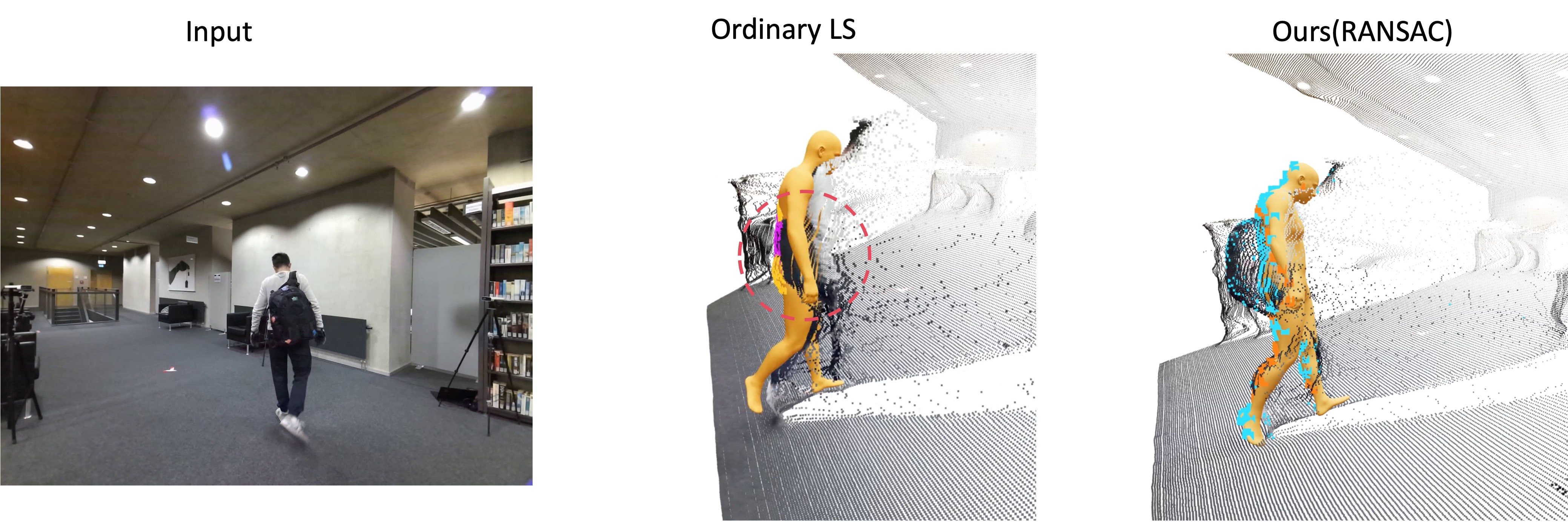}
    \caption{
        \textbf{Qualitative ablation of robust human--scene alignment on BEHAVE.}
        Without RANSAC, least-squares fitting can be biased by overlapping object geometry, while our robust estimator rejects such mismatches and recovers more accurate human--scene alignment.
    }
    \label{fig:ransac_ablation}
\end{figure*}

%% file: figs/HAMC_ablation.tex
\begin{figure*}[t]
    \centering
    \includegraphics[width=\linewidth]{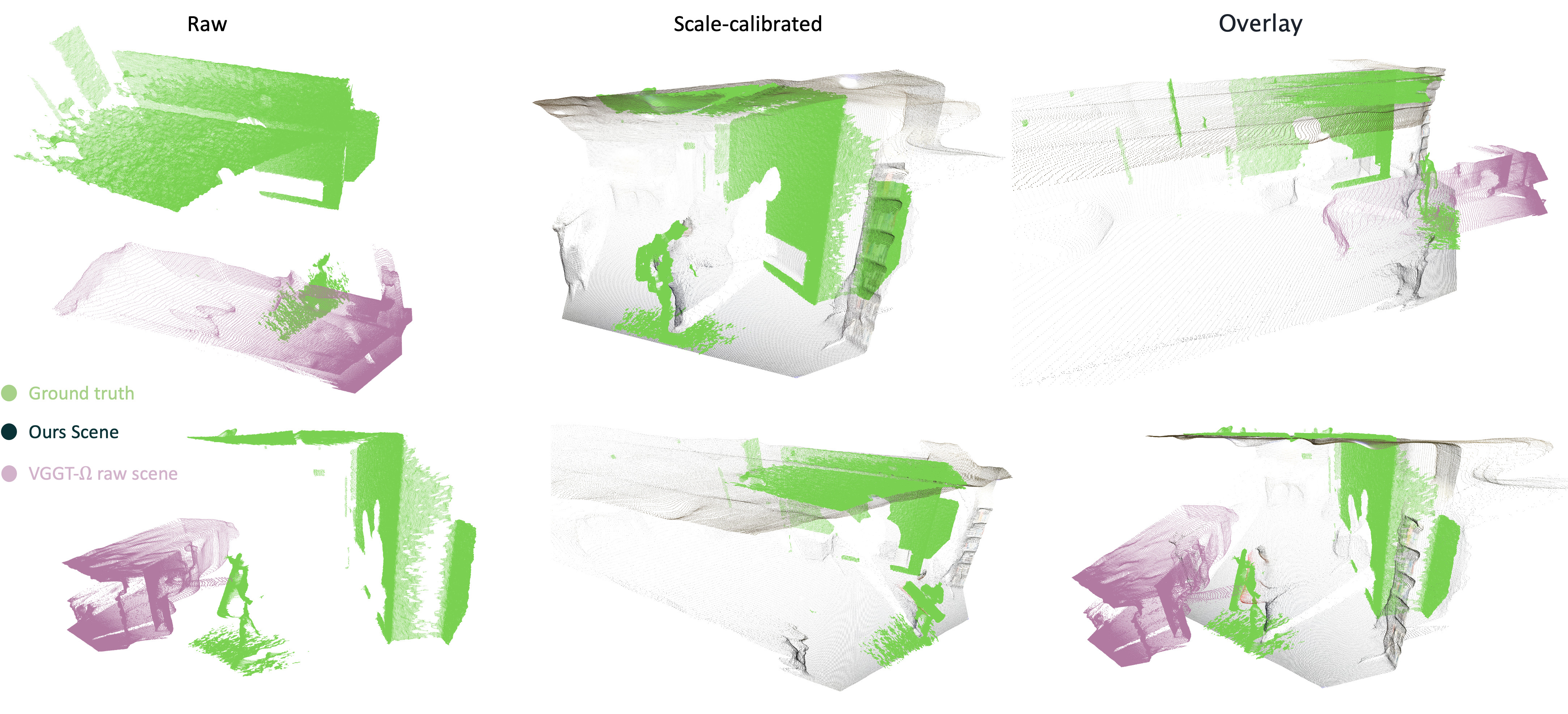}
    \vspace{-8mm}
    \caption{
        \textbf{Qualitative effect of human-anchored scale calibration on BEHAVE.} Our method calibrates the normalized scene prediction toward the ground-truth metric scale.
    }
    \label{fig:hamc_ablation}
\end{figure*}

%% file: figs/SGTR_ablation.tex
\begin{figure*}[t]
    \centering
    \vspace{-4mm}
    
    \includegraphics[width=\linewidth]{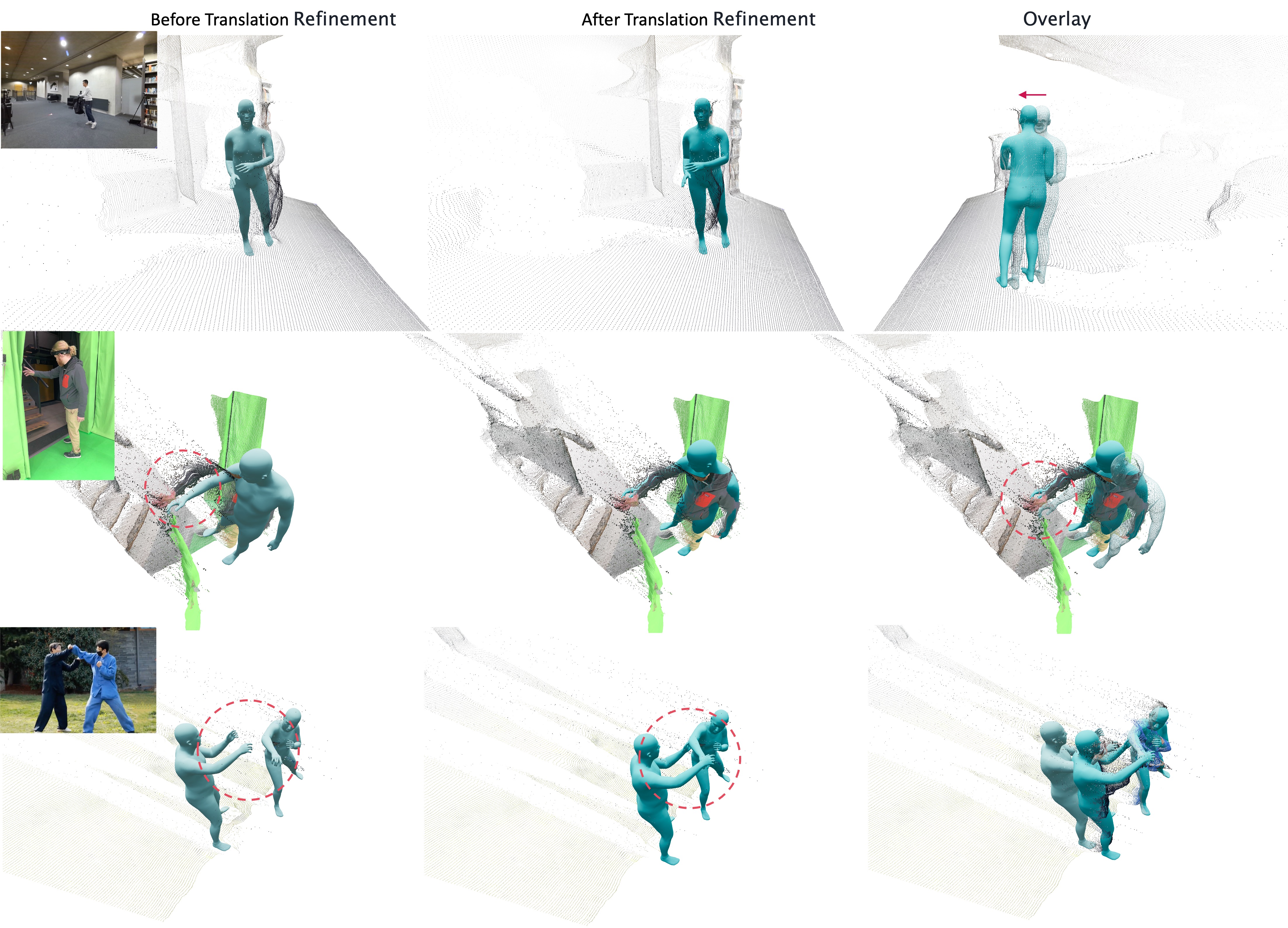}
    \vspace{-8mm}
    \caption{
        \textbf{Qualitative effect of scene-grounded translation refinement.} Our method refines the human translation to better align the mesh with the reconstructed scene.
        The last row shows that translation refinement improves multi-person contact and interaction.
    }
    \label{fig:sgtrablation}
\end{figure*}

%% file: tables/hac_comp_3db.tex
\begin{table*}[t]
    \centering

    \caption{
        \textbf{Comparison of contact-based and dense surface calibration} in human--scene alignment, metric scene and camera reconstruction, and global human motion.
        Both settings use \SAMThreeDBody{} and \VGGTOmega{}.
        BEDLAM2 scene metrics are evaluated at ground-truth depth $<70$ m.
    }
    \label{tab:hac_3db_comp}

    \setlength{\tabcolsep}{1.7pt}
    \renewcommand{\arraystretch}{1.05}

    \resizebox{\linewidth}{!}{    \begin{tabular}{@{}lcc|cc|ccccc|ccc@{}}
        \toprule

        \multirow{3}{*}{Method}
        & \multicolumn{4}{c|}{Human--Scene Alignment}
        & \multicolumn{5}{c|}{Metric Scene / Camera}
        & \multicolumn{3}{c}{Global Human Motion} \\

        \cmidrule(lr){2-5}
        \cmidrule(lr){6-10}
        \cmidrule(lr){11-13}

        & \multicolumn{2}{c|}{3DPW}
        & \multicolumn{2}{c|}{EMDB-1}
        & \multicolumn{5}{c|}{BEDLAM2}
        & \multicolumn{3}{c}{EMDB-2} \\

        \cmidrule(lr){2-3}
        \cmidrule(lr){4-5}
        \cmidrule(lr){6-10}
        \cmidrule(lr){11-13}

        & HS-CF$_5$ $\downarrow$
        & HS-CF$_{10}$ $\downarrow$
        & HS-CF$_5$ $\downarrow$
        & HS-CF$_{10}$ $\downarrow$

        & P-Err. $\downarrow$
        & AbsRel $\downarrow$
        & $\delta<1.25$ $\uparrow$
        & ATE $\downarrow$
        & RPE-t $\downarrow$

        & G-MPJPE $\downarrow$
        & W-MPJPE $\downarrow$
        & WA-MPJPE $\downarrow$ \\

        \midrule

        Contact-based (HAC)

                & 4.417
        & 4.864
        & 0.829
        & 1.031

                & 315.94
        & 0.254
        & 86.37
        & 29.16
        & 12.69

                & 1227.3
        & 156.0
        & \textbf{67.9} \\

        \midrule

        Dense Surface (Ours)

                & \textbf{0.641}
        & \textbf{0.856}
        & \textbf{0.401}
        & \textbf{0.601}

                & \textbf{143.30}
        & \textbf{0.131}
        & \textbf{88.11}
        & \textbf{16.55}
        & \textbf{7.24}

                & \textbf{1203.9}
        & \textbf{152.9}
        & 68.1 \\

        \bottomrule
    \end{tabular}    }

\end{table*}

%% file: figs/hac.tex
\begin{figure*}[t]
    \centering
    \includegraphics[width=\linewidth]{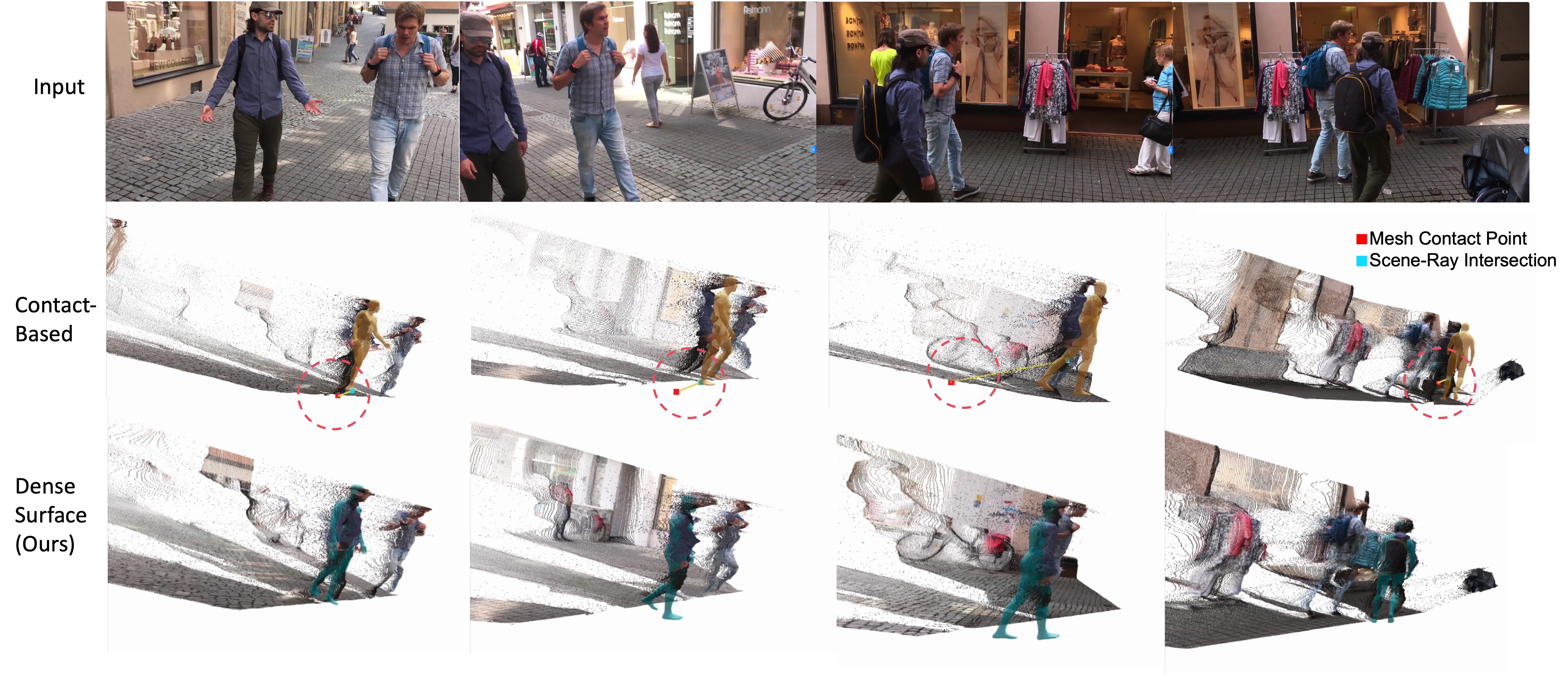}
        \caption{
        \textbf{Qualitative comparison of contact-based and dense surface correspondence calibration on 3DPW.}
        Both variants use the same human and scene backbones.
        Middle: the contact-based method exhibits scale drift under lower-body truncation.
        Bottom: our dense surface correspondence remains effective in challenging scenes.
        }
    \label{fig:hac}
\end{figure*}

%% file: tables/scene_ablation.tex
\begin{table*}[t]
    \centering

    \caption{
        \textbf{Ablation of scene chunk stitching and shared-scale selection}
        for metric scene reconstruction and global human motion.
        All settings use \MHMRtwo{}~\citep{fiche2026multi} as the human backbone.
        For \PiThreeX{}, we compare SE(3) and Sim(3) chunk stitching.
        For \VGGTOmega{}, we compare sequence-wide scale estimation
        with our streaming first-chunk scale selection.
                        The best result within each backbone block is boldfaced.
    }
    \label{tab:scene_ablation}

    \setlength{\tabcolsep}{3.0pt}
    \renewcommand{\arraystretch}{1.05}

    \resizebox{\linewidth}{!}{    \begin{tabular}{@{}llccccc|ccc@{}}
        \toprule

        \multirow{3}{*}{Scene Backbone}
        & \multirow{3}{*}{Setting}
        & \multicolumn{5}{c|}{Metric Scene Reconstruction}
        & \multicolumn{3}{c}{Global Human Motion} \\

        \cmidrule(lr){3-7}
        \cmidrule(lr){8-10}

        &
        & \multicolumn{5}{c|}{Bonn}
        & \multicolumn{3}{c}{EMDB-2} \\

        \cmidrule(lr){3-7}
        \cmidrule(lr){8-10}

        &
        & P-Err. $\downarrow$
        & AbsRel $\downarrow$
        & $\delta < 1.25$ $\uparrow$
        & ATE $\downarrow$
        & RPE-t $\downarrow$

        & G-MPJPE $\downarrow$
        & W-MPJPE $\downarrow$
        & WA-MPJPE $\downarrow$ \\

        \midrule

        \multirow{2}{*}{\PiThreeX{}~\citep{wang2026pi}}
        & w/ SE(3)
        & 14.26
        & 0.0478
        & 98.66
        & \textbf{5.02}
        & \textbf{2.66}

        & 3491.6
        & 332.5
        & 120.3 \\

        & w/ Sim(3) (Ours)
        & \textbf{13.08}
        & \textbf{0.0360}
        & \textbf{98.67}
        & 6.15
        & 3.15

        & \textbf{3313.8}
        & \textbf{328.2}
        & \textbf{119.7} \\

        \midrule

        \multirow{2}{*}{\VGGTOmega{}~\citep{wang2026vggt}}
        & w/ All-seq. Scale
        & 17.21
        & 0.0655
        & 97.70
        & \textbf{3.50}
        & 2.17

        & \textbf{1449.3}
        & \textbf{198.7}
        & \textbf{75.5} \\

        & w/ First-chunk Scale (Ours)
        & \textbf{16.75}
        & \textbf{0.0640}
        & \textbf{97.72}
        & 3.58
        & \textbf{2.16}

        & 1580.1
        & 210.9
        & 78.9 \\

        \bottomrule
    \end{tabular}    }

\end{table*}

%% file: tables/runtime.tex
\begin{table*}[t]
    \centering

    \caption{
        Runtime comparison on a single NVIDIA RTX 4090.
        End-to-end inference FPS includes image preprocessing and is evaluated
        on BEHAVE videos at the original resolution of $2048\times1536$.
        Core method inference FPS excludes image preprocessing but includes
        model decoding and subsequent post-processing.
        For our streaming pipeline, we report both initial and steady-state FPS.
        \HumanThreeR{} and \UniSHMethod{} do not distinguish between these two
        stages, and thus a single FPS is reported for each configuration.
        Higher is better.
    }
    \label{tab:runtime}

    \setlength{\tabcolsep}{4.0pt}
    \renewcommand{\arraystretch}{1.08}

    \resizebox{\linewidth}{!}{    \begin{tabular}{@{}lccc|cc|cc@{}}
        \toprule

        \multirow{2}{*}{Method}
        & \multirow{2}{*}{Variant}
        & \multicolumn{2}{c|}{Core Input Resolution}
        & \multicolumn{2}{c|}{End-to-End Inference FPS $\uparrow$}
        & \multicolumn{2}{c}{Core Method Inference FPS $\uparrow$} \\

        \cmidrule(lr){3-4}
        \cmidrule(lr){5-6}
        \cmidrule(lr){7-8}

        &
        & Human
        & Scene
        & Initial
        & Steady
        & Initial
        & Steady \\

        \midrule

        \multirow{3}{*}{\HumanThreeR{}~\citep{chen2025human3r}}
        & 672S
        & $672\times672$
        & $512\times384$
        & \multicolumn{2}{c|}{9.8}
        & \multicolumn{2}{c}{13.4} \\

        & 672B
        & $672\times672$
        & $512\times384$
        & \multicolumn{2}{c|}{8.7}
        & \multicolumn{2}{c}{11.5} \\

        & 896L
        & $896\times896$
        & $512\times384$
        & \multicolumn{2}{c|}{4.3}
        & \multicolumn{2}{c}{4.9} \\

        \UniSHMethod{}~\citep{li2026unish}
        & --
        & $192\times256$
        & $518\times392$
        & \multicolumn{2}{c|}{5.8}
        & \multicolumn{2}{c}{6.2} \\

        \midrule

        \multirow{6}{*}{Ours}

        & \MHMRtwo{}+\DAThree{}~\citep{fiche2026multi,lin2025depth}
        & $768\times576$
        & $504\times378$
        & 5.2
        & 5.6
        & 7.4
        & 6.9 \\

        & \MHMRtwo{}+\PiThreeX{}~\citep{fiche2026multi,wang2026pi}
        & $768\times576$
        & $574\times434$
        & 5.1
        & 5.4
        & 7.3
        & 6.5 \\

        & \MHMRtwo{}+\VGGTOmega{}~\citep{fiche2026multi,wang2026vggt}
        & $768\times576$
        & $512\times384$
        & 6.2
        & 12.8
        & 11.5
        & 13.8 \\

        \addlinespace[1pt]

        & \FastSAMThreeDBody{}+\DAThree{}~\citep{yang2026fast,lin2025depth}
        & $512\times512$
        & $504\times378$
        & 4.7
        & 5.3
        & 5.5
        & 5.9 \\

        & \FastSAMThreeDBody{}+\PiThreeX{}~\citep{yang2026fast,wang2026pi}
        & $512\times512$
        & $574\times434$
        & 4.4
        & 4.7
        & 5.1
        & 5.7 \\

        & \FastSAMThreeDBody{}+\VGGTOmega{}~\citep{yang2026fast,wang2026vggt}
        & $512\times512$
        & $512\times384$
        & 5.5
        & 6.9
        & 6.9
        & 10.3 \\

        \bottomrule
    \end{tabular}    }

\end{table*}

%% file: figs/HSCF.tex
\begin{figure*}[t]
    \centering
    \includegraphics[width=\linewidth]{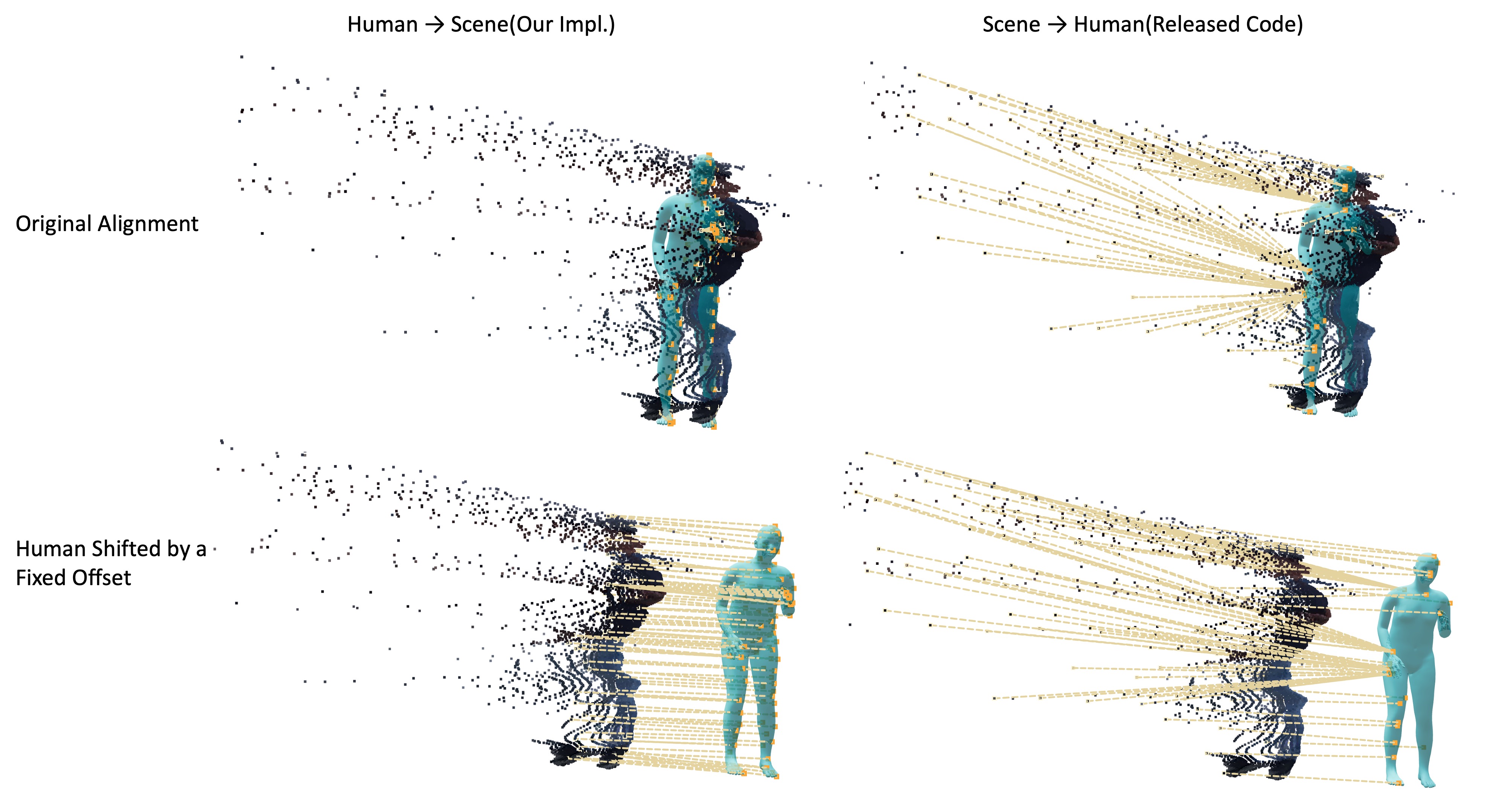}
    \caption{
        \textbf{Effect of the nearest-neighbor (NN) query direction on the HS-CF metric on 3DPW.}
        We visualize the correspondences (orange dashed lines) between the visible human surface and the scene points.
        \textbf{Left column (Human$\rightarrow$Scene):}
        The protocol defined in the SHOW paper and used in our implementation.
        \textbf{Right column (Scene$\rightarrow$Human):}
        The protocol from the officially released SHOW codebase.
        The top row shows the original alignment, while the bottom row applies a controlled translation to the human mesh to better visualize the correspondences.
    }
    \label{fig:hscf}
\end{figure*}

%% file: tables/HSCF.tex
\begin{wraptable}{r}{0.48\textwidth}
    \centering
    \vspace{-8pt}

    \caption{
        \textbf{Quantitative comparison of human--scene alignment under the \SHOWMethod{} released-code protocol}
        on 3DPW and EMDB-1 using HS-CF$_5$ and HS-CF$_{10}$.
        \bestmark~Best result; \secondmark~Second-best result.
        * denotes metrics that are directly copied from the original
        \SHOWMethod{} paper; these metrics are not directly comparable
        with the rest of the table.
    }
    \label{tab:hscf}

    \setlength{\tabcolsep}{3.0pt}
    \renewcommand{\arraystretch}{0.95}

    \resizebox{\linewidth}{!}{    \begin{tabular}{@{}lcc|cc@{}}
        \toprule
        \multirow{2}{*}{Method}
        & \multicolumn{2}{c|}{3DPW}
        & \multicolumn{2}{c}{EMDB-1} \\

        \cmidrule(lr){2-3}
        \cmidrule(lr){4-5}

        & HS-CF$_5$ $\downarrow$
        & HS-CF$_{10}$ $\downarrow$
        & HS-CF$_5$ $\downarrow$
        & HS-CF$_{10}$ $\downarrow$ \\

        \midrule

        \HumanThreeR{}$^{*}$~\citep{chen2025human3r}
        & 1.300 & 1.320
        & 1.130 & 1.310 \\

        \UniSHMethod{}$^{*}$~\citep{li2026unish}
        & 7.650 & 7.520
        & 7.100 & 6.990 \\

        \SHOWMethod{}$^{*}$~\citep{shi2026scene}
        & 5.690 & 5.440
        & 5.880 & 5.790 \\

        \midrule

        \HumanThreeR{}~\citep{chen2025human3r}
        & 2.419 & 2.275
        & 3.330 & 3.247 \\

        \UniSHMethod{}~\citep{li2026unish}
        & 7.096 & 7.016
        & 7.106 & 7.001 \\

        \midrule

        \multicolumn{5}{l}{\textbf{Ours}} \\

        \MHMRtwo{} + \DAThree{}~\citep{fiche2026multi,lin2025depth}
        & \cellsecond{1.673}
        & \cellsecond{1.473}
        & 2.520
        & 2.384 \\

        \MHMRtwo{} + \PiThreeX{}~\citep{fiche2026multi,wang2026pi}
        & \cellfirst{\textbf{1.363}}
        & \cellfirst{\textbf{1.229}}
        & \cellsecond{2.080}
        & \cellsecond{1.970} \\

        \MHMRtwo{} + \VGGTOmega{}~\citep{fiche2026multi,wang2026vggt}
        & 1.977
        & 1.751
        & \cellfirst{\textbf{2.013}}
        & \cellfirst{\textbf{1.924}} \\

        \bottomrule
    \end{tabular}    }

    \vspace{-10pt}
\end{wraptable}

%% file: figs/occlusion.tex
\begin{figure*}[t]
    \centering
    \includegraphics[width=\linewidth]{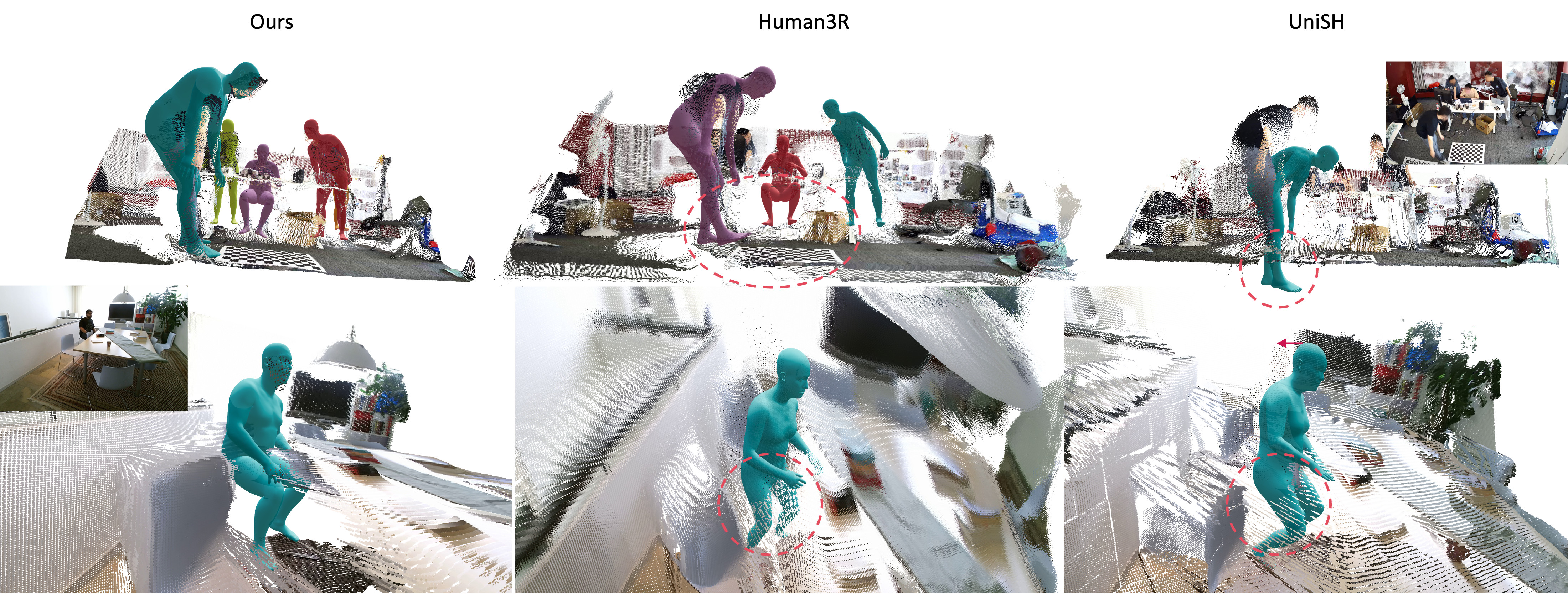}
    \vspace{-4mm}
    \caption{
        \textbf{Qualitative results under severe occlusion} on Internet videos and PROX.
        \OurMethod{} maintains better human--scene alignment and more plausible reconstruction of heavily occluded body parts.
    }
    \label{fig:occlusion}
\end{figure*}

%% file: figs/interaction.tex
\begin{figure*}[t]
    \centering
    \includegraphics[width=\linewidth]{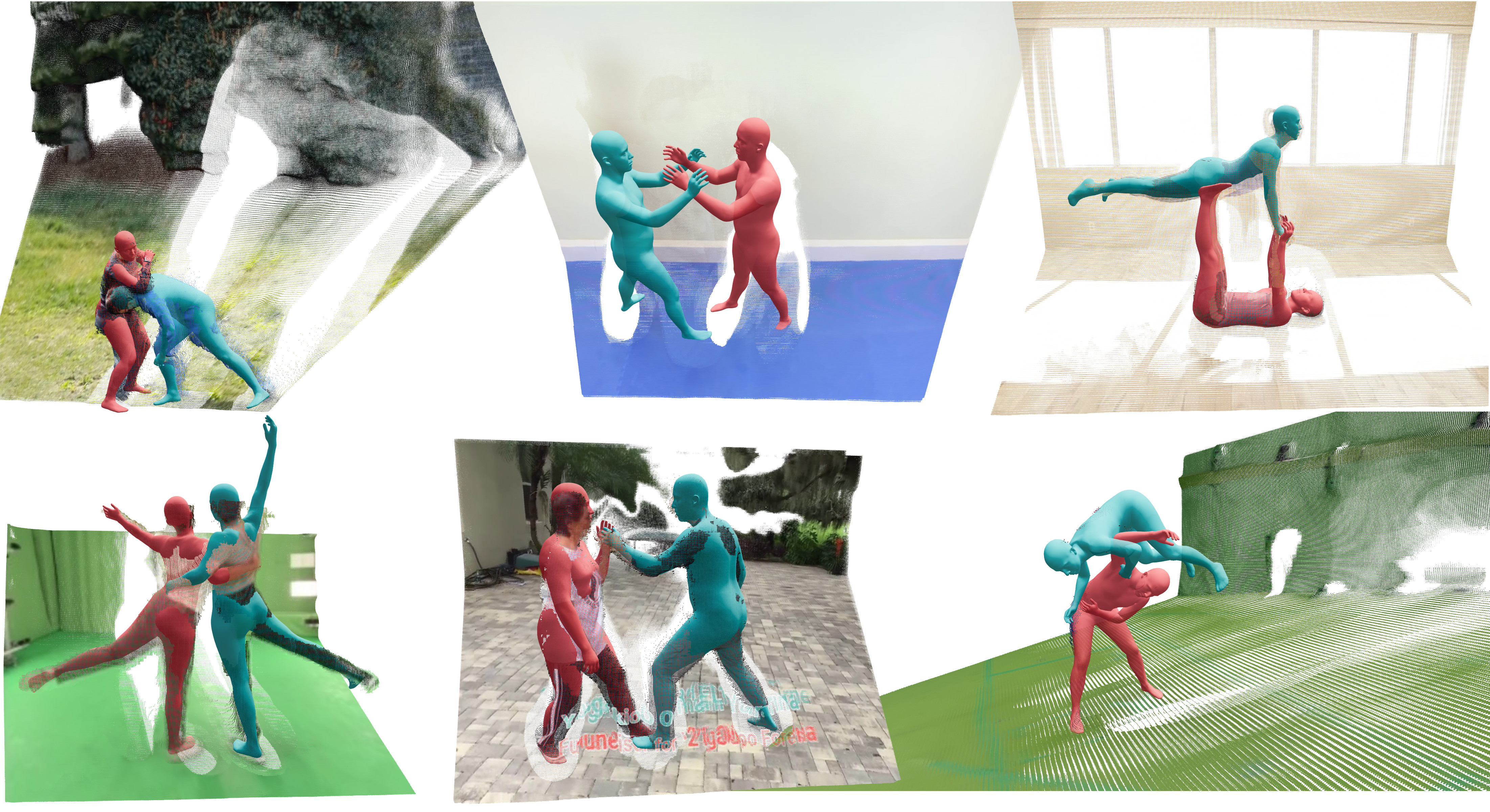}
    \vspace{-6mm}
    \caption{
        \textbf{Qualitative results for multi-person interaction} on Internet videos and Hi4D.
        Scene grounding improves relative human positioning and produces more plausible contact during interaction.
    }
    \label{fig:interaction}
\end{figure*}

%% file: figs/internet.tex
\begin{figure*}[t]
    \centering
    \includegraphics[width=\linewidth]{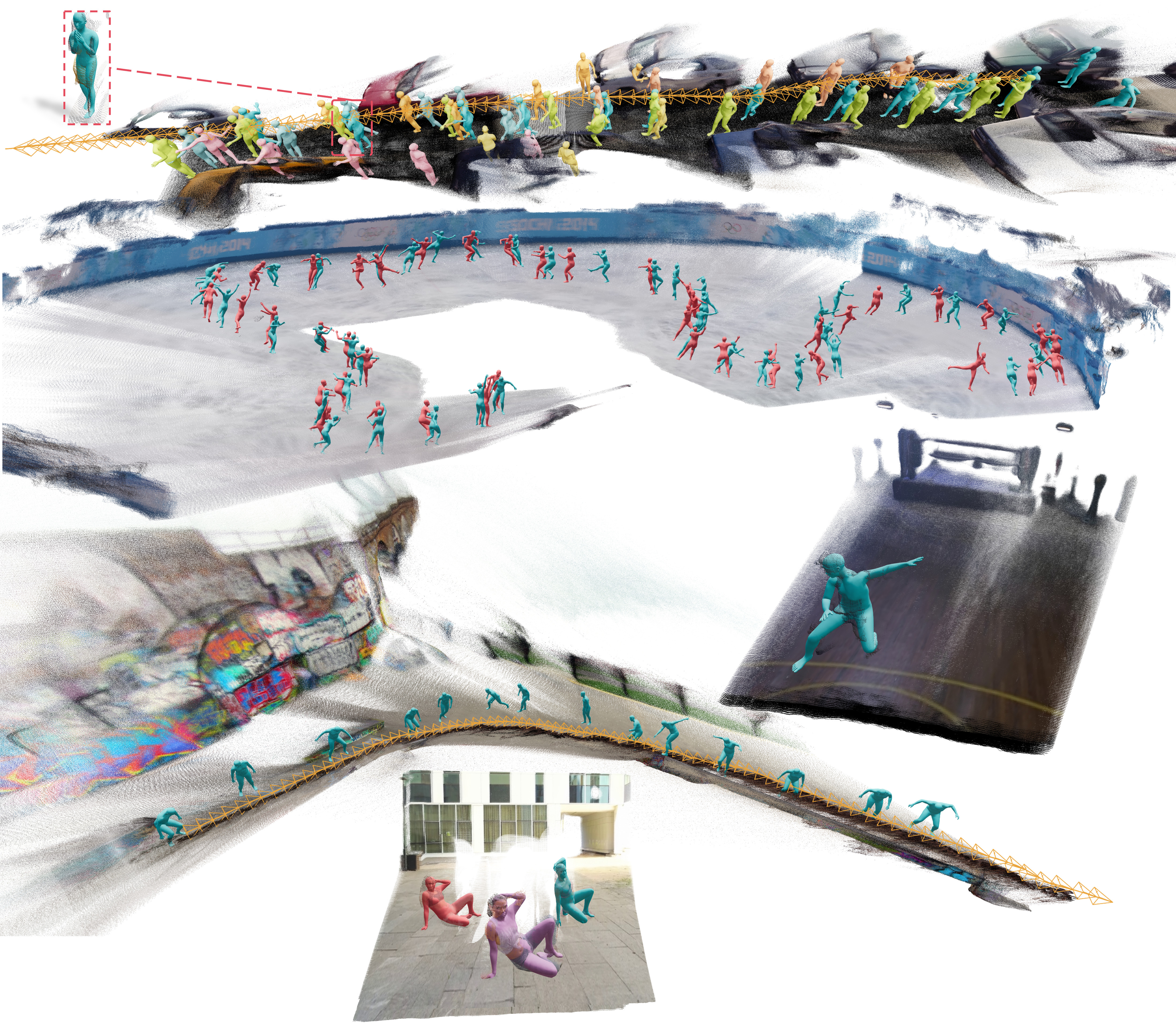}
    \vspace{-8mm}
    \caption{
        \textbf{Qualitative results on in-the-wild Internet videos.} \OurMethod{} maintains coherent human--scene alignment and long-term consistency across diverse multi-person, dancing, sports, and long-sequence videos.
    }
    \label{fig:internet}
\end{figure*}

%% file: figs/failure.tex
\begin{figure*}[t]
    \centering
    \vspace{-3mm}
    \includegraphics[width=\linewidth]{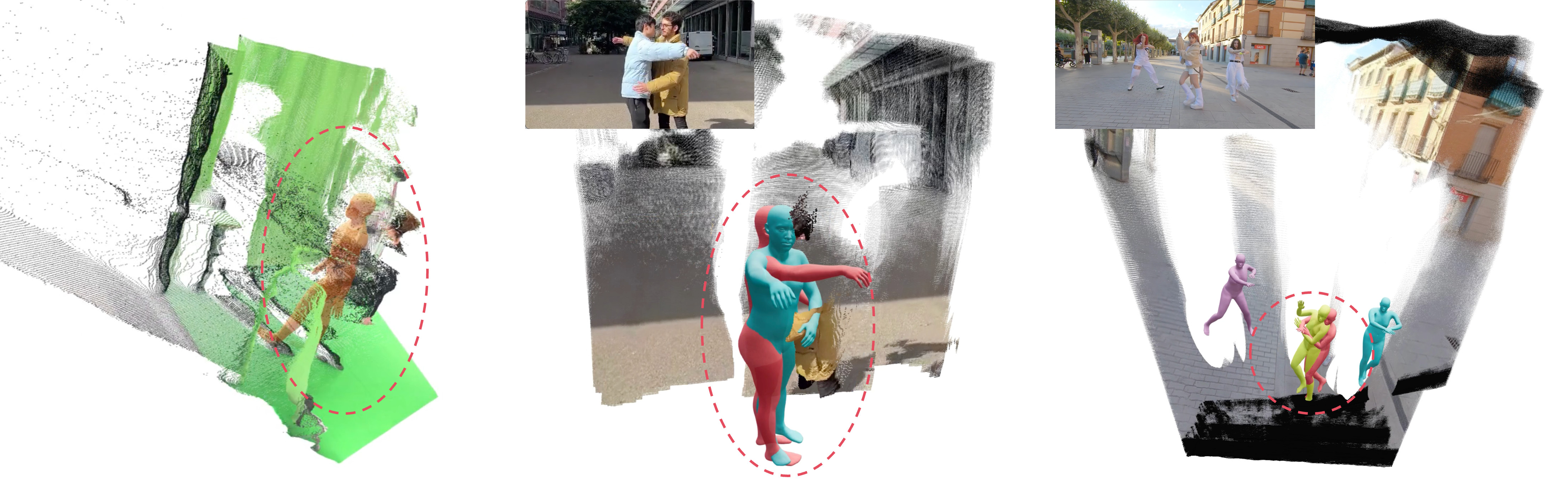}
    \vspace{-9mm}
    \caption{
       \textbf{Representative failure cases of \OurMethod{}.}
        Left: unreliable foreground geometry under challenging camera motion can cause occasional alignment errors.
        Right: close human interactions may lead to identity association errors in the underlying human reconstruction model.
    }
    \label{fig:failure}
\end{figure*}